\documentclass{article}

\PassOptionsToPackage{numbers, sort&compress}{natbib}
\usepackage[preprint]{neurips_2026}

\usepackage[utf8]{inputenc} % allow utf-8 input
\usepackage[T1]{fontenc}    % use 8-bit T1 fonts
\usepackage{hyperref}       % hyperlinks
\usepackage{url}            % simple URL typesetting
\usepackage{booktabs}       % professional-quality tables
\usepackage{amsfonts}       % blackboard math symbols
\usepackage{nicefrac}       % compact symbols for 1/2, etc.
\usepackage{microtype}      % microtypography
\usepackage{graphicx}
\usepackage[all]{hypcap}
\usepackage{siunitx}
\usepackage{xcolor}
\usepackage{listings}

\usepackage{amsmath}
\usepackage{adjustbox}

\usepackage{algorithm}
\usepackage{algpseudocode}
\usepackage{makecell}
\usepackage{float}

\graphicspath{{prelab/}}
\usepackage{multirow}

\title{NASDAQ: Normalized Observation Space Dynamics-Augmented Q-Learning}

\author{
  Xinwei Liu$^{1}$\quad
  Junyuan Liang$^{1}$\thanks{Corresponding author.}\quad
  Zicong Hong$^{2}$\quad
  Jianting Zhang$^{3}$\quad
  Wuhui Chen$^{1}$
    \\
  $^{1}${Sun Yat-sen University, China} \\
  $^{2}${EPFL, Switzerland} \\
  $^{3}${Purdue University, USA} \\
  \texttt{\{liuxw73, liangjy53\}@mail2.sysu.edu.cn}  \\
}

\begin{document}
\maketitle

\begin{abstract}
Augmenting model-free reinforcement learning (RL) with representations learned through observation dynamics prediction (observation-predictive RL) can improve sample efficiency and performance, with minor modifications and limited additional computation. However, this approach still struggles in challenging tasks with low-dimensional observations. In this paper, we identify a key factor behind this problem: unbalanced reconstruction losses across observation dimensions, where dimensions with larger value ranges dominate the loss. This encourages the agent to neglect dimensions with relatively small ranges, leading to degraded performance. To address this issue, we propose a novel normalization method tailored to online RL, which normalizes low-dimensional observations and balances the resulting losses and gradients. Beyond balancing reconstruction losses, observation normalization enables dynamics prediction to be performed in a normalized observation space, thereby providing a unified treatment of low- and high-dimensional inputs (e.g., physical states and images). Building on this idea, we further introduce Normalized Observation Space Dynamics-Augmented Q-learning (NASDAQ), a framework for observation-predictive RL applicable across diverse domains. NASDAQ learns state-action representations by coupling value learning with two auxiliary tasks: short-term value prediction and next \textit{normalized} observation prediction. Extensive experiments demonstrate that NASDAQ achieves competitive or superior performance compared with state-of-the-art model-based and self-predictive RL methods, while requiring significantly less training wall-time.
\end{abstract}

\section{Introduction}
Reinforcement learning (RL) \citep{sutton1998reinforcement} has proven to be a powerful framework for solving various sequential decision-making problems. However, model-free RL \citep{mnih2015human, schulman2017proximal, fujimoto2018addressing} is notoriously sample inefficient.
Model-based RL improves sample efficiency and performance by introducing world models \citep{hafner2019dream, hafner2023mastering} and planning \citep{hansen2022temporal, hansen2023td, schrittwieser2020mastering}, but incurs substantial computational costs.
Augmenting model-free RL with representations learned through observation dynamics prediction \citep{van2016stable, zhang1804decoupling, gelada2019deepmdp, ota2020can}, which we call observation-predictive RL, can also improve sample efficiency and performance, with less algorithmic and computational complexity. 
However, these approaches still struggle with challenging low-dimensional control tasks.
In contrast, self-predictive RL \citep{guo2020bootstrap, schwarzer2020data, fujimoto2023sale, fujimoto2025towards}, which learns representations by predicting latent embeddings of future observations instead, has recently shown promising results across a broader range of domains \citep{fujimoto2025towards}. 
This discrepancy raises the following questions: 
\begin{enumerate}
\item What is the bottleneck in observation-predictive RL?
\item Is it possible to enable observation-predictive RL to match or surpass model-based and self-predictive RL across diverse domains, while maintaining high computational efficiency?
\end{enumerate}

To answer these questions, we start by examining the training metrics of OFENet \citep{ota2020can}, an observation-predictive RL method, and identify the cause of its limited performance in challenging low-dimensional tasks. The root cause lies in unbalanced reconstruction losses across observation dimensions arising from differences in their value ranges. Consequently, the gradients induced by large losses dominate the overall gradient, encouraging the learned representations to neglect dimensions with relatively small ranges. 
To address this issue, we propose Shift-Adaptive Robust Observation Normalization (SARON), a novel normalization method for low-dimensional settings. SARON is tailored to online RL, accounting for distributional shifts and distributional variability induced by exploration. It normalizes streaming observations and balances the resulting reconstruction losses and gradients across dimensions.

Beyond mitigating this imbalance, observation normalization allows dynamics prediction to be performed in a normalized observation space, thereby providing a unified treatment of low- and high-dimensional observations (e.g., physical states and images). 
Building on this insight, we introduce Normalized Observation Space Dynamics-Augmented Q-learning (NASDAQ), a unified framework for observation-predictive RL applicable across diverse domains. Unlike complex self-predictive methods \citep{schwarzer2020data, fujimoto2023sale, fujimoto2025towards, guo2020bootstrap} that iteratively predict latent embeddings of future observations over extended horizons, NASDAQ learns state-action representations by combining value learning with two auxiliary tasks: short-term value prediction and next \textit{normalized} observation prediction.

We evaluate NASDAQ (with SARON for low-dimensional settings) across 87 environments from four widely used benchmarks. Extensive experiments demonstrate that our method achieves competitive or superior performance compared with state-of-the-art model-based \citep{hafner2023mastering, hansen2023td} and self-predictive approaches \citep{fujimoto2023sale, fujimoto2025towards, schwarzer2020data}, while requiring substantially less training wall-time. Interestingly, even without auxiliary tasks, NASDAQ with SARON remains a strong model-free RL baseline, trailing the state-of-the-art self-predictive method only slightly in low-dimensional settings.
We hope this work can inspire further research on observation-predictive and model-free methods.

\section{Related work}
\label{sec:related_work}
\paragraph{Data-efficient RL} A large body of work aims to improve sample efficiency and performance via dynamics modeling. Model-based methods incorporate world models into RL \citep{hafner2019dream, hafner2023mastering, hansen2022temporal, hansen2023td, schrittwieser2020mastering}. While these methods achieve impressive performance, they incur significant algorithmic and computational complexity. 
An alternative line of work \citep{fujimoto2023sale, fujimoto2025towards, guo2020bootstrap, schwarzer2020data, van2016stable, zhang1804decoupling, gelada2019deepmdp, ota2020can} augments model-free RL with representations learned through dynamics prediction without explicitly using world models.
We refer to this class of approaches as \textit{dynamics-augmented methods}, which can be divided into self-predictive RL \citep{fujimoto2023sale, fujimoto2025towards, guo2020bootstrap, schwarzer2020data} and observation-predictive RL \citep{van2016stable, zhang1804decoupling, gelada2019deepmdp, ota2020can}.
Self-predictive RL learns representations by predicting latent embeddings of future observations. Recently, the self-predictive method MR.Q \citep{fujimoto2025towards} has achieved performance competitive with model-based approaches, DreamerV3 \citep{hafner2023mastering} and TD-MPC2 \citep{hansen2023td}, across diverse domains. 
Inspired by model-based methods, MR.Q also includes model-based objectives, such as reward and termination predictions, in its representation learning.
Observation-predictive RL performs dynamics modeling directly in the observation space. Our framework, NASDAQ, falls into this category. The most related work to ours is OFENet \citep{ota2020can}, which learns state-action representations in a decoupled manner for downstream RL by predicting the next \textit{raw} observation. However, OFENet is designed for low-dimensional inputs and struggles in challenging tasks. In contrast, NASDAQ adopts coupled representation and value learning and addresses this bottleneck via a novel observation normalization method, while being broadly applicable across domains.

\paragraph{Normalization for RL} 
Prior studies have investigated various normalization techniques in RL \citep{ioffe2015batch, ba2016layer, bjorck2021towards, gogianu2021spectral, li2023normalization}. These methods are applied to internal components of neural networks. While observation normalization is commonly used in visual RL \citep{fujimoto2025towards, kostrikov2020image, yarats2021mastering}, it is often overlooked for low-dimensional observations. Its effects in such settings thus remain underexplored. A simple normalization method for low-dimensional tasks is implemented in the Stable-Baselines3 (SB3) library \citep{raffin2021stable}. The approach normalizes each dimension using the mean and standard deviation computed over all historical observations. In contrast, our proposed method, SARON, computes statistics by accounting for the characteristics of online RL. 

\paragraph{Self-predictive vs. observation-predictive RL}
Recent work has compared the two types of approaches. Ni et al. \citep{ni2024bridging} empirically show that self-predictive methods outperform observation-predictive ones in noisy or distracting tasks, whereas the opposite holds in sparse-reward tasks. Voelcker et al. \citep{voelcker2024does} theoretically show that features learned via observation prediction are generally superior to those learned via latent self-prediction, but the latter are more robust to observation perturbations. While these studies identify conditions under which observation-predictive RL is weak, they do not propose methods to address them. In this work, we empirically identify a key factor limiting observation-predictive RL and propose a solution.

\section{Background}
\label{sec:preliminaries}
RL problems are formulated as Markov Decision Processes (MDPs). An MDP can be described as a tuple \begin{math} (\mathcal{S}, \mathcal{A}, p, R, \gamma) \end{math}, where \begin{math} \mathcal{S} \end{math} and \begin{math} \mathcal{A} \end{math} respectively denote the state and action spaces, \begin{math} p(s_{t+1}|s_t,a_t) \end{math} is the transition dynamics, $R$ is a reward function, and \begin{math} \gamma \end{math} is a discount factor. The objective is to find a policy \begin{math} \pi: \mathcal{S} \rightarrow \mathcal{A} \end{math} that maximizes \begin{math} \sum_{t=0}^{\infty} \gamma^{t} r_{t} \end{math}, the cumulative discounted return when following the policy. Value-based RL methods learn a value function $ Q^{\pi}(s,a):= \mathbb{E}_{\pi}[\sum_{t=0}^{\infty} \gamma^{t} r_{t}|s_0=s,a_0=a] $ that models the expected return starting from an initial state \begin{math}s\end{math} and action \begin{math}a\end{math}. 

\section{Understanding the bottleneck in observation-predictive RL}
\label{sec:understanding}
In this section, we identify a key bottleneck in OFENet \citep{ota2020can}, a strong observation-predictive RL method that learns state-action representations separately from downstream model-free RL algorithms (e.g., TD3 \citep{fujimoto2018addressing}, PPO \citep{schulman2017proximal}, and SAC \citep{haarnoja2018soft}). We consider the OFENet+TD3 variant in our experiments. The method is evaluated on a compact diagnostic set of six low-dimensional tasks, comprising five OpenAI Gym tasks \citep{towers2024gymnasium} used in OFENet and the \textit{dog-run} task from the DeepMind Control suite (DMC) \citep{tassa2018deepmind}. Experimental setup and implementation details are deferred to Appendix~\ref{app:benchmark} and~\ref{app:imple_detail}, respectively.

\begin{table}[htbp]
    \caption{
        Final performance and training metrics of OFENet+TD3 at the end of training over 5 seeds. The {\textcolor{gray}{[bracketed values]}} represent a 95\% bootstrap confidence interval. The performance of MR.Q is included for comparison. 
    \textbf{{Bold numbers}} indicate better results.
    }
    \label{tab:table1}
    \centering
    % \begin{adjustbox}{max width=\textwidth}
    \begin{tabular}{lccc}
        \toprule
        \textbf{Tasks} & \textbf{MR.Q} & \textbf{OFENet+TD3} & \textbf{Auxiliary loss}  \\
        \midrule
        % \multicolumn{4}{l}{\textbf{Gym}} \\
        % \midrule
        Ant-v5 & 7514 {\textcolor{gray}{[6977, 7911]}} & \textbf{{8156}} {\textcolor{gray}{[8047, 8280]}} & 0.015 {\textcolor{gray}{[0.013, 0.016]}} \\
        Hopper-v5 & 2699 {\textcolor{gray}{[2245, 3179]}} & \textbf{{2853}} {\textcolor{gray}{[1837, 3637]}} & 4.8e-4 {\textcolor{gray}{[3.6, 5.9]e-4}}  \\
        Walker2d-v5 & 5431 {\textcolor{gray}{[4793, 5998]}} & \textbf{{6009}} {\textcolor{gray}{[5812, 6206]}} & 0.054 {\textcolor{gray}{[0.044, 0.067]}} \\
        HalfCheetah-v5 & \textbf{{13823}} {\textcolor{gray}{[13412, 14251]}} & 13548 {\textcolor{gray}{[12867, 14309]}} & 0.16 {\textcolor{gray}{[0.13, 0.19]}} \\
        Humanoid-v5 & \textbf{{7207}} {\textcolor{gray}{[5950, 8147]}} & 6063 {\textcolor{gray}{[5853, 6273]}} & 139 {\textcolor{gray}{[133, 149]}}  \\
        % \midrule
        % \multicolumn{4}{l}{\textbf{DMC}} \\
        % \midrule
        dog-run & \textbf{{300}} {\textcolor{gray}{[281, 319]}} & 66 {\textcolor{gray}{[22, 110]}} & 272 {\textcolor{gray}{[130, 436]}} \\
        \bottomrule
    \end{tabular}
    % \end{adjustbox}
\end{table}

\begin{figure}[htbp]
\includegraphics[width=0.32\linewidth]{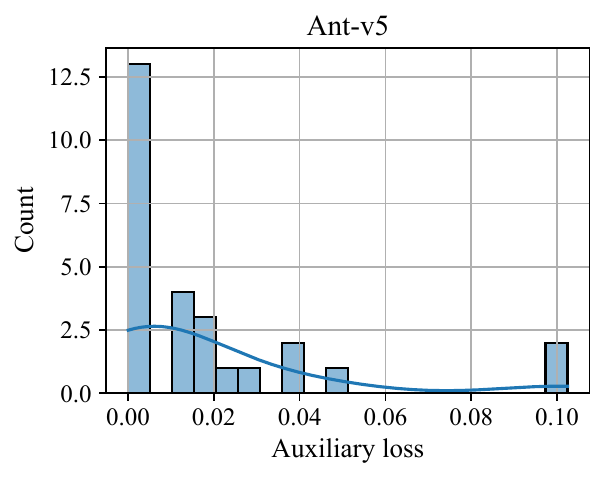}
\includegraphics[width=0.32\linewidth]{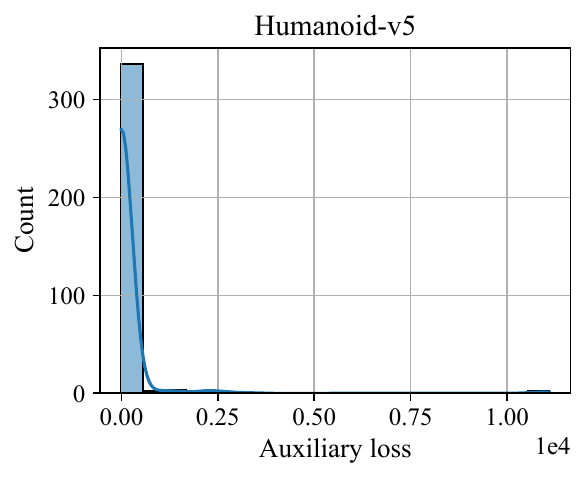}
\includegraphics[width=0.32\linewidth]{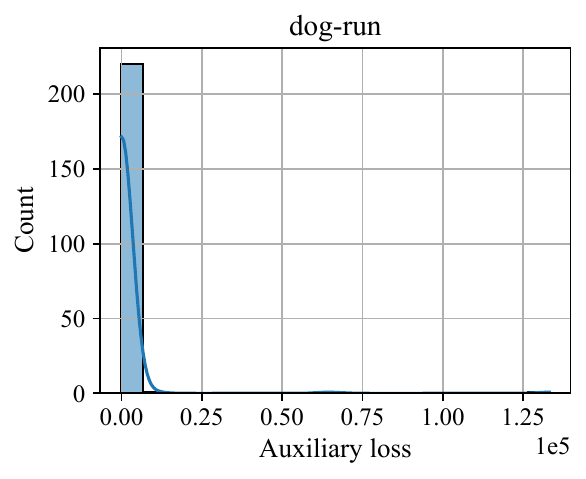}

\includegraphics[width=0.32\linewidth]{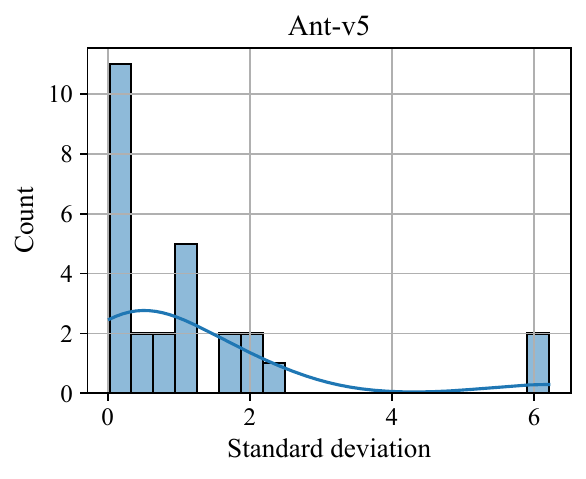}
\includegraphics[width=0.32\linewidth]{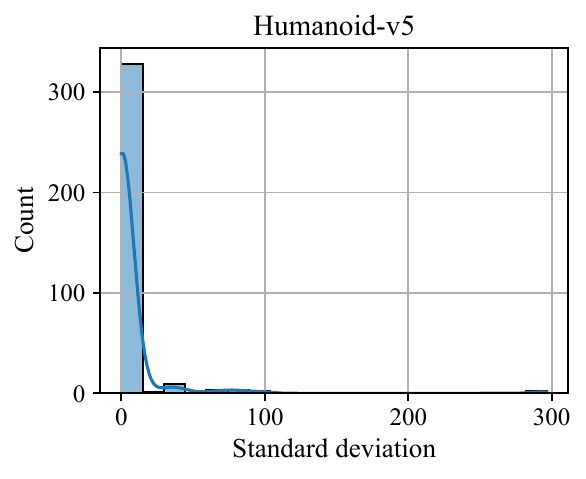}
\includegraphics[width=0.32\linewidth]{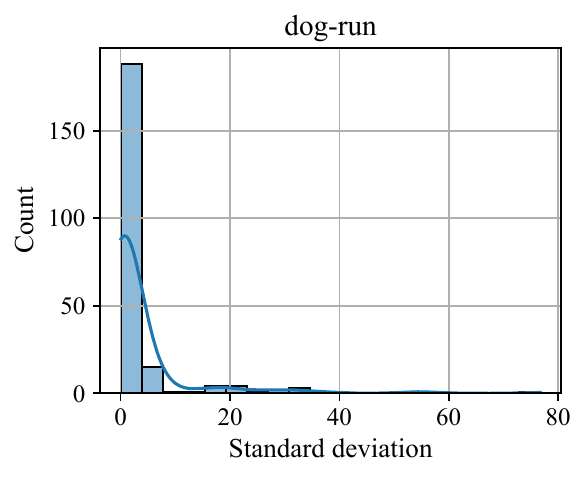}

\caption{Histograms and kernel density–estimated PDFs of per-dimension statistics, obtained from a single run of OFENet+TD3 on three tasks. 
\textit{Top}: distribution of auxiliary losses for each observation dimension, where the value for each dimension is computed by averaging the recorded auxiliary losses over the final 20k time steps.
\textit{Bottom}: distribution of per-dimension standard deviations, computed from the final 20k samples. See Appendix~\ref{app:add_result} for results on the remaining tasks.}
\label{tab:prelab_fig1}
\end{figure}

In Table~\ref{tab:table1}, we examine the auxiliary loss (observation reconstruction loss) of OFENet+TD3 and compare its performance with that of MR.Q \citep{fujimoto2025towards}, a state-of-the-art self-predictive RL method. 
As reported in OFENet, smaller auxiliary losses weakly correlate with better performance. Notably, auxiliary losses on challenging tasks (\textit{Humanoid-v5} and \textit{dog-run}) are far larger than those on the other tasks. To investigate the underlying cause, we further analyze auxiliary losses across observation dimensions. As shown in Figure~\ref{tab:prelab_fig1}, auxiliary losses across dimensions exhibit a skewed distribution with a heavy tail in challenging tasks.
This indicates that the large auxiliary loss is primarily driven by a small subset of dimensions with disproportionately high error contributions.
Moreover, the auxiliary loss distribution mirrors that of standard deviation, suggesting a positive correlation between a dimension’s value range and its auxiliary loss. A quantitative assessment of this correlation is provided in Figure~\ref{tab:prelab_scatter} of Appendix~\ref{app:add_result}. Taken together, these findings suggest that a key bottleneck in observation-predictive RL is unbalanced reconstruction losses across dimensions due to their value ranges. As a result, the overall gradient is dominated by gradients from large-loss dimensions, encouraging the learned representations to ignore dimensions of relatively small ranges.

To some extent, our findings are consistent with prior theoretical results that observation-predictive methods are more vulnerable than self-predictive ones when observation vectors are distracted by multiplying a random binary matrix \citep{voelcker2024does}. When the learned representations are encouraged to focus on a subset of observation dimensions, the effect resembles multiplying the observation vector by a diagonal binary matrix.

\section{Methods}
\label{sec:methods}
\subsection{Normalization for low-dimensional observations}
\label{subsec:SARON}
Motivated by the insight obtained in Section~\ref{sec:understanding}, we aim to balance the auxiliary losses across observation dimensions. Specifically, we normalize each dimension by its mean and standard deviation. A simple approach is to compute these statistics over all historical observations \citep{raffin2021stable}. However, such a method neglects both distributional shifts during online RL and distributional variability induced by exploration. Therefore, we propose Shift-Adaptive Robust Observation Normalization (SARON), a novel normalization method designed for the online RL paradigm.
SARON computes episode-wise means $\mu$ and uncentered second moments $\nu$, and maintains their exponential moving averages (EMAs) with coefficient $\beta$ to obtain the running statistics: 
\begin{equation}
\label{eq:1}
\begin{aligned}
m \gets \beta m + (1-\beta)\mu, \quad\quad s \gets \beta s + (1-\beta)\nu.
\end{aligned}
\end{equation}
As EMA statistics ($m$ and $s$) are initialized to zero, leading to biased estimates in early updates, we apply bias correction to these statistics following Adam \citep{diederik2014adam}:
\begin{equation}
\label{eq:2}
\begin{aligned}
\hat{m} = \frac m {1-\beta^k} , \quad\quad \hat{s} = \frac s {1-\beta^k},
\end{aligned}
\end{equation}
where $k$ denotes the number of EMA updates applied so far. Raw observations $o$ are normalized using these corrected statistics and clipped to $[-O, O]$ for numerical stability:
\begin{equation}
\label{eq:3}
\begin{aligned}
\tilde{o} =\operatorname{clip}( \frac {o - \hat{m}} {\hat{\sigma}}, -O, O) , \quad\quad \hat{\sigma} = \sqrt{\hat{s}-{\hat{m}}^2},
\end{aligned}
\end{equation}
where $\hat{\sigma}$ is the corrected standard deviation.
To obtain robust statistics from noisy online observation streams, we exclude episode-wise statistics associated with abnormally low returns from the EMA updates. These statistics, probably induced by exploration, are treated as outliers and identified using the interquartile range (IQR) over recent returns. 

\begin{figure}[t]
    \centering
    \begin{minipage}[t]{0.95\textwidth}
        \centering
        \includegraphics[width=\linewidth]{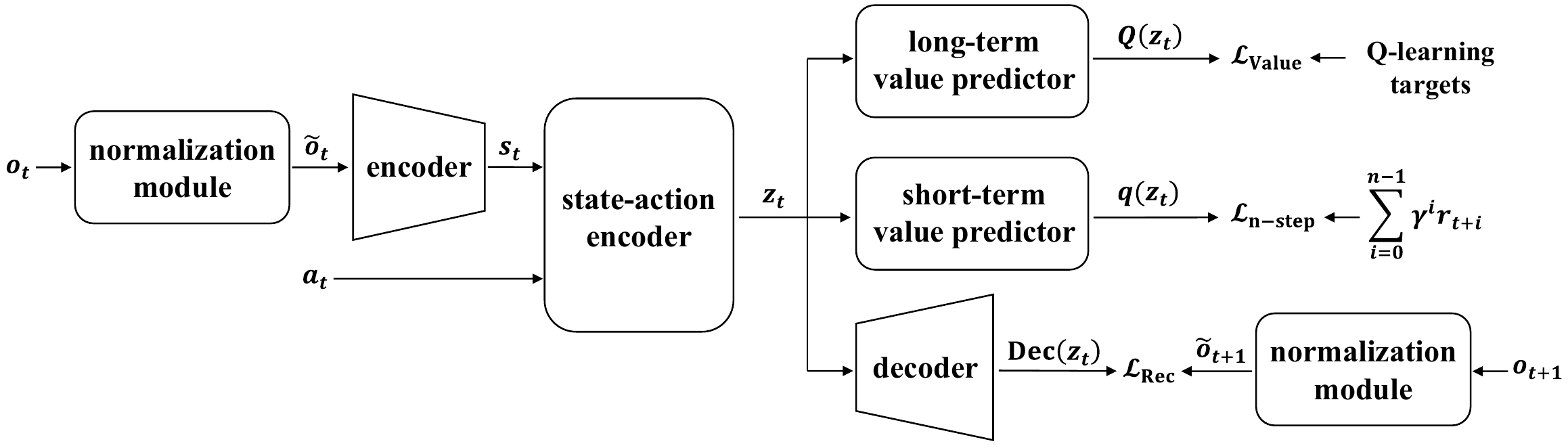}\\
        {(a) Value network with two auxiliary prediction heads}
        \label{framework:1}
    \end{minipage}

    \begin{minipage}[t]{0.95\textwidth}
        \centering
        \includegraphics[width=0.6\linewidth]{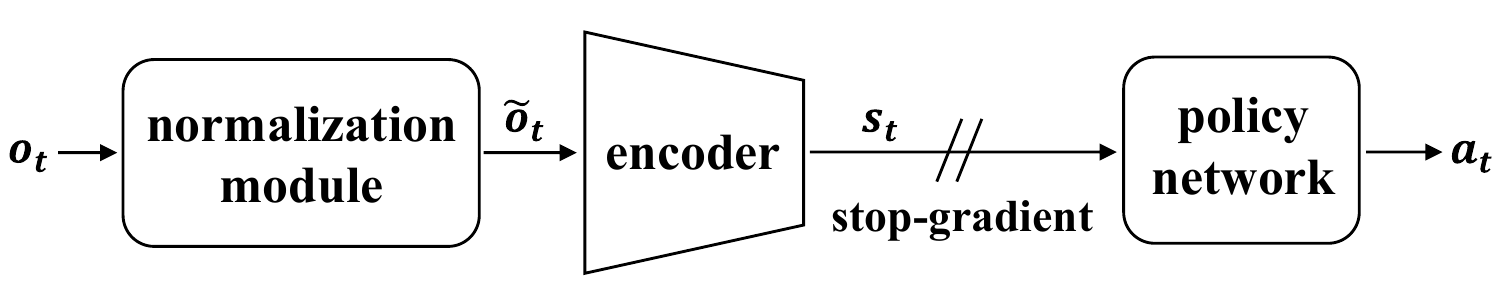}\\
        {(b) Policy network with detached inputs}
        \label{framework:2}
    \end{minipage}

    \caption{Overview of the NASDAQ framework. (a) The value network contains three predictors for value learning (long-term value prediction), and two auxiliary tasks of predicting the short-term value and the next \textit{normalized} observation. 
    (b) The gradients from the policy network are not backpropagated to the observation encoder.}
    \label{fig:framework}
\end{figure}

\subsection{Normalized observation space dynamics-augmented Q-learning (NASDAQ)}
\label{subsec:NASDAQ}
Beyond balancing auxiliary losses, observation normalization enables dynamics prediction to be formulated in a normalized observation space, thereby providing a unified way to learn representations from low- and high-dimensional observations (e.g., physical states and images). Motivated by this perspective, we introduce Normalized Observation Space Dynamics-Augmented Q-learning (NASDAQ), a unified framework for observation-predictive RL applicable to diverse domains. As illustrated in Figure~\ref{fig:framework}, the raw observation $o$ is first normalized by a normalization module, e.g., min-max normalization for pixel inputs, and SARON for low-dimensional observations. 
The normalized observation $\tilde{o}$ is then fed to an encoder $f$ to obtain a latent state $s$, which is further processed by a state-action encoder $g$ together with the action $a$ to produce a state-action representation $z$:
\begin{equation}
\label{eq:4}
\begin{aligned}
s = f(\tilde{o}), \quad\quad z = g(s, a).
\end{aligned}
\end{equation}
$s$ is fed to a policy network with gradients stopped, whereas $z$ is fed to three predictors ($Q$, $q$, and $\operatorname{Dec}$) for value learning and two auxiliary tasks.

\paragraph{Value learning} The value target is constructed similarly to TD3 \citep{fujimoto2018addressing}. Specifically, we train two separate value networks $V_{1}$ and $V_{2}$, each with the same structure shown in Figure~\ref{fig:framework}. 
We use numerical subscripts to refer to components within each network (e.g., the encoder $f_1$ of $V_{1}$). Target networks, which are used to produce stationary prediction targets, are denoted by adding a prime to their online counterparts (e.g., $f'_1$ and $V'_{1}$).
Online and target policy networks ($\pi$ and $\pi'$) use the detached outputs from $f_1$ and $f'_1$, respectively.
All target parameters are periodically synchronized with their online counterparts \citep{fujimoto2023sale, fujimoto2025towards}. The target action is produced by the target policy network, perturbed by clipped Gaussian noise:
\begin{equation}
\label{eq:5}
\begin{aligned}
a^{\pi'}= \begin{cases}
\operatorname{argmax}{a'}  & \text{for discrete}\ a' \\
\operatorname{clip}(a',-1, 1)  & \text{for continuous}\ a'
\end{cases} \ ,
\quad a'=\pi'(s') + \operatorname{clip}(\epsilon, -c, c), \quad\epsilon \sim \mathcal{N}(0, \sigma^2),
\end{aligned}
\end{equation}
where $s' = f_1'(\tilde{o}')$. Discrete actions are represented by a one-hot encoding, with the Gaussian noise added to each dimension. Following prior work \citep{fujimoto2025towards, yarats2021mastering, hessel2018rainbow}, we predict $n$-step returns. Given a normalized transition $\tau = (\tilde{o}_t, a_t, r_{t:t+n-1}, \tilde{o}_{t+n})$, the value target $y$ is computed as:
\begin{equation}
\label{eq:6}
\begin{aligned}
y = \sum_{i=0}^{n-1} \gamma^i r_{t+i} + \gamma^n  \min_{j \in \{1,2\}}{Q'_{j}(z_{t+n}^{j})} , \quad\quad z_{t+n}^{j} = g'_{j}\left( s_{t+n}^{j},a_{t+n}^{\pi'} \right),
\end{aligned}
\end{equation}
where $s_{t+n}^{j} = f'_j(\tilde{o}_{t+n})$, and $a_{t+n}^{\pi'}$ is computed based on $\tilde{o}_{t+n}$ via Equation~\ref{eq:5}. The value learning loss is computed using the Huber loss \citep{huber1992robust} instead of mean squared error (MSE) to eliminate bias from prioritized sampling \citep{fujimoto2025towards, fujimoto2020equivalence}:
\begin{equation}
\label{eq:7}
\begin{aligned}
\mathcal{L}_\text{Value} = \operatorname{Huber}\left(Q_{1}, y\right) + \operatorname{Huber}\left(Q_{2}, y\right).
\end{aligned}
\end{equation}

\paragraph{Policy learning} 
The policy network $\pi$ is updated using the deterministic policy gradient \citep{silver2014deterministic}. Following prior work \citep{fujimoto2025towards, bjorck2021high}, we add L2 regularization to the pre-activation policy outputs $h^\pi$ to help avoid local minima when rewards and value estimates are sparse :
\begin{equation}
\label{eq:8}
\begin{aligned}
\mathcal{L}_\text{Policy} = - \frac{1}{2} \sum_{i \in\{1,2\}}{Q_{i}(z^{i})} + \lambda_\text{pre-activ} \|h^\pi\|_2^2,
\quad\quad a^{{\pi}} = \operatorname{activ}(h^{{\pi}}),
\end{aligned}
\end{equation}
where $z^i = g_i(f_i(\tilde{o}), a^\pi)$, and $a^\pi$ is computed from $\tilde{o}$ using online networks and Equation~\ref{eq:5} but without action noise.
To accommodate different action spaces, the \textit{activ} function is Tanh for continuous actions and Gumbel-SoftMax \citep{lowe2017multi, cianflone2019discrete, fujimoto2025towards} for discrete ones.

\paragraph{Auxiliary tasks} We introduce two auxiliary tasks in NASDAQ. The first task is short-term value prediction, where a predictor $q$ takes state-action representations $z$ as input and predicts the cumulative discounted reward over the next $n$ steps. Following prior work \citep{schrittwieser2020mastering, hafner2023mastering, fujimoto2025towards}, we optimize this task with a cross-entropy (CE) loss due to its empirical benefit in sparse-reward settings. The auxiliary loss is computed as the sum of CE losses between the predicted logits and a two-hot encoding of the target:  
\begin{equation}
\label{eq:9}
\begin{aligned}
\mathcal{L}_\text{n-step}
=
\sum_{i\in\{1,2\}}
\operatorname{CE}\left(q_i(z_t^i), \operatorname{TwoHot}(R)\right),\quad\quad R=\sum_{j=0}^{n-1} \gamma^j r_{t+j},
\end{aligned}
\end{equation}
where $R$ is the first term of the value target in Equation~\ref{eq:6}. This auxiliary task provides training signals of lower variance than those from value estimation, thereby stabilizing representation learning when value targets exhibit high variance. The second auxiliary task predicts the next \textit{normalized} observation by feeding $z$ to a decoder $\operatorname{Dec}$, with MSE loss to minimize reconstruction errors:
\begin{equation}
\label{eq:10}
\begin{aligned}
\mathcal{L}_\text{Rec} = \sum_{i\in\{1,2\}}
\left\|\operatorname{Dec}_i(z_t^i)-\tilde{o}_{t+1}\right\|_2^2.
\end{aligned}
\end{equation}
During training, gradients from the policy loss are not propagated to the observation encoder $f_1$ \citep{yarats2021mastering, bjorck2021towards, ota2020can}, whereas all components of $V_{1}$ and $V_{2}$ are jointly optimized using the combined loss $\mathcal{L}$:
\begin{equation}
\label{eq:11}
\begin{aligned}
\mathcal{L} = \mathcal{L}_\text{Value} + \lambda_\text{Rec}\mathcal{L}_\text{Rec} + \lambda_\text{n-step}\mathcal{L}_\text{n-step},
\end{aligned}
\end{equation}
where $\lambda_\text{Rec}$ and $\lambda_\text{n-step}$ denote the auxiliary loss weights. Pseudocode of SARON and NASDAQ is provided in Appendix~\ref{app:algo}.

\section{Experiments}
\label{sec:exp}
In our experiments, we aim to answer the following questions:
\begin{enumerate}
    \item How does NASDAQ (with SARON) compare with state-of-the-art data-efficient RL approaches in terms of performance and computational efficiency? (See Section~\ref{subsec:q1})
    \item Can observation normalization address the bottleneck in observation-predictive RL identified in Section~\ref{sec:understanding}? (See Section~\ref{subsec:q2})
    \item Do the two auxiliary tasks (\textit{normalized} observation prediction and short-term value prediction) used in NASDAQ improve RL performance? (See Section~\ref{subsec:q3})
\end{enumerate}

\subsection{Experimental setup}
\paragraph{Benchmarks} We evaluate NASDAQ (with SARON for low-dimensional tasks) on four widely used RL benchmarks, spanning continuous and discrete control, as well as low- and high-dimensional observations: (1) \textbf{Gym} \citep{towers2024gymnasium}, a set of five continuous control locomotion tasks with low-dimensional observations, (2) \textbf{DMC (proprioceptive)} \citep{tassa2018deepmind}, a collection of 28 continuous control tasks with low-dimensional proprioceptive observations, (3) \textbf{DMC (visual)}, the same tasks as DMC (proprioceptive) but with image-based observations, (4) \textbf{Atari100k}, a benchmark of 26 Atari games \citep{bellemare2013arcade} for discrete control with pixel-based inputs. See Appendix~\ref{app:benchmark} for the evaluation protocol and a complete description of the benchmarks.

\paragraph{RL baselines} We consider the following data-efficient RL methods as baselines: (1) \textbf{OFENet+TD3} \citep{ota2020can}, an observation-predictive RL method for low-dimensional continuous control tasks, (2) \textbf{TD7} \citep{fujimoto2023sale}, a self-predictive method built on OFENet and TD3, which achieves state-of-the-art results on the Gym benchmark, (3) \textbf{SPR} \citep{schwarzer2020data}, a self-predictive method for visual discrete control tasks, (4) \textbf{TD-MPC2} \citep{hansen2023td}, a strong model-based method for continuous control tasks, (5) \textbf{DreamerV3} \citep{hafner2023mastering}, a strong model-based RL algorithm that performs well across diverse domains, (6) \textbf{MR.Q} \citep{fujimoto2025towards}, a state-of-the-art self-predictive method that achieves performance competitive with model-based RL approaches, DreamerV3 and TD-MPC2, across diverse benchmarks,
(7) \textbf{DrQ-v2} \citep{yarats2021mastering}, a simple yet strong model-free RL baseline for visual continuous control tasks, (8) \textbf{Rainbow (enhanced)}, an improved variant of Rainbow \citep{hessel2018rainbow} tailored for Atari100k. Details on the RL baselines are provided in Appendix~\ref{app:baselines}.

\paragraph{Observation normalization (ON) baseline} We consider a simple observation normalization (simple ON) method implemented in SB3 \citep{raffin2021stable}, which normalizes observations using means and standard deviations computed over all historical observations.  

\paragraph{Implementation overview} 
To enable fair comparisons with other TD3‑based methods (OFENet+TD3, TD7, and MR.Q), we take two control measures. First, we align the design choices of OFENet+TD3 and NASDAQ with those of MR.Q for components unrelated to representation learning (e.g., replay buffer and optimizer). Second, since these TD3‑based methods share similar principles for integrating learned representations into model‑free RL (whether in a coupled or decoupled manner), we control model capacity by keeping parameters of the RL components comparable across methods. Parameters dedicated to representation learning are not constrained, as they are determined by each method’s design. These controls allow performance differences to be more reliably attributed to algorithmic choices rather than parameter scale or implementation discrepancies. Within each benchmark, all tasks share fixed auxiliary loss weights. A simple method for determining these weights is described in Appendix~\ref{app:hyper_params}. Network structures, implementation details, and the parameter control protocol are provided in Appendices~\ref{app:structure},~\ref{app:imple_detail}, and~\ref{app:para_control}, respectively.

\subsection{Comparison with state-of-the-art data-efficient methods}
\label{subsec:q1}
\paragraph{Performance comparison} As shown in Table~\ref{tab:main_results1}, NASDAQ with SARON achieves the best performance in low-dimensional settings. For pixel-based tasks, NASDAQ outperforms other methods on DMC (visual), as shown in Table~\ref{tab:main_results2}. On Atari100k, our method achieves performance competitive with the state-of-the-art self-predictive method MR.Q. While DreamerV3 outperforms NASDAQ on Atari100k, it employs a substantially larger model with approximately 200M parameters, which is 40$\times$ larger than ours.

\begin{table}[htbp]
    \caption{
    Aggregate results, mean, median, and interquartile mean (IQM), on the two low-dimensional benchmarks, Gym and DMC (proprioceptive). The scores on Gym are normalized by the scores of a \textit{deep} variant of TD3 before aggregation (see Appendix~\ref{app:benchmark}). \textbf{{Bold numbers}} indicate the best performance. Full results are provided in Appendices~\ref{app:full_gym} and~\ref{app:full_dmc_p}.
    }
    \centering

    \begin{adjustbox}{max width=\textwidth}
    \begin{tabular}{lccc c ccc}
        \toprule
        \multirow{2}{*}{\textbf{Methods}}  & \multicolumn{3}{c}{\textbf{Gym}} & & \multicolumn{3}{c}{\textbf{DMC (proprioceptive)}} \\
        \cmidrule(lr){2-4} \cmidrule(lr){6-8}
        & Mean & Median & IQM && Mean & Median & IQM \\
        \midrule

        NASDAQ with SARON (ours)
        & \textbf{1.79} 
        & 1.26
        & \textbf{1.50}
        &
        & \textbf{805}
        & \textbf{927}
        & \textbf{898} \\

        OFENET+TD3
        & 1.48
        & 1.24
        & 1.44
        &
        & 567
        & 696
        & 600 \\
        
        TD7
        & 1.16
        & \textbf{1.30}
        & 1.14
        &
        & 660
        & 810
        & 726 \\
    
        MR.Q
        & 1.50
        & 1.12
        & 1.36
        &
        & 791
        & 918
        & 886 \\ 

        TD-MPC2
        & 0.36
        & 0.20
        & 0.20
        &
        & 783
        & 896
        & 868 \\ 
        
        DreamerV3
        & 0.66
        & 0.62
        & 0.68
        &
        & 530
        & 700
        & 577 \\ 
        \bottomrule
    \end{tabular}
    \end{adjustbox}

\label{tab:main_results1}
\end{table}

\begin{table}[htbp]
    \caption{
    Aggregated results on two high-dimensional benchmarks. 
    The results on Atari100k are normalized by human scores before aggregation (see Appendix~\ref{app:benchmark}). \textbf{{Bold numbers}} indicate the best performance. The full results are provided in Appendices~\ref{app:full_dmc_v} and~\ref{app:full_atari}.
    }
    \centering

    \begin{adjustbox}{max width=\textwidth}
    \begin{tabular}{lccc c ccc}
        \toprule
        \multirow{2}{*}{\textbf{Methods}}  & \multicolumn{3}{c}{\textbf{DMC (visual)}} & & \multicolumn{3}{c}{\textbf{Atari100k}} \\
        \cmidrule(lr){2-4} \cmidrule(lr){6-8}
        & Mean & Median & IQM && Mean & Median & IQM \\
        \midrule

        NASDAQ (ours)
        & \textbf{606}
        & \textbf{836}
        & \textbf{701}
        &
        & 0.93
        & 0.34
        & 0.42 \\

        MR.Q
        & 598
        & 809
        & 686
        &
        & 0.91
        & 0.40
        & 0.41 \\
    
        DreamerV3
        & 463
        & 493
        & 452
        &
        & \textbf{1.25}
        & \textbf{0.49}
        & \textbf{0.54} \\ 

        DrQ-v2
        & 510
        & 626
        & 545
        &
        & -
        & -
        & - \\ 

        TD-MPC2
        & 492
        & 572
        & 501
        &
        & -
        & -
        & - \\ 

        SPR
        & -
        & -
        & -
        &
        & 0.65
        & 0.42
        & 0.43 \\ 
        
        Rainbow (enhanced)
        & -
        & -
        & -
        &
        & 0.54
        & 0.29
        & 0.35 \\ 
        \bottomrule
    \end{tabular}
    \end{adjustbox}

\label{tab:main_results2}
\end{table}

\begin{figure}[htbp]
    \centering
    % \hfill
    % 左边子图
    \begin{minipage}[t]{0.47\textwidth}
        \centering
        % \textbf{cheetah-run (DMC-proprioceptive)} \\
        \includegraphics[width=\linewidth]{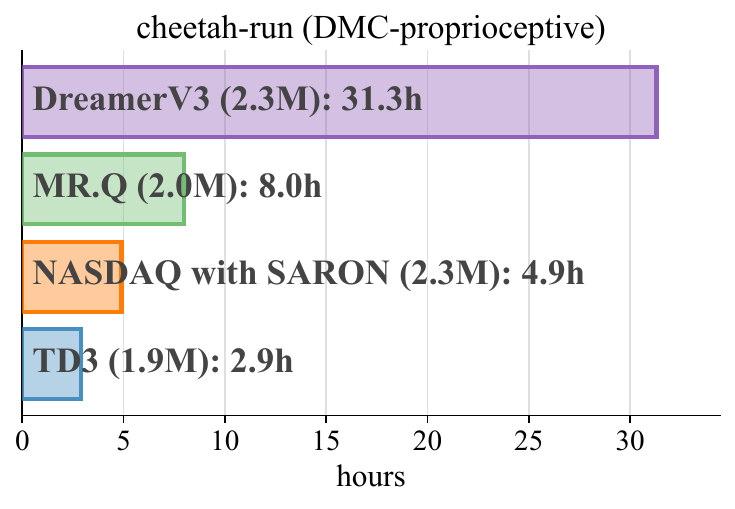}
    \end{minipage}
    % \hfill
    % 右边子图
    \begin{minipage}[t]{0.47\textwidth}
        \centering
        % \textbf{cheetah-run (DMC-visual)} \\
        \includegraphics[width=\linewidth]{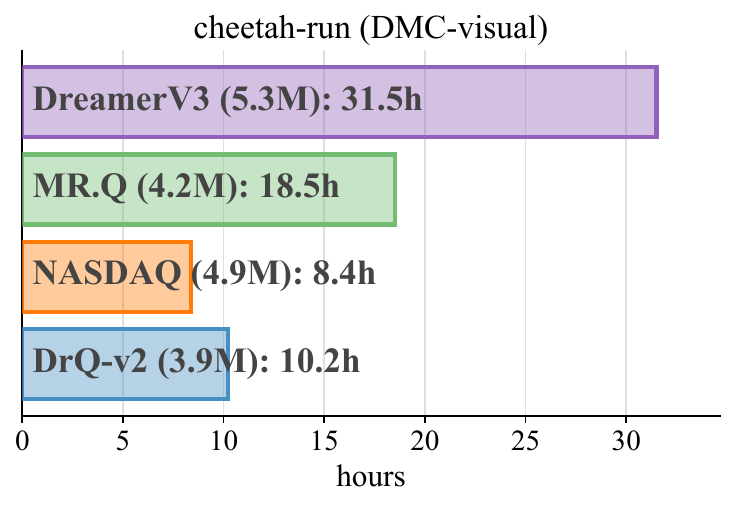}
    \end{minipage}
    % \hfill
    \caption{Training wall-time (in hours) comparisons on the \textit{cheetah-run} task in low- and high-dimensional settings. The model size (in millions) of each method is shown in parentheses.}
    \label{fig:runtime}
\end{figure}

\paragraph{Computational efficiency comparison} Figure~\ref{fig:runtime} compares the training wall-time of NASDAQ (with SARON) with that of two general-purpose RL algorithms (MR.Q \citep{fujimoto2025towards} and DreamerV3 \citep{hafner2023mastering}) and two model-free baselines (TD3 \citep{fujimoto2018addressing} and DrQ-v2 \citep{yarats2021mastering}). We scale down DreamerV3 to make its model size comparable to those of the other methods. All methods are implemented in PyTorch \citep{paszke2019pytorch} and run on a single RTX 4090 GPU.
In low-dimensional settings, NASDAQ with SARON has training wall-time closest to that of model-free baselines, indicating it incurs the least computational overhead among data-efficient methods. This efficiency stems from the fact that NASDAQ introduces only minor modifications to standard model-free RL. Notably, NASDAQ even trains faster than the model-free baseline in visual settings. The reason behind this interesting result is that NASDAQ allocates a large proportion of parameters to the observation encoder (17$\%$), whose computations are reused across learning objectives. In contrast, the observation encoder in DrQ-v2 accounts for only 0.7$\%$ of the total parameters. 

\subsection{On the effectiveness of observation normalization in low-dimensional settings}
\label{subsec:q2}
As shown in Table~\ref{tab:ab1}, SARON significantly improves OFENet+TD3.
It balances auxiliary losses across dimensions and addresses the identified bottleneck. The auxiliary-loss distribution of OFENet+TD3 with SARON is provided in Figure~\ref{tab:after_obs_norm} of Appendix~\ref{app:add_result}. Notably, observation normalization also benefits model-free RL even without any auxiliary tasks. NASDAQ with SARON (w/o auxiliary tasks) not only significantly outperforms its ON-ablated version but also achieves performance close to that of MR.Q. Note that MR.Q requires multi-step latent dynamics prediction. 
We thus recommend this variant as a strong model-free RL baseline for low-dimensional settings. Moreover, SARON is generally superior to simple ON. This highlights the importance of accounting for distributional shifts and distributional variability when normalizing online observation streams.

\begin{table}[htbp]
    \caption{
    Aggregated results on two low-dimensional benchmarks. We use $\uparrow$ and $\downarrow$ to denote the performance changes of SARON relative to the simple ON baseline. \textbf{Bold numbers} indicate the best performance. Full results are provided in Appendix~\ref{app:ab_on}.
    }
    \centering

    \begin{adjustbox}{max width=\textwidth}
    \begin{tabular}{lccc c ccc}
        \toprule
        \multirow{2}{*}{\textbf{Methods}}  & \multicolumn{3}{c}{\textbf{Gym}} & & \multicolumn{3}{c}{\textbf{DMC (proprioceptive)}} \\
        \cmidrule(lr){2-4} \cmidrule(lr){6-8}
        & Mean & Median & IQM && Mean & Median & IQM \\
        \midrule
        OFENET+TD3
        & 1.48 & 1.24 & 1.44 && 567 & 696 & 600 \\
        
        OFENET+TD3 with SARON
        & 1.64 & \textbf{1.29} & \textbf{1.54} && 587 & 738 & 656 \\
        
        \midrule
        NASDAQ \quad \quad \ (w/o ON,  w/o auxiliary tasks)
        & 1.29 & 1.20 & 1.35 && 593 & 762 & 645 \\
        
        NASDAQ with simple ON (w/o auxiliary tasks)
        & 1.63 & 1.21 & 1.40 && 713 & 808 & 804 \\
        
        NASDAQ with SARON \ \ (w/o auxiliary tasks)
        & 1.55 \rlap{$\,\downarrow$} & 1.22 \rlap{$\,\uparrow$} & 1.30 \rlap{$\,\downarrow$} && 744 \rlap{$\,\uparrow$} & 839 \rlap{$\,\uparrow$} & 834 \rlap{$\,\uparrow$} \\

        NASDAQ with simple ON
        & 1.75 & 1.25 & 1.50 && 773 & 906 & 886 \\
        
        NASDAQ with SARON  \ \ (complete)
        & \textbf{1.79} \rlap{$\,\uparrow$} & 1.26 \rlap{$\,\uparrow$} & 1.50 && \textbf{805} \rlap{$\,\uparrow$} & \textbf{927} \rlap{$\,\uparrow$} & \textbf{898} \rlap{$\,\uparrow$} \\
        
        \bottomrule
    \end{tabular}
    \end{adjustbox}
    \label{tab:ab1}
\end{table}

\subsection{Ablation study of auxiliary tasks}
\label{subsec:q3}
Table~\ref{tab:full_ab} summarizes the ablation results on the four benchmarks, where $\lambda >0$ and $\lambda=0$ denote that the auxiliary task is enabled and disabled, respectively. Details on how $\lambda$ is selected when $\lambda > 0$ are provided in Appendix~\ref{app:hyper_params}. Across all benchmarks, normalized observation prediction brings significant improvements over the pure model-free variant ($\lambda_\text{Rec} =0$, $\lambda_\text{n-step} =0$). While further introducing short-term value prediction ($\lambda_\text{Rec} >0$, $\lambda_\text{n-step} >0$) consistently improves performance on visual RL benchmarks, it leads to a slightly negative effect in non-visual settings. We hypothesize that this reflects a bias--variance trade-off. On the one hand, the auxiliary task provides lower-variance learning signals than those from value estimation, thereby stabilizing learning. This advantage is particularly pronounced in visual RL tasks, where bootstrapped targets tend to have high variance due to high-dimensional inputs. On the other hand, it may bias the agent toward short-term value prediction and interfere with long-term value estimation. This negative effect may exist in both visual and non-visual settings, but it becomes more evident in the latter, where bootstrapped targets have relatively low variance and the benefit of variance reduction is thus diminished.

\begin{table}[htbp]
    \caption{
     Aggregate performance of NASDAQ (with SARON) variants across benchmarks. 
     % $\lambda>0$ denotes the corresponding auxiliary task is included, whereas $\lambda=0$ means it is removed. 
     % The scores on Gym and Atari100k are normalized by that of TD3 and human, respectively, before aggregation (see Appendix~\ref{app:benchmark}). 
     \textbf{Bold numbers} indicate the best performance. The full results are provided in Appendix~\ref{app:full_ab}.
    }
    \centering

    \begin{adjustbox}{max width=\textwidth}
    \begin{tabular}{lccc c ccc c ccc c ccc}
        \toprule
        \multirow{2}{*}{\textbf{Versions}}  &
        \multicolumn{3}{c}{\textbf{Gym}} & &
        \multicolumn{3}{c}{\textbf{DMC (proprioceptive)}} & &
        \multicolumn{3}{c}{\textbf{DMC (visual)}} & & 
        \multicolumn{3}{c}{\textbf{Atari100k}} \\
        \cmidrule(lr){2-4} \cmidrule(lr){6-8} \cmidrule(lr){10-12} \cmidrule(lr){14-16}
        & Mean & Median & IQM && Mean & Median & IQM && Mean & Median & IQM && Mean & Median & IQM \\
        \midrule
        
        % \multicolumn{8}{l}{\quad \textit{NASDAQ-based methods (ours)}} \\
        $\lambda_\text{Rec} =0$, $\lambda_\text{n-step} =0$
        & 1.55 & 1.22 & 1.30 && 744 & 839 & 834&& 422 & 372 & 375&& 0.38 & 0.16 & 0.23 \\
        
        $\lambda_\text{Rec} >0$, $\lambda_\text{n-step} =0$
        & \textbf{1.79} & \textbf{1.26} & \textbf{1.50} && \textbf{805} & \textbf{927} & \textbf{898}  && 589  & 792  & 668  && 0.86  & 0.16  & 0.36\\

        $\lambda_\text{Rec} >0$, $\lambda_\text{n-step} >0$
        & 1.78 & 1.21 & 1.47 && 762 & 899 & 874  && \textbf{606}  & \textbf{836}  & \textbf{701}  && \textbf{0.93}  & \textbf{0.34}  & \textbf{0.42} \\
        
        \bottomrule
    \end{tabular}
    \end{adjustbox}
    \label{tab:full_ab}
\end{table}

\section{Conclusion and discussion}
\label{sec:conclusion}
This paper identifies a key bottleneck in observation-predictive RL, i.e., unbalanced observation reconstruction losses across dimensions. To address this bottleneck, we propose SARON, a novel observation normalization method for low-dimensional settings. We further introduce NASDAQ, a unified framework for observation-predictive RL applicable across diverse domains. Extensive experiments demonstrate that NASDAQ (with SARON) achieves competitive or superior performance compared with state-of-the-art self-predictive and model-based approaches, while incurring significantly less computational overhead. We discuss broader impacts in Appendix~\ref{app:impacts}.

\paragraph{Generality of NASDAQ} While many RL approaches showcase their generality by employing a unified network architecture with a single set of hyperparameters for tasks across domains \citep{fujimoto2025towards, hafner2023mastering, hansen2023td}, NASDAQ characterizes generality from an alternative perspective. Its generality lies in its principled compatibility with a wide range of existing value-based RL methods, requiring only modest modifications (e.g., value targets). 
We believe that this form of generality, centered on extensibility, has the potential to advance research across diverse RL domains.

\paragraph{Limitations and future directions} Although we provide a simple and efficient method for selecting benchmark-level auxiliary loss weights in Appendix~\ref{app:hyper_params}, NASDAQ falls short of achieving a universal configuration of these weights, mainly due to the heterogeneity of observation spaces across benchmarks. Besides, the two auxiliary loss weights are fixed during training. An important direction for future research is to design an approach that can adaptively adjust the weights according to observation spaces, training metrics, and possibly other indicators. Another limitation is that SARON only applies to low-dimensional settings. Given its promising performance, future work could focus on designing normalization methods for high-dimensional observations (e.g., images).

{
% \small
\bibliographystyle{unsrtnat}  % 或 unsrtnat, abbrvnat
\bibliography{refs}     % 不加 .bib 扩展名
}
%%%%%%%%%%%%%%%%%%%%%%%%%%%%%%%%%%%%%%%%%%%%%%%%%%%%%%%%%%%%
\appendix

% ====== 可选：总标题 ======
% \section*{Appendix}
% \addcontentsline{toc}{section}{Appendix}
\begin{minipage}{\textwidth}
\section{Pseudocode}
\label{app:algo}

\begin{algorithm}[H]
\caption{RL with SARON}
\label{alg:saron}
\begin{algorithmic}[1]
\State \textbf{Input:} environment $\mathrm{env}$
\State Initialize policy $\pi$ and value function $V$
\State Initialize running means and second moments $(m, s)$
\State Initialize replay buffer $\mathcal{D}$, observation buffer $\mathcal{O}$, and return queue $\mathcal{Q}$ with capacity $Q$
\State $o \gets env.\mathrm{reset}()$, $R \gets 0$

\For{$t = 1$ to $T$}
    \If{t > {warmup steps}}
    \State Obtain normalized observation $\tilde{o}$ based on $o$ and $(m, s)$ using Equation~\ref{eq:2} and~\ref{eq:3}
    \State Select action $a \sim \pi(\cdot \mid \tilde{o})$ \Comment{Inference with the normalized observation}
    \Else
    \State Sample random action $a$
    \EndIf
    % \Statex \Comment{Environment interaction and data collection}
    \State Add raw observation $o$ to $\mathcal{O}$
    \State $o^{\text{next}}, r, d \gets env.\mathrm{step}(a)$
    \State $\mathcal{D} \gets \mathcal{D}\cup \{(o,a,r,o^{\text {next}},d) \}$
    \State $R \gets R + r$ \Comment{Record episodic returns}
    \State $o \gets o^{\text {next}}$
    \If{episode ends}
        \If{$|\mathcal{Q}| = Q$ and $R$ is a low-return outlier identified by the IQR method}
            \State Skip EMA updates on $(m, s)$
        \Else
            \State Compute current episode statistics $(\mu, \nu)$ from $\mathcal{O}$
            \State Update $(m, s)$ by $(\mu, \nu)$ using Equation~\ref{eq:1}
        \EndIf
        \State Append $R$ to $\mathcal{Q}$ \Comment{Maintain recent returns for IQR-based outlier detection}
        \State Clear $\mathcal{O}$  \Comment{Clear observation buffer for new episode}
        \State $o \gets env.\mathrm{reset}()$, $R \gets 0$
    \EndIf
    \If{t > {warmup steps}}
    \State Sample mini-batch $B$ from $\mathcal{D}$
    \State Normalize the observations in $B$ based on $(m, s)$ using Equation~\ref{eq:2} and ~\ref{eq:3}
    \State Train $\pi$ and $V$ with the processed $B$ \Comment{Training with normalized observations}
    \EndIf
\EndFor
\end{algorithmic}
\end{algorithm}

% \subsection{NASDAQ}
\label{app:algo_NASDAQ}
\begin{algorithm}[H]
\caption{Updates of NASDAQ}
\label{alg:nasdaq}
\begin{algorithmic}[1]
\State \textbf{Input:} online networks $V_{1}$, $V_{2}$, ${\pi}$
\State \textbf{Input:} target networks $V'_{1}$, $V'_{2}$, ${\pi}'$
\State \textbf{Input:} replay buffer $\mathcal{D}$, current training step $t$, multi-step $n$ for \textit{n}-step returns
\State Sample a minibatch of $({o}_t, a_t, r_{t:t+n-1}, {o}_{t+1}, {o}_{t+n}) \sim \mathcal{D}$
\State Obtain $(\tilde{o}_t, a_t, r_{t:t+n-1}, \tilde{o}_{t+1}, \tilde{o}_{t+n})$ by observation normalization
% \State Compute the combined loss $\mathcal{L}$ via Equation~\ref{eq:11}
\State Optimize $V_{1}$ and $V_{2}$ against the combined loss $\mathcal{L}$ computed via Equation~\ref{eq:11}

\If{t $\%$ {policy update frequency} $=0$}
\State Optimize ${\pi}$ against the loss $\mathcal{L}_\text{Policy}$ computed via Equation~\ref{eq:8}
\EndIf

\State $t \gets t+1$

\If{t $\%$ {target update frequency} $=0$}
\State Synchronize target networks with online networks
\EndIf

\end{algorithmic}
\end{algorithm}
\end{minipage}

% ====== Appendix A ======
\section{Experimental details}
\label{app:exp_details}
\subsection{Benchmarks}
\label{app:benchmark}
\paragraph{Evaluation protocol} All experiments are run for 5 seeds. For each seed, the evaluation score is computed as the mean over 10 episodes, measured every 5k time steps. For Atari100k, however, the final evaluation (at the end of training) follows the protocol of SPR \citep{schwarzer2020data}, where the score is averaged over 100 episodes.

\paragraph{Gym} This benchmark consists of five common locomotion tasks defined by OpenAI Gym \citep{towers2024gymnasium} in the MuJoCo simulator \citep{todorov2012mujoco}, with continuous actions and low-dimensional observations. We use the \texttt{-v5} version, which is quite different from the previous versions. See the version history at \url{https://gymnasium.farama.org/environments/mujoco/}. Agents are trained for 1M time steps. Similar to MR.Q \citep{fujimoto2025towards}, we use the scores of a \textit{deep} variant of TD3 \citep{fujimoto2018addressing} to normalize the scores of other methods when aggregating results:
\begin{equation}
\label{eq:13}
\begin{aligned}
\operatorname{Deep-TD3-Normalized}(x) = \frac{x-\text{random score}}{{\text{Deep-TD3 score} - \text{random score}}}.
\end{aligned}
\end{equation}
Deep-TD3 is constructed by increasing the network size of the original TD3 to match the model capacity of other TD3-based methods on this benchmark. Specifically, we increase the depth of the value networks from three to five layers, with spectral normalization \citep{bjorck2021towards, gogianu2021spectral, miyato2018spectral} applied to the 2nd through 4th layers to stabilize training. The hidden units of each layer are 450.

\begin{table}[htbp]
\centering
\caption{The scores of Deep-TD3 and a random policy on the Gym benchmark.}
\label{tab:deepTD3}
\begin{tabular}{lcc}
\toprule
\textbf{Environments} & \textbf{Random} & \textbf{Deep-TD3} \\
\midrule

\multirow{1}{*}{Ant-v5} 
    & -0.6 & 3908 \\

\multirow{1}{*}{Humanoid-v5} 
    & 92.2 & 2897 \\

\multirow{1}{*}{HalfCheetah-v5} 
    & -265.0 &  13538 \\

\multirow{1}{*}{Hopper-v5} 
    & 25.1 &  3009 \\

\multirow{1}{*}{Walker2d-v5} 
    & 5.1 &  4831 \\
\bottomrule
\end{tabular}
\end{table}

\paragraph{DMC (proprioceptive)} The DeepMind Control suite (DMC) \citep{tassa2018deepmind} is a collection of continuous control tasks built on the MuJoCo simulator \citep{todorov2012mujoco}. These tasks use low-dimensional proprioceptive data as observations. The maximum total reward for each episode is 1000, making it easy to aggregate results. We report results on 28 tasks evaluated in MR.Q \citep{fujimoto2025towards} and DreamerV3 \citep{hafner2023mastering}. Agents are trained for 500k time steps, equivalent to 1M frames in the original environment due to an action repeat of 2.

\paragraph{DMC (visual)} Visual DMC includes the same 28 tasks as the proprioceptive benchmark but uses image-based observations instead. Consistent with the proprioceptive setting, agents are trained for 500k time steps. The input observation is composed of the previous 3 frames, which are resized to $84 \times 84$ pixels in RGB format \citep{fujimoto2025towards}.

\paragraph{Atari100k} This data-efficiency benchmark contains 26 Atari games \citep{bellemare2013arcade} for discrete control. Agents are allowed only 100k steps of environment interaction with an action repeat of 4 (producing 400k frames), amounting to 2 hours of game time. We use the no sticky action setting, following the default configuration used by DreamerV3 in this benchmark (\url{https://github.com/danijar/dreamerv3/blob/main/dreamerv3/configs.yaml}). The observation is composed of the previous 4 frames, which are gray-scaled, resized to $84 \times 84$ pixels and set to the max between the 3rd and 4th frame \citep{fujimoto2025towards}. When aggregating scores, we normalize with human scores reported by \citep{wang2016dueling} (see Table~\ref{app:human}):

\begin{equation}
\label{eq:14}
\begin{aligned}
\operatorname{Human-Normalized}(x) = \frac{x-\text{random score}}{\text{Human score - random score}}.
\end{aligned}
\end{equation}

\begin{table}[htbp]
\centering
\caption{Human and random policy scores on Atari100k games.}
\label{tab:human}
\begin{tabular}{lcc}
\toprule
\textbf{Games} & \textbf{Random} & \textbf{Human} \\
\midrule

Alien & 227.8 & 7127.7 \\
Amidar & 5.8 & 1719.5 \\
Assault & 222.4 & 742 \\
Asterix & 210 & 8503.3 \\
BankHeist & 14.2 & 753.1 \\
BattleZone & 2360.0 & 37187.5 \\
Boxing & 0.1 & 12.1 \\
Breakout & 1.7 & 30.5 \\
ChooperCommand & 811.0 & 7387.8 \\
CrazyClimber & 10780.5 & 35829.4 \\
DemonAttack & 152.1 & 1971.0 \\
Freeway & 0.0 & 29.6 \\
Frostbite & 65.2 & 4334.7 \\
Gopher & 257.6 & 2412.5 \\
Hero & 1027.0 & 30826.4 \\
Jamesbond & 29.0 & 302.8 \\
Kangaroo & 52.0 & 3035.0 \\
Krull & 1598.0 & 2665.5 \\
KungFuMaster & 258.5 & 22736.3 \\
MsPacman & 307.3 & 6951.6 \\
Pong & -20.7 & 14.6 \\
PrivateEye & 24.9 & 69571.3 \\
Qbert & 163.9 & 13455.0 \\
RoadRunner & 11.5 & 7845.0 \\
Seaquest & 68.4 & 42054.7 \\
UpNDown & 533.4 & 11693.2 \\

\bottomrule
\end{tabular}
\label{app:human}
\end{table}

\subsection{Baselines}
\label{app:baselines}
Unless otherwise specified, baseline results are obtained by running the corresponding algorithms.

\paragraph{OFENet+TD3} The Online Feature Extractor Network (OFENet) \citep{ota2020can} was proposed to learn state-action representations through observation dynamics prediction for decoupled model-free RL algorithms. It is an observation-predictive method for continuous control tasks with low-dimensional observations. We adopt the OFENet+TD3 variant in our experiments.

\paragraph{TD7} TD7 \citep{fujimoto2023sale} is a self-predictive method built on OFENet and TD3 that achieves state-of-the-art results on the Gym benchmark.

\paragraph{SPR} Self-Predictive Representation (SPR) \citep{schwarzer2020data} is a self-predictive method for visual discrete control tasks. It builds upon Rainbow and demonstrates high data efficiency in Atari100k.

\paragraph{TD-MPC2} TD-MPC2 \citep{hansen2023td} is a model-based method for continuous-control tasks that performs local trajectory optimization (planning) in the latent space of a learned world model. While the results for DMC (proprioceptive) were obtained from MR.Q, we re-run the authors’ code to gather the results for Gym because we use the \texttt{-v5} version, which is quite different from the previous versions.

\paragraph{DreamerV3} DreamerV3 \citep{hafner2023mastering} is a strong model-based RL algorithm that masters diverse domains, where a model-free policy is optimized through the rollouts from a learned world model. The results for Gym were obtained by re-running the algorithm. The results for DMC and Atari100k were obtained from MR.Q and the original paper, respectively. Note that while DreamerV3 was originally implemented by JAX \citep{jax2018github}, we use a PyTorch \citep{paszke2019pytorch} implementation (\url{https://github.com/NM512/dreamerv3-torch/}) in the experiments of computational efficiency, for a fair comparison with other methods implemented by PyTorch.

\paragraph{MR.Q} Model-based Representations for Q-learning (MR.Q) \citep{fujimoto2025towards} is a state-of-the-art self-predictive method that achieves performance competitive with model-based RL approaches, DreamerV3 and TD-MPC2, on diverse benchmarks with much less algorithmic complexity and computational overhead. Similar to OFENet+TD3 and TD7, MR.Q builds on TD3 and learns state-action representations for decoupled model-free RL algorithms. In addition to latent dynamics prediction, MR.Q introduces two extra auxiliary tasks of reward and terminal signal prediction.

\paragraph{DrQ-v2} DrQ-v2 \citep{yarats2021mastering} is a simple yet strong model-free RL baseline based on data augmentation for visual continuous control tasks. Results were obtained from MR.Q.

\paragraph{Rainbow (enhanced)} SPR introduces a series of improvements to a model-free algorithm Rainbow \citep{hessel2018rainbow}, tailored to the characteristics of the Atari100k benchmark. We keep these design choices, except the core latent dynamics learning process, to construct an enhanced variant of Rainbow for the benchmark.

\subsection{Network architecture of NASDAQ}
\label{app:structure}
\paragraph{Low-dimensional settings} The observation encoder is simply an identity function, whereas the state-action encoder is a three-layer multilayer perceptron (MLP) with an Exponential Linear Unit (ELU) activation \citep{clevert2015fast} applied after each layer. Spectral normalization (SN) \citep{miyato2018spectral, bjorck2021towards, gogianu2021spectral} is applied to the last two layers. The short-term value predictor is a two-layer MLP, where the first layer is followed by layer normalization (LayerNorm) \citep{ba2016layer} and an ELU activation. We use a non-linear decoder instead of the linear one adopted by OFENet \citep{ota2020can} to avoid overly strong regularization on state-action representations. The decoder is a two-layer MLP with ELU activation applied after the first layer. Following TD3 \citep{fujimoto2018addressing}, the policy network is a three-layer MLP where the first two layers are followed by ReLU activations. The long-term value predictor comprises three linear layers, with SN applied to the first two layers. By default, we employ ReLU activations \citep{nair2010rectified, glorot2011deep} in the first two layers of the long-term value predictor, following TD3 \citep{fujimoto2018addressing}. However, we find that using ELU, as adopted in MR.Q \citep{fujimoto2025towards}, yields better performance on the \textit{acrobot-swingup} and \textit{pendulum-swingup} tasks from the DMC (proprioceptive) benchmark. We thus replace ReLU with ELU only for these two tasks. We conjecture that ELU performs better in these tasks due to their task-specific properties. \textit{Pendulum-swingup} is a sparse-reward task, where ReLU may suffer from dead neurons \citep{nair2010rectified, lu2019dying}, while \textit{acrobot-swingup} is a challenging task (as noted in the DeepMind Control Suite \citep{tassa2018deepmind}), potentially requiring a more expressive activation function to model its value function. ELU may alleviate both issues. A more thorough investigation of activation function choices is left for future work. We also include a PyTorch-style \citep{paszke2019pytorch} code snippet to illustrate the architecture, omitting non-critical implementation details for clarity.

\begin{lstlisting}[language=Python, title={Network Architecture in low-dimensional settings}]
import torch
import torch.nn as nn
import torch.nn.functional as F
from torch.nn.utils import spectral_norm as SN
from functools import partial


# Value network with auxiliary prediction heads
class ValueNetwork(nn.Module):
    def __init__(self, obs_dim, action_dim, hidden_dim, zsa_dim, rew_bins):
        self.obs_enc = nn.Identity()
        self.state_action_enc = nn.Sequential(
            nn.Linear(int(obs_dim + action_dim), hidden_dim),
            nn.ELU(),
            SN(nn.Linear(hidden_dim, hidden_dim)),
            nn.ELU(),
            SN(nn.Linear(hidden_dim, zsa_dim)),
            nn.ELU(),
        )

        self.long_term_val_predictor = nn.Sequential(
            SN(nn.Linear(zsa_dim, hidden_dim)),
            nn.ReLU(),
            nn.Linear(hidden_dim, hidden_dim),
            nn.ReLU(),
            nn.Linear(hidden_dim, 1)
        )

        self.short_term_val_predictor = nn.Sequential(
            nn.Linear(zsa_dim, hidden_dim),
            nn.LayerNorm(hidden_dim, elementwise_affine=False),
            nn.ELU(),
            nn.Linear(hidden_dim, rew_bins)
        )

        self.decoder = nn.Sequential(
            nn.Linear(zsa_dim, hidden_dim),
            nn.ELU(),
            nn.Linear(hidden_dim, obs_dim)
        )

    def forward(self, obs, action):
        s = self.obs_enc(obs)
        zsa = torch.cat([s, action], -1)
        zsa = self.state_action_enc(zsa)

        q_val = self.long_term_val_predictor(zsa)
        short_term_val = self.short_term_val_predictor(zsa)
        next_obs = self.decoder(zsa)
        return q_val, short_term_val, next_obs


class PolicyNetork(nn.Module):
    def __init__(self, obs_dim, action_dim, hidden_dim):
        self.policy = nn.Sequential(
            nn.Linear(obs_dim, hidden_dim),
            nn.ReLU(),
            nn.Linear(hidden_dim, hidden_dim),
            nn.ReLU(),
            nn.Linear(hidden_dim, action_dim)
        )

        if discrete_action_space:
            self.final_activ = partial(F.gumbel_softmax, tau=10)
        else:
            self.final_activ = torch.tanh

    def forward(self, s):
        pre_activ = self.policy(s)
        action = self.final_activ(pre_activ)
        return action
\end{lstlisting}

\paragraph{High-dimensional settings} We adopt the same architectural design as MR.Q \citep{fujimoto2025towards} for the observation encoder, state-action encoder, long-term value predictor, and policy network. Specifically, for the observation encoder, four convolutional (Conv) layers \citep{lecun1989backpropagation, lecun2002gradient} are used, each with 32 output channels, a kernel size of 3, strides of (2, 2, 2, 1), and ELU activations. The convolutional layers are followed by a linear layer taking in the flattened output, followed by LayerNorm and a final ELU activation. The state-action encoder is a three-layer MLP with LayerNorm followed by ELU after the first two layers. The long-term value predictor is a four-layer MLP with LayerNorm followed by ELU after the first three layers. We adopt a decoder similar to that used in Dreamer \citep{hafner2019dream}, consisting of a linear layer that projects the input vector to a feature map of size (\textit{latent\_channels}, 1, 1), followed by four transposed convolutional (TransConv) \citep{zeiler2010deconvolutional} layers with strides 2 and kernel sizes of (7, 6, 6, 6). The channel dimension decreases from \textit{latent\_channels} to 64, 64, 32, and finally \textit{out\_channels}. We additionally apply spectral normalization \citep{miyato2018spectral, bjorck2021towards, gogianu2021spectral} to the first two layers of the decoder to restrict its capacity, encouraging upstream representations to carry more predictive information. The policy network is a three-layer MLP with LayerNorm followed by ReLU activations after the first two layers. Below is the corresponding PyTorch-style code snippet, including only the components that differ from the low-dimensional setting.

\begin{lstlisting}[language=Python, title={Network Architecture in high-dimensional settings}]
# Value network with auxiliary prediction heads
class ValueNetwork(nn.Module):
    def __init__(self, in_channels, state_dim, action_dim, hidden_dim, zsa_dim, latent_channels, out_channels):
        self.obs_enc = nn.Sequential(
            nn.Conv2d(in_channels, 32, kernel_size=3, stride=2),
            nn.ELU(),
            nn.Conv2d(32, 32, kernel_size=3, stride=2),
            nn.ELU(),
            nn.Conv2d(32, 32, kernel_size=3, stride=2),
            nn.ELU(),
            nn.Conv2d(32, 32, kernel_size=3, stride=1),
            nn.ELU(), nn.Flatten(),
            nn.Linear(flatten_dim, state_dim),
            nn.LayerNorm(state_dim, elementwise_affine=False),
            nn.ELU()
        )

        self.state_action_enc = nn.Sequential(
            nn.Linear(int(state_dim + action_dim), hidden_dim),
            nn.LayerNorm(state_dim, elementwise_affine=False),
            nn.ELU(),
            nn.Linear(hidden_dim, hidden_dim),
            nn.LayerNorm(state_dim, elementwise_affine=False),
            nn.ELU(),
            nn.Linear(hidden_dim, zsa_dim),
        )

        self.long_term_val_predictor = nn.Sequential(
            nn.Linear(zsa_dim, hidden_dim),
            nn.LayerNorm(hidden_dim, elementwise_affine=False),
            nn.ELU(inplace=True),
            nn.Linear(hidden_dim, hidden_dim),
            nn.LayerNorm(hidden_dim, elementwise_affine=False),
            nn.ELU(inplace=True),
            nn.Linear(hidden_dim, hidden_dim),
            nn.LayerNorm(hidden_dim, elementwise_affine=False),
            nn.ELU(inplace=True),
            nn.Linear(hidden_dim, 1)
        )

        self.decoder = nn.Sequential(
            SN(nn.Linear(zsa_dim, latent_channels)),
            nn.Unflatten(1, (latent_channels, 1, 1)),
            SN(nn.ConvTranspose2d(latent_channels, 64, kernel_size=7, stride=2)),
            nn.ReLU(),
            nn.ConvTranspose2d(64, 64, kernel_size=6, stride=2),
            nn.ReLU(),
            nn.ConvTranspose2d(64, 32, kernel_size=6, stride=2),
            nn.ReLU(),
            nn.ConvTranspose2d(32, out_channels, kernel_size=6, stride=2)
        )


class PolicyNetork(nn.Module):
    def __init__(self, obs_dim, action_dim, hidden_dim):
        self.policy = nn.Sequential(
            nn.Linear(obs_dim, hidden_dim),
            nn.LayerNorm(hidden_dim, elementwise_affine=False),
            nn.ReLU(),
            nn.Linear(hidden_dim, hidden_dim),
            nn.LayerNorm(hidden_dim, elementwise_affine=False),
            nn.ReLU(),
            nn.Linear(hidden_dim, action_dim)
        )
\end{lstlisting}

\subsection{Implementation details}
\label{app:imple_detail}
\paragraph{Design choices alignment} To reduce potential confounding factors when comparing against MR.Q \citep{fujimoto2025towards}, we keep certain design choices of OFENet+TD3 and NASDAQ consistent with those of MR.Q for components not directly related to representation learning. Specifically, we use the LAP replay buffer \citep{fujimoto2020equivalence}, and adopt the same target network update strategy, where all target parameters are periodically synchronized with their online counterparts. The value learning loss is computed using the Huber loss \citep{huber1992robust, fujimoto2020equivalence}. All networks are trained with the AdamW \citep{diederik2014adam} optimizer. 

\paragraph{Parameter sharing} Similar to prior work \citep{kostrikov2020image, hafner2023mastering, fujimoto2025towards}, we share the parameters of the observation and state-action encoders between the two value networks of NASDAQ, i.e., $f_1=f_2$, $g_1=g_2$, in high-dimensional settings to stabilize and accelerate learning. As a result of sharing these encoders, the inputs to the auxiliary prediction heads become identical across the two networks. We therefore merge the corresponding auxiliary heads ($q_1 = q_2$, $\operatorname{Dec}_1 = \operatorname{Dec}_2$).

\paragraph{Benchmark-specific modifications} For the Atari100k benchmark, NASDAQ clips rewards to $\{-1, 0, +1\}$ based on their sign, which is a preprocessing method commonly used for Atari games \citep{mnih2015human, hessel2018rainbow, schwarzer2020data}. Following SPR \citep{schwarzer2020data}, both NASDAQ and MR.Q perform \textit{two} updates of all networks per training step in this benchmark. Following DrQ-v2 \citep{yarats2021mastering} and MR.Q \citep{fujimoto2025towards}, we apply random-shift image augmentation to pixel observations on the DMC (visual) and Atari100k benchmarks.

\paragraph{Observation prediction targets in image-based tasks} As described in Appendix~\ref{app:benchmark}, the observation in pixel-based benchmarks is composed of several previous frames. This raises the question of how to construct learning targets for next observation prediction. A straightforward approach is to use the \textit{complete} next observation as the target. However, this target contains redundant information from the current observation, which may reduce the efficiency of learning dynamics-aware representations. We thus remove the overlapping frames between the current and next observations when constructing targets. Note that since input observations undergo random-shift augmentation, we apply the same sampled spatial shift to each corresponding prediction target to preserve input-target alignment in spatial shift.

\subsection{Hyper-parameters}
\label{app:hyper_params}
Table~\ref{tab:hyperparameters} and Table~\ref{tab:hyperparameters2} summarize the default hyperparameters of NASDAQ and SARON in low- and high-dimensional settings, respectively.

\begin{figure}[htbp]
    \centering
    % % ===== 第一组标题 =====
    % \textbf{Auxiliary loss across observation dimensions}

    \begin{minipage}[t]{0.245\textwidth}
        \centering
        \textbf{\quad Gym}
        \includegraphics[width=\linewidth]{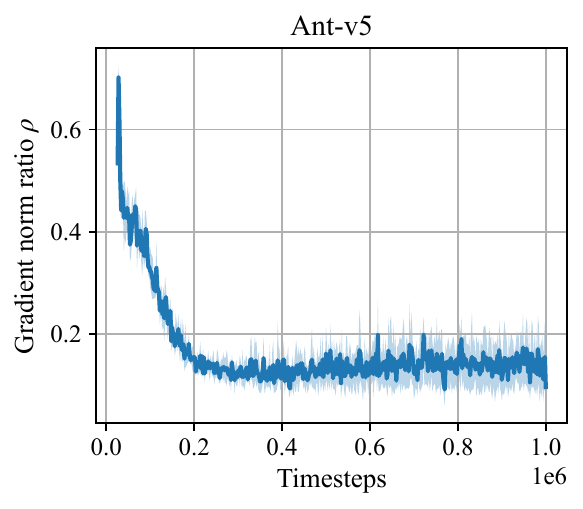}
    \end{minipage}
    \hfill
    \begin{minipage}[t]{0.245\textwidth}
        \centering
        \textbf{ DMC (proprioceptive)}
        \includegraphics[width=\linewidth]{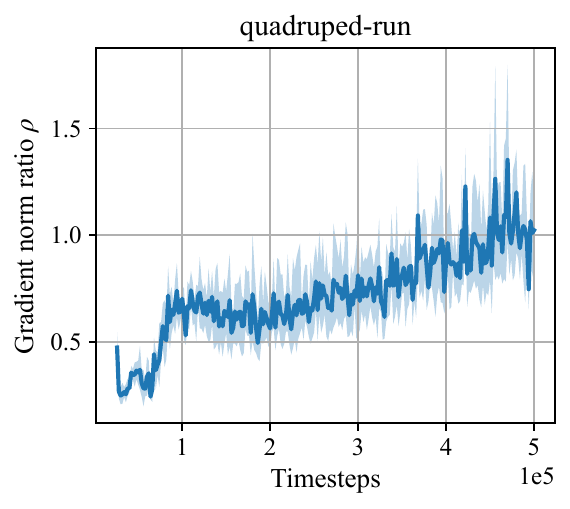}
    \end{minipage}
    \hfill
        \begin{minipage}[t]{0.245\textwidth}
        \centering
        \textbf{\quad DMC (visual)}
        \includegraphics[width=\linewidth]{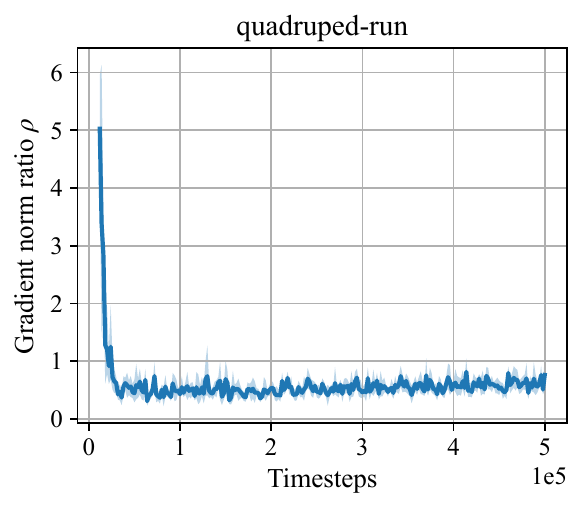}
    \end{minipage}
    \hfill
    \begin{minipage}[t]{0.245\textwidth}
        \centering
        \textbf{\quad Atari100k}
        \includegraphics[width=\linewidth]{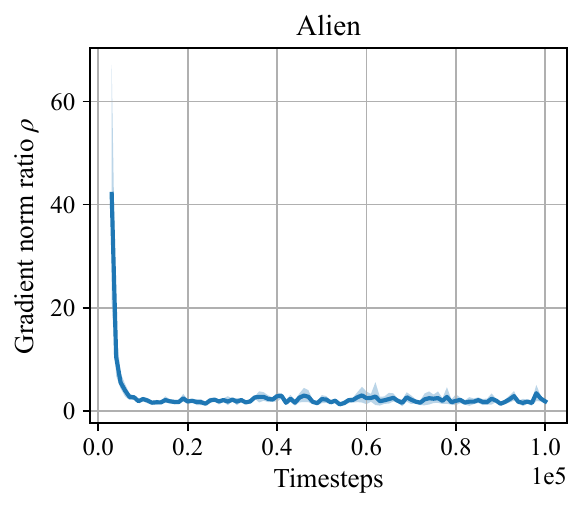}
    \end{minipage}

    % % ===== 第二组标题 =====
    % \textbf{Standard deviation across observation dimensions}

    \begin{minipage}[t]{0.245\textwidth}
        \centering
        \includegraphics[width=\linewidth]{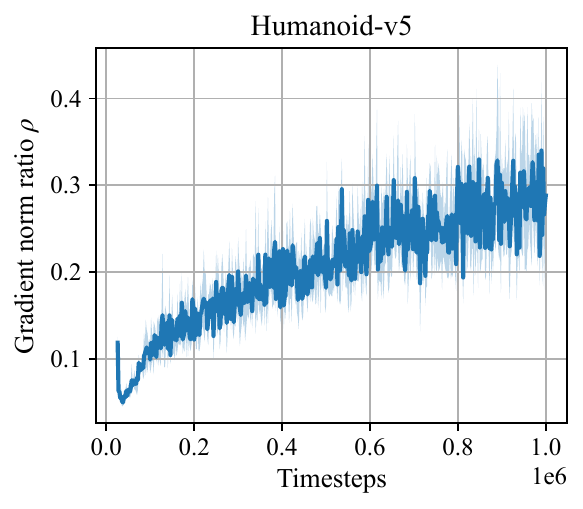}
    \end{minipage}
    \hfill
    \begin{minipage}[t]{0.245\textwidth}
        \centering
        \includegraphics[width=\linewidth]{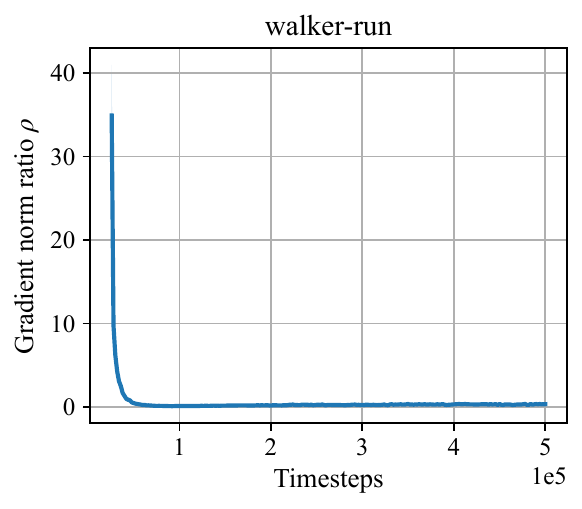}
    \end{minipage}
    \hfill
    \begin{minipage}[t]{0.245\textwidth}
        \centering
        \includegraphics[width=\linewidth]{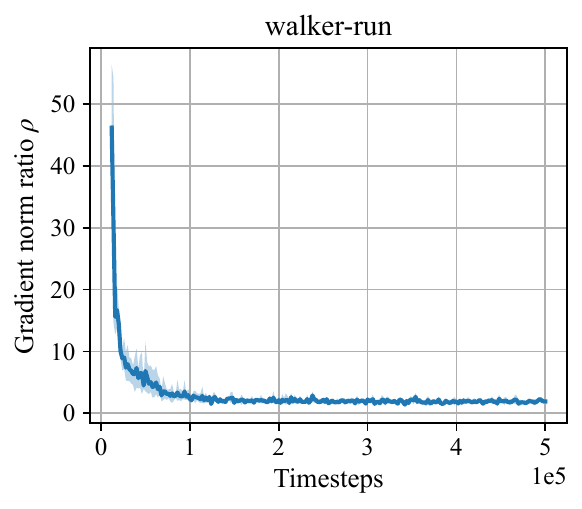}
    \end{minipage}
    \hfill
    \begin{minipage}[t]{0.245\textwidth}
        \centering
        \includegraphics[width=\linewidth]{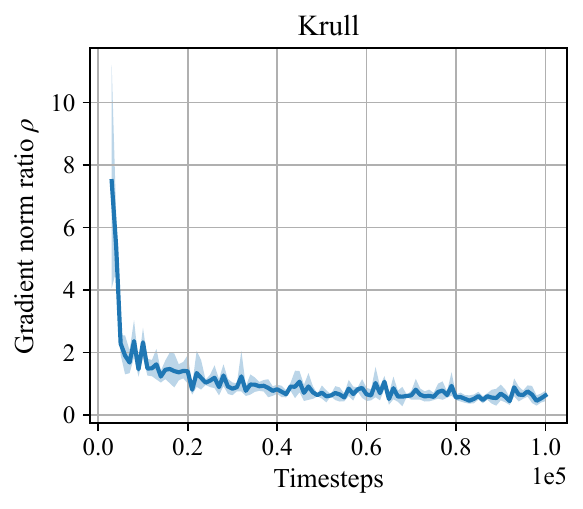}
    \end{minipage}

    \caption{Gradient norm ratio $\rho$ during training. Except for initial instability during early training, $\rho$ predominantly lies within the interval $[0.1, 2]$.}
    \label{fig:ratios}
\end{figure}

\paragraph{Hyperparameter selection for NASDAQ} We describe how the auxiliary loss weights, $\lambda_\text{Rec}$ and $ \lambda_\text{n-step}$, are determined. The two weights correspond to the two auxiliary tasks of observation prediction and short-term value prediction, respectively, as introduced in Section~\ref{subsec:NASDAQ}. For $\lambda_\text{Rec}$, we determine its value based on a training metric defined as the ratio between the gradient norm of the auxiliary loss $\mathcal{L}_\text{Rec}$ and that of the long-term value learning objective $\mathcal{L}_\text{Value}$, computed with respect to the shared observation encoder $f_{\theta_s}$ and state-action encoder $g_{\theta_{sa}}$:
\begin{equation}
\label{eq:12}
\begin{aligned}
\rho = \frac{\|\nabla_{\Theta} \mathcal{L}_{\text{Rec}}\|}{\|\nabla_{\Theta} \mathcal{L}_{\text{Value}}\|}, \quad \Theta =\{ \theta_s, \theta_{sa} \}.
\end{aligned}
\end{equation}
This gradient norm ratio provides a measure of the relative contributions of the auxiliary and main objectives to the shared parameters. If $\rho$ is too large, the auxiliary objective may dominate the gradient updates and hinder long-term value learning, and vice versa. To ensure that both objectives effectively influence the shared parameters, we determine benchmark-specific $\lambda_{\text{Rec}}$ by heuristically controlling $\rho \in [0.1,2]$ on a small subset of tasks, as shown in Figure~\ref{fig:ratios}. Note that while not all tasks satisfy the constraint, this heuristic offers a simple and effective method to select a decent $\lambda_\text{Rec}$ at the benchmark level, which serves as a starting point for further task-specific hyperparameter tuning. In our experiments, however, we use only the benchmark-level auxiliary loss weights across all tasks. The specific values of $\lambda_\text{Rec}$ and $\lambda_\text{n-step}$ are provided in Table~\ref{tab:hyperparameters} and Table~\ref{tab:hyperparameters2}.

For $\lambda_\text{n-step}$, we set its value to 1 for visual RL benchmarks without tuning. For benchmarks with low-dimensional observations, we simply set $\lambda_\text{n-step}$ to 0, indicating that short-term value prediction is not included, because we find that this auxiliary task, when $\lambda_\text{n-step}=1$, leads to slight negative effects in the non-visual setting, as shown in Table~\ref{tab:full_ab}. Although short-term value prediction benefits visual RL tasks across benchmarks, the situation is reversed in non-visual settings. We hypothesize that this phenomenon reflects a bias–variance trade-off. On the one hand, this auxiliary task provides lower-variance learning signals than those from long-term value prediction, which stabilizes the learning process. This advantage is particularly pronounced in visual RL tasks, where bootstrapped targets tend to be high-variance due to high-dimensional inputs. On the other hand, it may bias the agent toward short-term value prediction and interfere with long-term value estimation. While this negative effect may be present in both visual and non-visual settings, it becomes more evident in non-visual domains, where the variance of bootstrapped targets is relatively low, and the benefit of variance reduction is thus diminished.

\paragraph{Hyper-parameters selection for SARON} An EMA with coefficient $\beta$ can be viewed as having an effective window size of approximately $1/(1-\beta)$ \citep{fujimoto2025towards}. For the two low-dimensional benchmarks, Gym and DMC (proprioceptive), we set the effective window size to 200k, meaning that the statistics are computed using approximately 200k historical observations. Since the maximum episode lengths of Gym and DMC are 1000 and 500 (due to an action repeat of 2), respectively, the queue sizes of historical episode $|\mathcal{Q}|$, i.e., the effective window size, are thus 200 and 400, respectively.

\begin{table}[htbp]
\centering
\caption{Default hyperparameters of NASDAQ and SARON in non-visual settings.}
\label{tab:hyperparameters}
\begin{tabular}{lll}
\toprule
\textbf{Components} & \textbf{Hyperparameter} & \textbf{Value} \\
\midrule

\multirow{4}{*}{Value Learning} 
    & n-step returns & 1 \\
    & Auxiliary loss weights $(\lambda_\text{Rec}, \lambda_\text{n-step})$ & \makecell[l]{Gym: (10, 0) \\  DMC: (2, 0)} \\
    & Discount factor $\gamma$ & 0.99 \\
    % & Pre-activation loss weight $\lambda_\text{pre-activ}$ & 1e-5 \\
    % & Target policy noise clipping $c$ & (-0.5, 0.5) \\

\midrule

\multirow{5}{*}{SARON} 
    & Queue size of history episode $|\mathcal{Q}|$& \makecell[l]{Gym: 200 \\ DMC: 400} \\
    & EMA coefficient $\beta$ & \makecell[l]{Gym: $1-1 / 200$ \\ DMC: $1-1 / 400$} \\
    & Clip bounds $(-O, O)$ & (-10, 10) \\
\midrule

\multirow{2}{*}{TD3} 
    & Target policy noise $\sigma$ & $ \mathcal{N}(0,0.2^2)$ \\
    & Target policy noise clipping $c$ & (-0.5, 0.5) \\
\midrule

\multirow{2}{*}{LAP} 
    & Probability smoothing $\alpha$ & 0.4 \\
    & Minimum priority & 1 \\
\midrule
\multirow{7}{*}{Optimization} 
    & Optimizer & AdamW \\
    & Learning rate & 3e-4 \\
    & Weight decay & 1e-4 \\
    & Mini-batch size & 256 \\
    & Target update frequency & 250 \\
    & Gradient updates per training step &  \makecell[l]{value netowrk: 1 \\ policy network: 0.5} \\

\midrule

\multirow{2}{*}{Exploration} 
    & Initial random exploration time steps & 25k \\
    & Exploration noise & $ \mathcal{N}(0,0.1^2)$ \\

\midrule

\multirow{1}{*}{Observation Encoder} 
    & Structure & identity function \\
\midrule

\multirow{4}{*}{State-Action Encoder} 
    & Hidden dim & 450 \\
    & $z_{sa}$ dim & 450 \\
    & Activation function & ELU \\
    & Gradient clip norm & 20 \\
\midrule

\multirow{3}{*}{Long-term value predictor} 
    & Hidden dim & 450 \\
    & Activation function & ReLU \\
    & Gradient clip norm & 20 \\
\midrule

\multirow{4}{*}{Short-term value predictor} 
    & Hidden dim & 450 \\
    & Activation function & ELU \\
    & Reward bins & 65 \\
    & Gradient clip norm & 20 \\
\midrule

\multirow{4}{*}{Decoder} 
    & Structure & MLP \\
    & Hidden dim & 450 \\
    & Activation function & ELU \\
    & Gradient clip norm & 20 \\
\midrule

\multirow{3}{*}{Policy Network} 
    & Hidden dim & 450 \\
    & Activation function & ReLU \\
    & Gumbel-Softmax $\tau$ & 10 \\
\bottomrule
\end{tabular}
\end{table}

\begin{table}[htbp]
\centering
\caption{Default hyperparameters of NASDAQ in visual settings. For brevity, we only list hyperparameters that differ from the non-visual settings. SARON is not used in these settings.}
\label{tab:hyperparameters2}
\begin{adjustbox}{max width=\textwidth}
\begin{tabular}{lll}
\toprule
\textbf{Components} & \textbf{Hyperparameter} & \textbf{Value} \\
\midrule

\multirow{3}{*}{Value Learning} 
    & n-step returns & 3 \\
    & Auxiliary loss weights $(\lambda_\text{Rec}, \lambda_\text{n-step})$ & \makecell[l]{DMC: (0.1, 1) \\  Atari100k: (1, 1)} \\

\midrule

\multirow{1}{*}{TD3} 
    & Target policy noise clipping $c$ & (-0.3, 0.3) \\
\midrule

\multirow{1}{*}{Optimization} 
    & Gradient updates per training step &  \makecell[l]{value netowrk: 1 (DMC), 2 (Atari100k) \\ policy network: 1 (DMC), 2 (Atari100k)} \\
\midrule

\multirow{3}{*}{Exploration} 
    & Initial random exploration time steps $\alpha$ & \makecell[l]{DMC: 10k \\ Atari100k: 2k} \\
    & Exploration noise & \makecell[l]{DMC: $\mathcal{N}(0,0.1^2)$ \\ Atari100k: $\mathcal{N}(0,0.2^2)$}  \\
\midrule

\multirow{4}{*}{Observation Encoder} 
    & Structure & Conv + MLP \\
    & state dim & 512 \\
    & Activation function & ELU \\
    & Gradient clip norm & 20 \\
\midrule

\multirow{2}{*}{State-Action Encoder} 
    & Hidden dim & 580 \\
    & $z_{sa}$ dim & 512 \\
\midrule

\multirow{2}{*}{Long-term value predictor} 
    & Hidden dim & 512 \\
    & Activation function & ELU \\
\midrule

\multirow{1}{*}{Short-term value predictor} 
    & Hidden dim & 512 \\
\midrule

\multirow{5}{*}{Decoder} 
    & Structure & MLP + TransConv \\
    & Latent channels & 128 \\
    & Output channels & \makecell[l]{DMC: 3 \\ Atari100k: 1} \\
    & Activation function & ReLU \\
\midrule

\multirow{1}{*}{Policy Network} 
    & Hidden dim & 512 \\
\bottomrule
\end{tabular}
\end{adjustbox}
\end{table}

\subsection{Parameter control protocol}
\label{app:para_control}
We divide the parameters of the TD3-based methods (NASDAQ, MR.Q, OFENet+TD3, and TD7) into two categories: (1) RL components and (2) method-specific components. We define RL components as those involved in the RL training and inference processes, while method-specific components refer to the parameters used solely for representation learning. For a fair comparison, we keep the number of parameters in the RL components comparable across methods. Parameters used only for representation learning are not restricted, as they are inherently determined by the characteristics of each method. Note that we exclude the target networks when counting the number of parameters, since they are maintained as historical backups of the online counterparts and do not contribute additional model capacity \citep{hessel2018rainbow, fujimoto2018addressing, fujimoto2023sale, fujimoto2025towards}. Specifically, we control the number of parameters across these methods by adjusting their hidden sizes. Tables~\ref{tab:params} and~\ref{tab:params2} summarize the numbers of parameters across TD3-based methods in low- and high-dimensional settings, respectively. Table~\ref{tab:hidden_list} lists the specific hidden sizes for MR.Q, OFENet+TD3, and TD7. The hidden sizes of NASDAQ are provided in Tables~\ref{tab:hyperparameters} and~\ref{tab:hyperparameters2}.

\begin{table}[htbp]
    \caption{
    Parameter counts (in millions) for RL components and all components of TD3-based methods on low-dimensional benchmarks.
    }
    \centering

    \begin{adjustbox}{max width=\textwidth}
    \begin{tabular}{lcccc c cccc}
        \toprule
        \multirow{2}{*}{\textbf{Domains}}  & \multicolumn{4}{c}{\textbf{RL Components}} & & \multicolumn{4}{c}{\textbf{All the Components}} \\
        \cmidrule(lr){2-5} \cmidrule(lr){7-10}
        & NASDAQ & MR.Q & OFENet+TD3 & TD7 & & NASDAQ & MR.Q & OFENet+TD3 & TD7 \\
        \midrule
        \multicolumn{10}{l}{\textbf{Gym}} \\
        \midrule
        Ant
        & 1.88 & 1.92 & 1.96 & 1.90 &
        & 2.31 & 2.08 & 1.98 & 1.90 \\
        Humanoid
        & 2.32 & 2.41 & 2.16 & 2.35 &
        & 3.04 & 2.60 & 2.50 & 2.35 \\
        HalfCheetah
        & 1.86 & 1.91 & 1.93 & 1.89 &
        & 2.28 & 2.08 & 1.94 & 1.89 \\
        Hopper
        & 1.85 & 1.91 & 1.91 & 1.87 &
        & 2.26 & 2.08 & 1.92 & 1.87 \\
        Walker2d
        & 1.86 & 1.91 & 1.93 & 1.89 &
        & 2.28 & 2.08 & 1.94 & 1.89 \\
        \midrule
        \multicolumn{10}{l}{\textbf{DMC (proprioceptive)}} \\
        \midrule
        acrobot
        & 1.84 & 1.86 & 1.90 & 1.86 &
        & 2.25 & 2.02 & 1.90 & 1.86 \\
        \text{ball\_in\_cup}
        & 1.84 & 1.86 & 1.90 & 1.87 &
        & 2.26 & 2.02 & 1.91 & 1.87 \\    
        cartpole
        & 1.84 & 1.86 & 1.89 & 1.86 &
        & 2.25 & 2.02 & 1.90 & 1.86 \\          
        cheetah
        & 1.86 & 1.91 & 1.93 & 1.89 &
        & 2.28 & 2.08 & 1.94 & 1.89 \\
        dog
        & 2.18 & 2.28 & 2.19 & 2.21 &
        & 2.79 & 2.47 & 2.41 & 2.21 \\  
        finger
        & 1.84 & 1.86 & 1.91 & 1.87 &
        & 2.26 & 2.02 & 1.91 & 1.87 \\ 
        fish
        & 1.87 & 1.92 & 1.94 & 1.89 &
        & 2.30 & 2.08 & 1.96 & 1.89 \\ 
        hopper
        & 1.85 & 1.91 & 1.92 & 1.88 &
        & 2.27 & 2.08 & 1.93 & 1.88 \\  
        humanoid
        & 1.95 & 1.99 & 1.92 & 1.98 &
        & 2.41 & 2.16 & 1.93 & 1.98 \\  
        pendulum
        & 1.83 & 1.92 & 1.89 & 1.86 &
        & 2.24 & 2.16 & 1.89 & 1.86 \\  
        quadruped
        & 1.95 & 1.99 & 2.07 & 1.98 &
        & 2.43 & 2.16 & 2.15 & 1.98 \\  
        reacher
        & 1.84 & 1.86 & 1.90 & 1.87 &
        & 2.25 & 2.02 & 1.90 & 1.87 \\  
        walker  
        & 1.87 & 1.92 & 1.95 & 1.90 &
        & 2.30 & 2.08 & 1.97 & 1.90 \\  
        \bottomrule
    \end{tabular}
    \end{adjustbox}

\label{tab:params}
\end{table}

\begin{table}[htbp]
    \caption{
    Parameter counts (in millions) for RL components and all components of TD3-based methods on high-dimensional benchmarks.
    }
    \centering

    \begin{adjustbox}{max width=\textwidth}
    \begin{tabular}{lcc c cc}
        \toprule
        \multirow{2}{*}{\textbf{Domains}}  & \multicolumn{2}{c}{\textbf{RL Components}} & & \multicolumn{2}{c}{\textbf{All the Components}} \\
        \cmidrule(lr){2-3} \cmidrule(lr){5-6}
        & NASDAQ & MR.Q & & NASDAQ & MR.Q  \\
        \midrule
        \multicolumn{6}{l}{\textbf{DMC (proprioceptive)}} \\
        \midrule
        acrobot
        & 3.87 & 3.86 & 
        & 4.86 & 4.15  \\
        \text{ball\_in\_cup}
        & 3.87 & 3.86 & 
        & 4.86 & 4.15 \\    
        cartpole
        & 3.87 & 3.86 & 
        & 4.86 & 4.15  \\         
        cheetah
        & 3.87 & 3.86 & 
        & 4.86 & 4.16  \\ 
        dog
        & 3.91 & 3.88 & 
        & 4.90 & 4.18  \\ 
        finger
        & 3.87 & 3.86 & 
        & 4.86 & 4.15  \\ 
        fish
        & 3.87 & 3.86 & 
        & 4.86 & 4.16  \\ 
        hopper
        & 3.87 & 3.86 & 
        & 4.86 & 4.15  \\ 
        humanoid
        & 3.89 & 3.87 & 
        & 4.88 & 4.17  \\ 
        pendulum
        & 3.87 & 3.86 & 
        & 4.86 & 4.15  \\ 
        quadruped
        & 3.88 & 3.86 & 
        & 4.87 & 4.16  \\ 
        reacher
        & 3.87 & 3.86 & 
        & 4.86 & 4.15 \\ 
        walker  
        & 3.87 & 3.86 & 
        & 4.86 & 4.16 \\ 
        % acrobot
        % & 3.91 & 3.86 & 
        % & 4.90 & 4.15  \\
        % \text{ball\_in\_cup}
        % & 3.91 & 3.86 & 
        % & 4.90 & 4.15 \\    
        % cartpole
        % & 3.91 & 3.86 & 
        % & 4.90 & 4.15  \\         
        % cheetah
        % & 3.92 & 3.86 & 
        % & 4.91 & 4.16  \\ 
        % dog
        % & 3.95 & 3.88 & 
        % & 4.94 & 4.18  \\ 
        % finger
        % & 3.91 & 3.86 & 
        % & 4.90 & 4.15  \\ 
        % fish
        % & 3.92 & 3.86 & 
        % & 4.91 & 4.16  \\ 
        % hopper
        % & 3.92 & 3.86 & 
        % & 4.90 & 4.15  \\ 
        % humanoid
        % & 3.94 & 3.87 & 
        % & 4.92 & 4.17  \\ 
        % pendulum
        % & 3.91 & 3.86 & 
        % & 4.90 & 4.15  \\ 
        % quadruped
        % & 3.93 & 3.86 & 
        % & 4.91 & 4.16  \\ 
        % reacher
        % & 3.91 & 3.86 & 
        % & 4.90 & 4.15 \\ 
        % walker  
        % & 3.92 & 3.86 & 
        % & 4.91 & 4.16 \\ 
        \midrule
        \multicolumn{6}{l}{\textbf{Atari100k}} \\
        \midrule
        Alien
        & 3.89 & 3.87 & 
        & 4.87 & 4.16  \\
        Amidar
        & 3.88 & 3.86 & 
        & 4.86 & 4.16  \\         
        Assault
        & 3.87 & 3.86 & 
        & 4.86 & 4.16  \\ 
        Asterix
        & 3.88 & 3.86 & 
        & 4.86 & 4.16  \\ 
        BankHeist
        & 3.89 & 3.87 & 
        & 4.87 & 4.16  \\ 
        BattleZone
        & 3.89 & 3.87 & 
        & 4.87 & 4.16  \\ 
        Boxing
        & 3.89 & 3.87 & 
        & 4.87 & 4.16  \\ 
        Breakout
        & 3.87 & 3.86 & 
        & 4.86 & 4.15  \\ 
        ChopperCommand
        & 3.89 & 3.87 & 
        & 4.87 & 4.16  \\ 
        CrazyClimber
        & 3.88 & 3.86 & 
        & 4.86 & 4.16  \\ 
        DemonAttack
        & 3.87 & 3.86 & 
        & 4.86 & 4.15 \\ 
        Freeway  
        & 3.87 & 3.86 & 
        & 4.86 & 4.15 \\ 
        Frostbite  
        & 3.89 & 3.87 & 
        & 4.87 & 4.16 \\ 
        Gopher  
        & 3.88 & 3.86 & 
        & 4.86 & 4.16 \\ 
        Hero  
        & 3.89 & 3.87 & 
        & 4.87 & 4.16 \\ 
        Jamesbond  
        & 3.89 & 3.87 & 
        & 4.87 & 4.16 \\ 
        Kangaroo  
        & 3.89 & 3.87 & 
        & 4.87 & 4.16 \\ 
        Krull  
        & 3.89 & 3.87 & 
        & 4.87 & 4.16 \\ 
        KungFuMaster  
        & 3.88 & 3.86 & 
        & 4.87 & 4.16 \\ 
        MsPacman  
        & 3.88 & 3.86 & 
        & 4.86 & 4.16 \\ 
        Pong  
        & 3.87 & 3.86 & 
        & 4.86 & 4.15 \\ 
        PrivateEye  
        & 3.89 & 3.87 & 
        & 4.87 & 4.16 \\ 
        Qbert  
        & 3.87 & 3.86 & 
        & 4.86 & 4.15 \\ 
        RoadRunner  
        & 3.89 & 3.87 & 
        & 4.87 & 4.16 \\ 
        Seaquest  
        & 3.89 & 3.87 & 
        & 4.87 & 4.16 \\ 
        UpNDown  
        & 3.87 & 3.86 & 
        & 4.86 & 4.15 \\ 
        \bottomrule
    \end{tabular}
    \end{adjustbox}
\label{tab:params2}
\end{table}

\begin{table}
    \caption{
    Hidden size configurations of TD3-based methods across domains. 
    }
    \centering
    \begin{tabular}{lccc}
        \toprule
        \textbf{Domains} & \textbf{MR.Q} & \textbf{OFENet+TD3} & \textbf{TD7} \\
        \midrule
        \multicolumn{4}{l}{\textbf{Gym}} \\
        \midrule
        Ant & 375 & 420 & 340 \\
        Humanoid & 408 & 360 & 340 \\
        HalfCheetah & 375 & 420 & 340 \\
        Hopper & 375 & 420 & 340  \\
        Walker2d & 375 & 420 & 340 \\
        \midrule
        \multicolumn{4}{l}{\textbf{DMC (proprioceptive)}} \\
        \midrule
        acrobot
        & 370 & 360 & 340  \\
        \text{ball\_in\_cup}
        & 370 & 360 & 340  \\  
        cartpole
        & 370 & 360 & 340  \\      
        cheetah
        & 375 & 360 & 340  \\
        dog
        & 400 & 420 & 340  \\
        finger
        & 370 & 360 & 340  \\
        fish
        & 375 & 360 & 340  \\
        hopper
        & 375 & 360 & 340  \\
        humanoid
        & 380 & 360 & 340  \\
        pendulum
        & 370 & 360 & 340  \\
        quadruped
        & 380 & 360 & 340  \\
        reacher
        & 370 & 360 & 340  \\
        walker  
        & 375 & 360 & 340  \\      
        \midrule
        \multicolumn{4}{l}{\textbf{DMC (visual)}} \\
        \midrule
          & 512 & N/A & N/A  \\           
        \midrule
        \multicolumn{4}{l}{\textbf{Atari100k}} \\
        \midrule
          & 512 & N/A & N/A  \\  
        \bottomrule
    \end{tabular}
    \label{tab:hidden_list}
\end{table}
% \begin{table}[htbp]
% \centering
% \caption{Parameter Summary.}
% \label{tab:deepTD3}
% \begin{tabular}{lcccc}
% \toprule
% \textbf{Domains} & \textbf{NASDAQ} & \textbf{OFENet} & \textbf{MR.Q} & \textbf{TD7} \\
% \midrule

% Ant & -0.6 & 3908 & -0.6 & 3908 \\

% Humanoid & 92.2 & 2897 & -0.6 & 3908 \\

% HalfCheetah & -265.0 &  13538 & -0.6 & 3908 \\

% Hopper & 25.1 &  3009 & -0.6 & 3908 \\

% Walker2d & 5.1 &  4831 & -0.6 & 3908 \\

% \bottomrule
% \end{tabular}
% \end{table}

% ====== Appendix B ======
\clearpage
\begin{minipage}{\textwidth}
\section{Complete results}
\label{app:complete_results}
\subsection{Gym}
\label{app:full_gym}
\begin{table}[H]
    \caption{
    Final performance on the \textbf{Gym} benchmark at 1M time steps, averaged over 5 seeds. The {\textcolor{gray}{[bracketed values]}} represent a 95\% bootstrap confidence interval. The aggregate mean, median, and interquartile mean (IQM) are computed using the Deep-TD3-normalized score (see Appendix~\ref{app:benchmark}).
    }
    \centering
    \begin{adjustbox}{max width=\textwidth}
    \begin{tabular}{lcccccc}
        \toprule
        \textbf{Tasks} & \textbf{OFENet+TD3} & \textbf{TD7} & \textbf{TD-MPC2} &
        \textbf{DreamerV3} & \textbf{MR.Q} & \textbf{NASDAQ with SARON} \\
        \midrule
        Ant-v5 & 
        8156 {\textcolor{gray}{[8047, 8280]}} & 
        7181 {\textcolor{gray}{[6270, 8106]}} & 
        1120 {\textcolor{gray}{[555, 2117]}} & 
        1457 {\textcolor{gray}{[200, 2991]}} &
        7514 {\textcolor{gray}{[6977, 7911]}} & 
        7871 {\textcolor{gray}{[7664, 8103]}} \\
        
        HalfCheetah-v5 & 
        13548 {\textcolor{gray}{[12867, 14309]}} & 
        17723 {\textcolor{gray}{[17446, 18019]}} & 
        15127 {\textcolor{gray}{[14518, 15619]}} &  
        7107 {\textcolor{gray}{[6576, 7872]}} & 
        13823 {\textcolor{gray}{[13412, 14251]}} & 
        16742 {\textcolor{gray}{[16476, 17031]}} \\

        Hopper-v5 & 
        2853 {\textcolor{gray}{[1837, 3637]}} & 
        1612 {\textcolor{gray}{[1272, 1920]}} & 
        313 {\textcolor{gray}{[222, 431]}} &  
        2773 {\textcolor{gray}{[2066, 3379]}} & 
        2699 {\textcolor{gray}{[2245, 3179]}} & 
        2856 {\textcolor{gray}{[2340, 3372]}}  \\
        
        Humanoid-v5 & 
        6063 {\textcolor{gray}{[5853, 6273]}} & 
        4400 {\textcolor{gray}{[3872, 4924]}} & 
        383 {\textcolor{gray}{[266, 556]}} & 
        2568 {\textcolor{gray}{[1771, 3249]}} & 
        7207 {\textcolor{gray}{[5950, 8147]}} & 
        9938 {\textcolor{gray}{[9577, 10251]}} \\

        Walker2d-v5 & 
        6009 {\textcolor{gray}{[5812, 6206]}} & 
        2742 {\textcolor{gray}{[1953, 3554]}} & 
        980 {\textcolor{gray}{[611, 1555]}} &  
        3021 {\textcolor{gray}{[475, 5567]}} & 
        5431 {\textcolor{gray}{[4793, 5998]}} & 
        6077 {\textcolor{gray}{[5661, 6375]}} \\
        
        \midrule
        \multicolumn{4}{l}{\textbf{Aggregate Results}} \\
        \midrule
        Mean & 1.48 & 1.16 & 0.36 & 0.66 & 1.50 & 1.79 \\
        Median & 1.24 & 1.30 & 0.20 & 0.62 & 1.12 & 1.26  \\
        IQM & 1.44 & 1.14 & 0.20 & 0.68 & 1.36 & 1.50 \\
        \bottomrule
    \end{tabular}
    \label{tab:main_result_gym}
    \end{adjustbox}
\end{table}
\begin{figure}[H]
    \centering

    \begin{minipage}[t]{0.32\textwidth}
        % \caption*{\quad \quad \ \ Ant-v5}
        \centering
        % \textbf{Gym}
        \includegraphics[width=\linewidth]{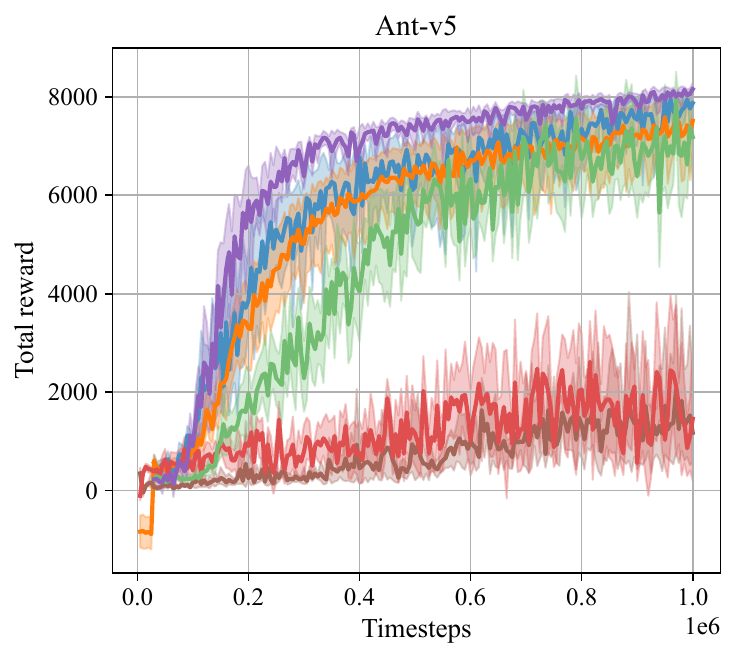}
    \end{minipage}
    \hfill
    \begin{minipage}[t]{0.32\textwidth}
        % \caption*{\quad \quad \ \ HalfCheetah-v5}
        \centering
        % \textbf{DMC (proprioceptive)}
        \includegraphics[width=\linewidth]{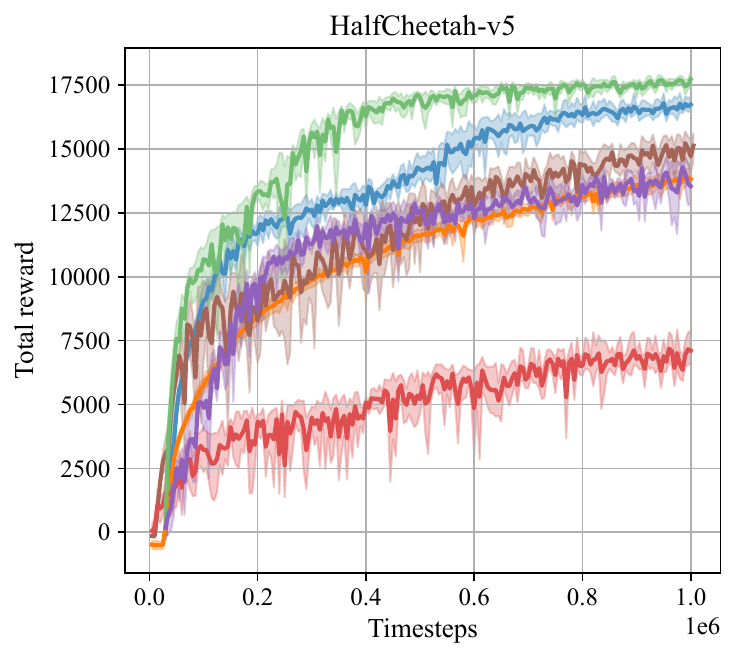}
    \end{minipage}
    \hfill
    \begin{minipage}[t]{0.32\textwidth}
        % \caption*{\quad \quad \ \ Hopper-v5}        
        \centering
        % \textbf{DMC (visual)}
        \includegraphics[width=\linewidth]{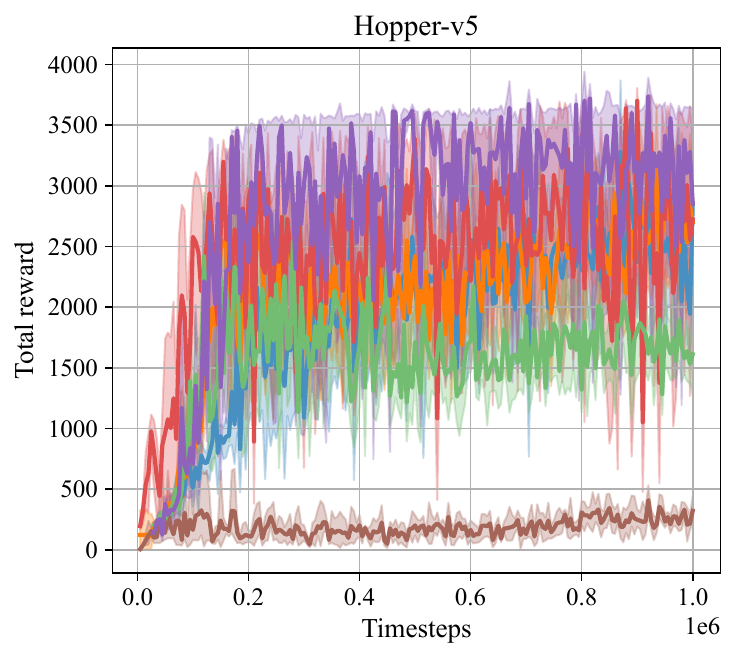}
    \end{minipage}

    \begin{minipage}{\textwidth}
        \begin{minipage}[t]{0.32\textwidth}
            % \caption*{\quad \quad \ \ Humanoid-v5}
            \centering
            \includegraphics[width=\linewidth]{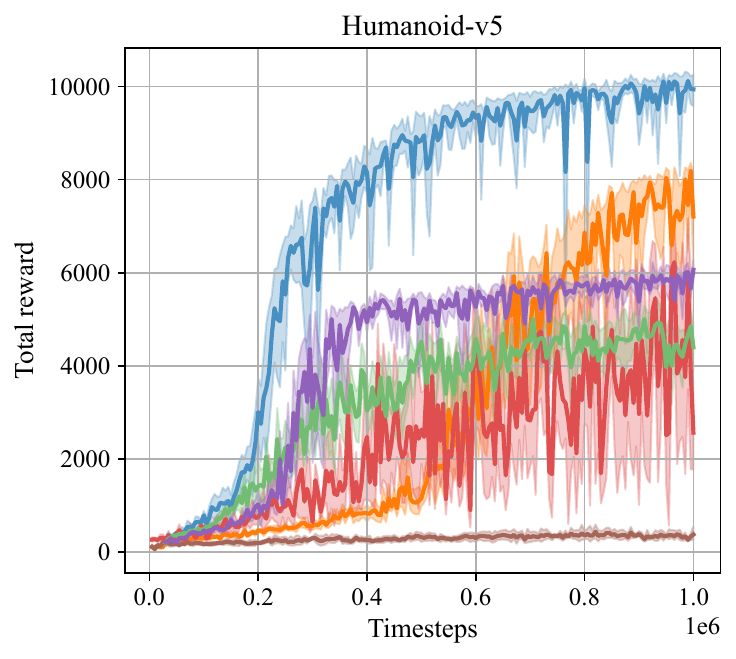}
        \end{minipage}
        % \hspace{0.01\textwidth}
        \begin{minipage}[t]{0.32\textwidth}
            % \caption*{\quad \quad \ \ Walker2d-v5}
            \centering
            \includegraphics[width=\linewidth]{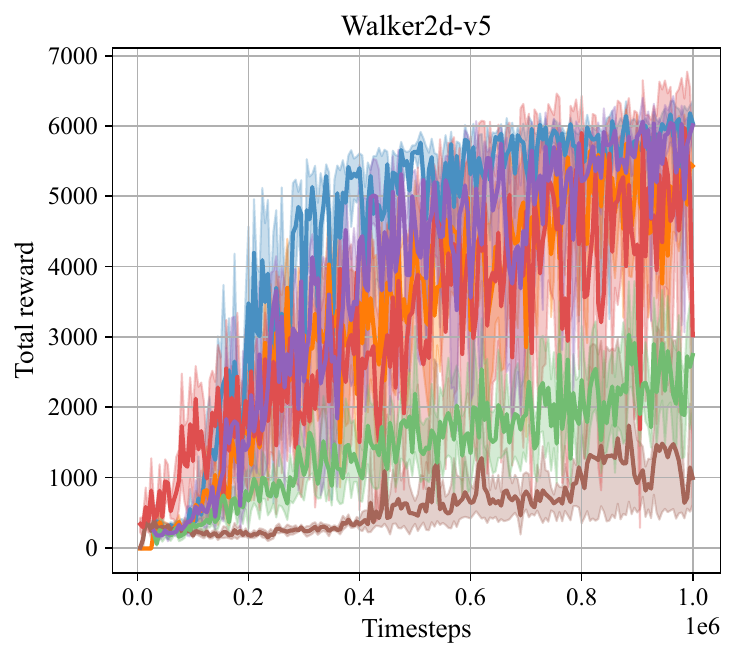}
        \end{minipage}
        % \hspace{1 em}
        \begin{minipage}[t]{0.2\textwidth}
            \centering
            % \vspace{-6.5 em}
        \includegraphics[width=\linewidth]{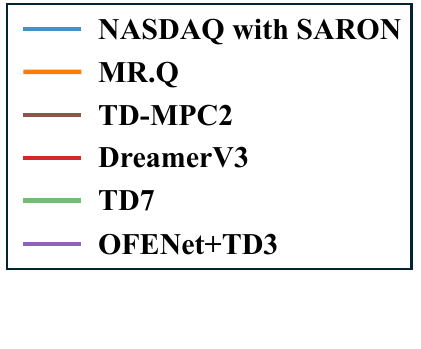}
    \end{minipage}    
    \end{minipage}
    \caption{
    Learning curves on \textbf{Gym}. Results are over 5 seeds. The shaded area captures a 95\% bootstrap confidence interval.
    }
    \label{fig:main_result_gym}
\end{figure}
\vfill
\end{minipage}

\begin{minipage}{\textwidth}
\subsection{DMC (proprioceptive)}
\label{app:full_dmc_p}
\begin{table}[H]
    \caption{
    Final performance on the \textbf{DMC (proprioceptive)} benchmark at 500k time steps (1M time steps in the original environment due to an action repeat of 2), averaged over 5 seeds. Results for TD-MPC2 and DreamerV3 are obtained from MR.Q. The {\textcolor{gray}{[bracketed values]}} represent a 95\% bootstrap confidence interval. 
    }
    \centering
    \begin{adjustbox}{max width=\textwidth}
    \begin{tabular}{lcccccc}
        \toprule
        \textbf{Tasks} & \textbf{OFENet+TD3} & \textbf{TD7} & \textbf{TD-MPC2} &
        \textbf{DreamerV3} & \textbf{MR.Q} & \textbf{NASDAQ with SARON} \\
        \midrule
        acrobot-swingup & 
        40 {\textcolor{gray}{[18, 62]}} & 
        257 {\textcolor{gray}{[228, 285]}} & 
        584 {\textcolor{gray}{[551, 615]}} & 
        230 {\textcolor{gray}{[193, 266]}} &
        234 {\textcolor{gray}{[201, 267]}} & 
        242 {\textcolor{gray}{[200, 279]}} \\
        
        ball\_in\_cup-catch & 
        981 {\textcolor{gray}{[975, 986]}} & 
        982 {\textcolor{gray}{[980, 986]}} & 
        984 {\textcolor{gray}{[982, 986]}} & 
        968 {\textcolor{gray}{[965, 973]}} & 
        982 {\textcolor{gray}{[980, 983]}} & 
        981 {\textcolor{gray}{[978, 983]}} \\
        
        cartpole-balance & 
        985 {\textcolor{gray}{[972, 997]}} & 
        996 {\textcolor{gray}{[994, 998]}} & 
        996 {\textcolor{gray}{[995, 998]}} &  
        998 {\textcolor{gray}{[997, 1000]}} & 
        998 {\textcolor{gray}{[996, 999]}} & 
        996 {\textcolor{gray}{[994, 998]}} \\
        
        cartpole-balance\_sparse & 
        678 {\textcolor{gray}{[325, 1000]}} & 
        1000 {\textcolor{gray}{[1000, 1000]}} & 
        1000 {\textcolor{gray}{[1000, 1000]}} &  
        999 {\textcolor{gray}{[999, 1000]}} & 
        966 {\textcolor{gray}{[898, 1000]}} & 
        1000 {\textcolor{gray}{[1000, 1000]}}  \\
        
        cartpole-swingup & 
        867 {\textcolor{gray}{[858, 875]}} & 
        858 {\textcolor{gray}{[844, 871]}} & 
        875 {\textcolor{gray}{[870, 880]}} &  
        736 {\textcolor{gray}{[591, 838]}} & 
        880 {\textcolor{gray}{[879, 881]}} & 
        881 {\textcolor{gray}{[880, 882]}} \\

        cartpole-swingup\_sparse & 
        168 {\textcolor{gray}{[0, 504]}} & 
        513 {\textcolor{gray}{[183, 833]}} & 
        845 {\textcolor{gray}{[839, 849]}} & 
        702 {\textcolor{gray}{[560, 792]}} &
        757 {\textcolor{gray}{[588, 843]}} & 
        827 {\textcolor{gray}{[814, 837]}} \\
        
        cheetah-run & 
        908 {\textcolor{gray}{[903, 913]}} & 
        832 {\textcolor{gray}{[670, 918]}} & 
        917 {\textcolor{gray}{[915, 920]}} & 
        699 {\textcolor{gray}{[655, 744]}} & 
        892 {\textcolor{gray}{[877, 905]}} & 
        920 {\textcolor{gray}{[918, 921]}} \\
        
        dog-run & 
        66 {\textcolor{gray}{[22, 110]}} & 
        176 {\textcolor{gray}{[146, 210]}} & 
        265 {\textcolor{gray}{[166, 342]}} &   
        4 {\textcolor{gray}{[4, 5]}} & 
        300 {\textcolor{gray}{[281, 319]}} & 
        296 {\textcolor{gray}{[272, 315]}} \\
        
        dog-stand & 
        347 {\textcolor{gray}{[268, 466]}} & 
        935 {\textcolor{gray}{[924, 945]}} & 
        506 {\textcolor{gray}{[266, 715]}} &  
        22 {\textcolor{gray}{[20, 27]}} & 
        946 {\textcolor{gray}{[935, 956]}} & 
        944 {\textcolor{gray}{[933, 955]}}  \\
        
        dog-trot & 
        98 {\textcolor{gray}{[73, 127]}} & 
        186 {\textcolor{gray}{[105, 261]}} & 
        407 {\textcolor{gray}{[265, 530]}} &  
        10 {\textcolor{gray}{[6, 17]}} & 
        593 {\textcolor{gray}{[516, 676]}} & 
        645 {\textcolor{gray}{[552, 738]}} \\

        dog-walk & 
        144 {\textcolor{gray}{[112, 176]}} & 
        687 {\textcolor{gray}{[444, 842]}} & 
        486 {\textcolor{gray}{[240, 704]}} & 
        17 {\textcolor{gray}{[15, 21]}} &
        773 {\textcolor{gray}{[724, 819]}} & 
        807 {\textcolor{gray}{[761, 856]}} \\
        
        finger-spin & 
        977 {\textcolor{gray}{[973, 982]}} & 
        794 {\textcolor{gray}{[415, 987]}} & 
        986 {\textcolor{gray}{[986, 988]}} & 
        666 {\textcolor{gray}{[577, 763]}} & 
        981 {\textcolor{gray}{[978, 983]}} & 
        979 {\textcolor{gray}{[974, 982]}} \\
        
        finger-turn\_easy & 
        771 {\textcolor{gray}{[535, 968]}} & 
        826 {\textcolor{gray}{[513, 984]}} & 
        979 {\textcolor{gray}{[975, 983]}} &  
        906 {\textcolor{gray}{[883, 927]}} & 
        968 {\textcolor{gray}{[947, 980]}} & 
        933 {\textcolor{gray}{[890, 971]}} \\
        
        finger-turn\_hard & 
        466 {\textcolor{gray}{[234, 698]}} & 
        427 {\textcolor{gray}{[81, 777]}} & 
        947 {\textcolor{gray}{[916, 977]}} &  
        864 {\textcolor{gray}{[812, 900]}} & 
        973 {\textcolor{gray}{[968, 978]}} & 
        959 {\textcolor{gray}{[936, 973]}}  \\
        
        fish-swim & 
        151 {\textcolor{gray}{[103, 190]}} & 
        223 {\textcolor{gray}{[118, 381]}} & 
        659 {\textcolor{gray}{[615, 706]}} &  
        813 {\textcolor{gray}{[808, 819]}} & 
        541 {\textcolor{gray}{[469, 609]}} & 
        803 {\textcolor{gray}{[793, 813]}} \\

        hopper-hop & 
        236 {\textcolor{gray}{[182, 290]}} & 
        247 {\textcolor{gray}{[57, 437]}} & 
        425 {\textcolor{gray}{[368, 500]}} & 
        116 {\textcolor{gray}{[66, 165]}} &
        329 {\textcolor{gray}{[302, 361]}} & 
        284 {\textcolor{gray}{[268, 300]}} \\
        
        hopper-stand & 
        930 {\textcolor{gray}{[922, 938]}} & 
        382 {\textcolor{gray}{[1, 762]}} & 
        952 {\textcolor{gray}{[944, 958]}} & 
        747 {\textcolor{gray}{[669, 806]}} & 
        948 {\textcolor{gray}{[926, 961]}} & 
        949 {\textcolor{gray}{[944, 953]}} \\
        
        humanoid-run & 
        57 {\textcolor{gray}{[13, 101]}} & 
        109 {\textcolor{gray}{[54, 142]}} & 
        181 {\textcolor{gray}{[121, 231]}} &  
        0 {\textcolor{gray}{[1, 1]}} & 
        148 {\textcolor{gray}{[128, 169]}} & 
        186 {\textcolor{gray}{[170, 204]}} \\
        
        humanoid-stand & 
        112 {\textcolor{gray}{[7, 221]}} & 
        656 {\textcolor{gray}{[543, 770]}} & 
        658 {\textcolor{gray}{[506, 745]}} &  
        5 {\textcolor{gray}{[5, 6]}} & 
        811 {\textcolor{gray}{[731, 876]}} & 
        860 {\textcolor{gray}{[797, 909]}}  \\
        
        humanoid-walk & 
        156 {\textcolor{gray}{[8, 312]}} & 
        547 {\textcolor{gray}{[449, 658]}} & 
        754 {\textcolor{gray}{[725, 791]}} &  
        1 {\textcolor{gray}{[1, 2]}} & 
        719 {\textcolor{gray}{[630, 809]}} & 
        622 {\textcolor{gray}{[606, 639]}} \\

        pendulum-swingup & 
        380 {\textcolor{gray}{[60, 700]}} & 
        211 {\textcolor{gray}{[19, 532]}} & 
        846 {\textcolor{gray}{[830, 862]}} & 
        774 {\textcolor{gray}{[740, 802]}} &
        832 {\textcolor{gray}{[792, 864]}} & 
        822 {\textcolor{gray}{[808, 838]}} \\
        
        quadruped-run & 
        852 {\textcolor{gray}{[745, 918]}} & 
        918 {\textcolor{gray}{[867, 947]}} & 
        942 {\textcolor{gray}{[938, 947]}} & 
        130 {\textcolor{gray}{[92, 169]}} & 
        944 {\textcolor{gray}{[938, 951]}} & 
        935 {\textcolor{gray}{[909, 942]}} \\
        
        quadruped-walk & 
        929 {\textcolor{gray}{[920, 937]}} & 
        948 {\textcolor{gray}{[931, 963]}} & 
        963 {\textcolor{gray}{[959, 967]}} &  
        193 {\textcolor{gray}{[137, 243]}} & 
        962 {\textcolor{gray}{[956, 968]}} & 
        960 {\textcolor{gray}{[954, 966]}} \\
        
        reacher-easy & 
        977 {\textcolor{gray}{[973, 979]}} & 
        969 {\textcolor{gray}{[931, 989]}} & 
        983 {\textcolor{gray}{[980, 986]}} &  
        966 {\textcolor{gray}{[964, 970]}} & 
        952 {\textcolor{gray}{[917, 983]}} & 
        983 {\textcolor{gray}{[982, 985]}}  \\
        
        reacher-hard & 
        956 {\textcolor{gray}{[935, 970]}} & 
        976 {\textcolor{gray}{[972, 979]}} & 
        960 {\textcolor{gray}{[936, 979]}} &  
        919 {\textcolor{gray}{[864, 955]}} & 
        953 {\textcolor{gray}{[927, 976]}} & 
        978 {\textcolor{gray}{[974, 983]}} \\

        walker-run & 
        714 {\textcolor{gray}{[668, 759]}} & 
        847 {\textcolor{gray}{[843, 852]}} & 
        854 {\textcolor{gray}{[851, 859]}} &  
        510 {\textcolor{gray}{[430, 588]}} & 
        790 {\textcolor{gray}{[772, 806]}} & 
        794 {\textcolor{gray}{[786, 803]}} \\
        
        walker-stand & 
        979 {\textcolor{gray}{[973, 984]}} & 
        991 {\textcolor{gray}{[988, 994]}} & 
        991 {\textcolor{gray}{[990, 994]}} &  
        941 {\textcolor{gray}{[934, 948]}} & 
        990 {\textcolor{gray}{[988, 992]}} & 
        987 {\textcolor{gray}{[985, 989]}}  \\
        
        walker-walk & 
        959 {\textcolor{gray}{[949, 968]}} & 
        980 {\textcolor{gray}{[978, 983]}} & 
        981 {\textcolor{gray}{[979, 984]}} &  
        898 {\textcolor{gray}{[875, 919]}} & 
        978 {\textcolor{gray}{[976, 980]}} & 
        976 {\textcolor{gray}{[975, 977]}} \\
        \midrule
        \multicolumn{4}{l}{\textbf{Aggregate Results}} \\
        \midrule
        Mean & 567 & 660 & 783 & 530 & 791 & 805 \\
        Median & 696 & 810 & 896 & 700 & 918 & 927  \\
        IQM & 600 & 726 & 868 & 577 & 886 & 898 \\
        \bottomrule
    \end{tabular}
    \label{tab:main_result_dmc1}
    \end{adjustbox}
\end{table}

\end{minipage}
\begin{figure}[H]
    \centering
    \begin{minipage}[t]{0.245\textwidth}
        % \caption*{\quad \quad acrobot-swingup}
        \centering
        \includegraphics[width=\linewidth]{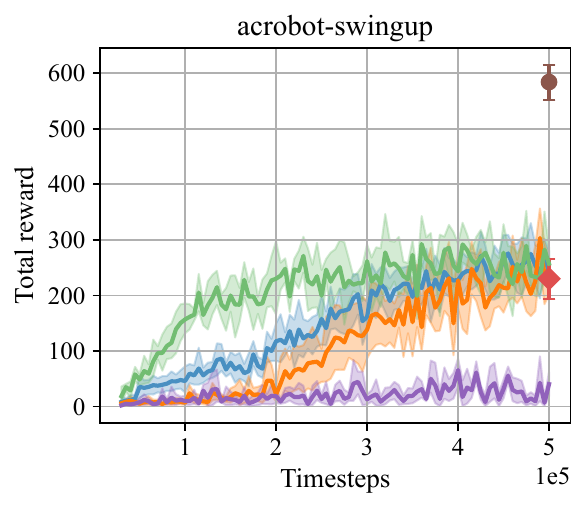}
    \end{minipage}
    \hfill
    \begin{minipage}[t]{0.245\textwidth}
        % \caption*{\quad \quad ball\_in\_cup-catch}
        \centering
        \includegraphics[width=\linewidth]{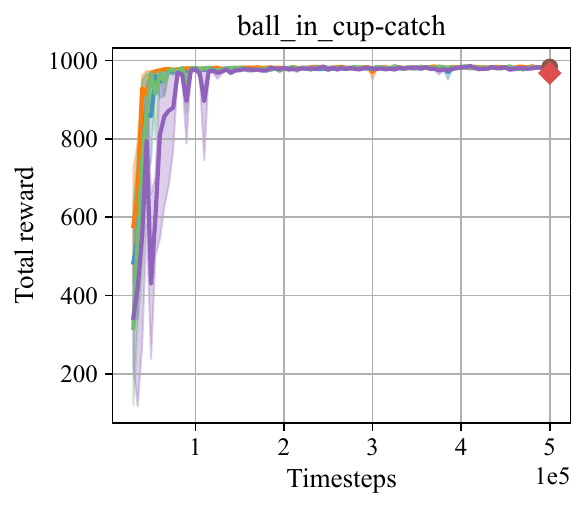}
    \end{minipage}
    \hfill
    \begin{minipage}[t]{0.245\textwidth}
        % \caption*{\quad \quad cartpole-balance}
        \centering
        \includegraphics[width=\linewidth]{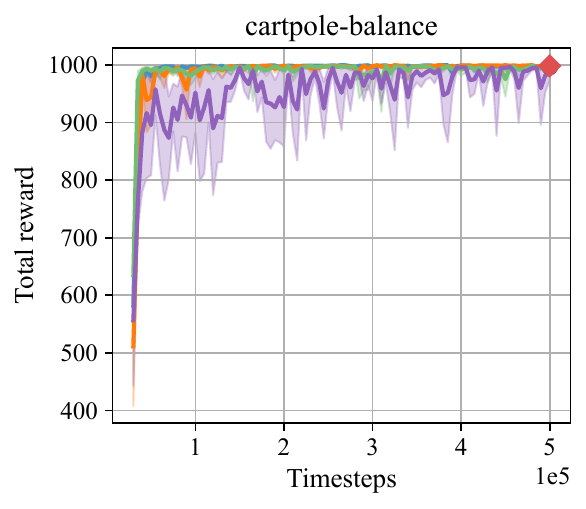}
    \end{minipage}
    \hfill
    \begin{minipage}[t]{0.245\textwidth}
        % \caption*{\quad cartpole-balance\_sparse}
        \centering
        \includegraphics[width=\linewidth]{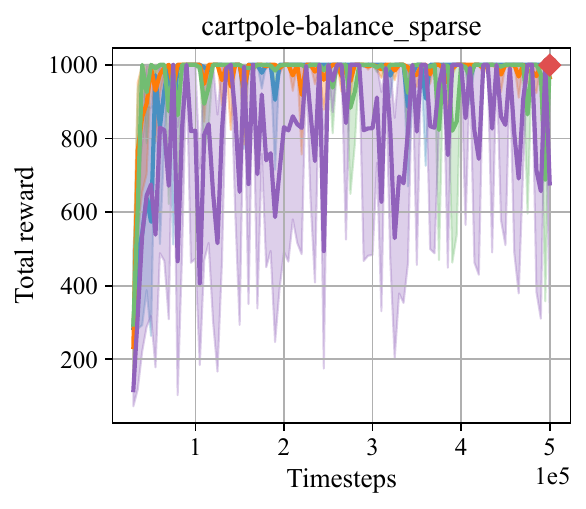}
    \end{minipage}

    \begin{minipage}[t]{0.245\textwidth}
        % \caption*{\quad \quad cartpole-swingup}
        \centering
        \includegraphics[width=\linewidth]{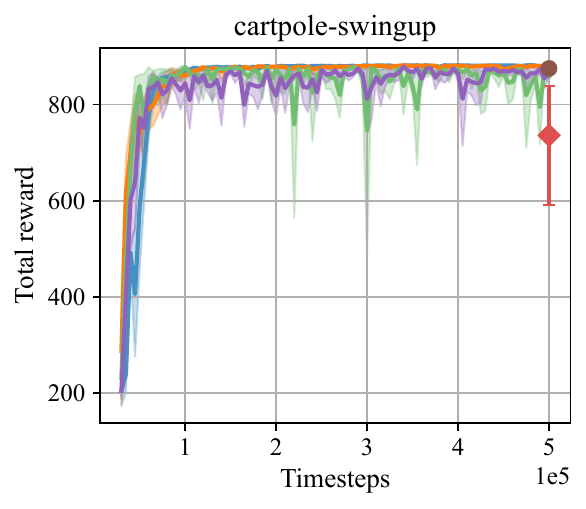}
    \end{minipage}
    \hfill
    \begin{minipage}[t]{0.245\textwidth}
        % \caption*{ \ \ cartpole-swingup\_sparse}
        \centering
        % \textbf{DMC (proprioceptive)}
        \includegraphics[width=\linewidth]{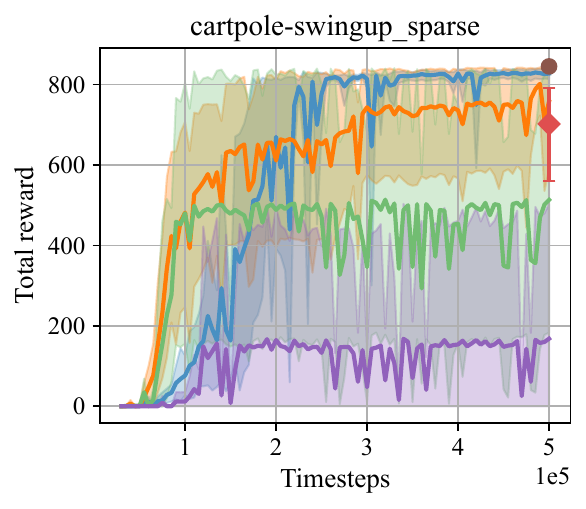}
    \end{minipage}
    \hfill
    \begin{minipage}[t]{0.245\textwidth}
        % \caption*{\quad \quad cheetah-run}
        \centering
        % \textbf{DMC (visual)}
        \includegraphics[width=\linewidth]{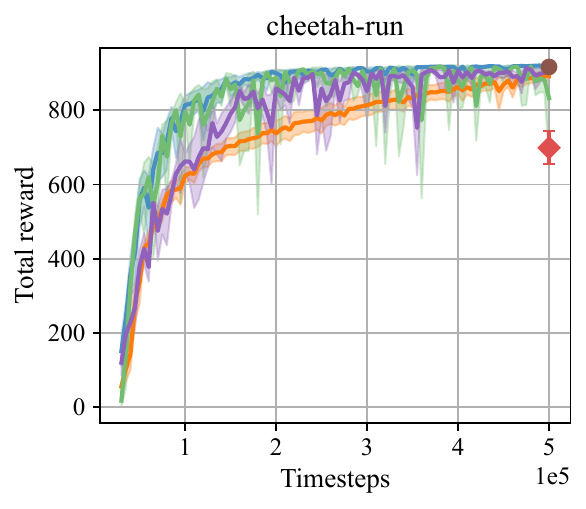}
    \end{minipage}
    \hfill
    \begin{minipage}[t]{0.245\textwidth}
        % \caption*{\quad \quad dog-run}
        \centering
        % \textbf{DMC (visual)}
        \includegraphics[width=\linewidth]{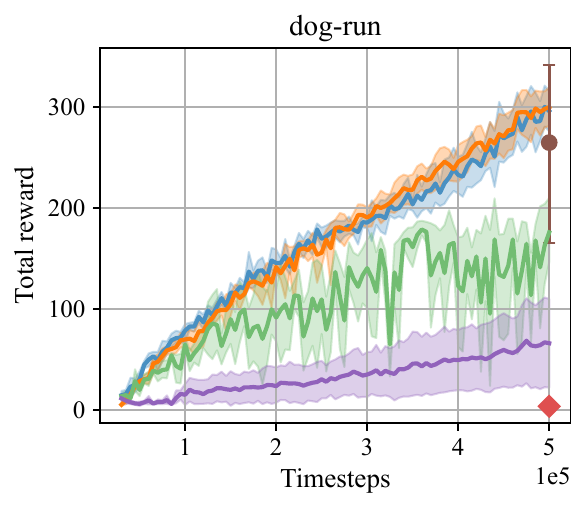}
    \end{minipage}
    
    \begin{minipage}[t]{0.245\textwidth}
        % \caption*{\quad \quad dog-stand}
        \centering
        \includegraphics[width=\linewidth]{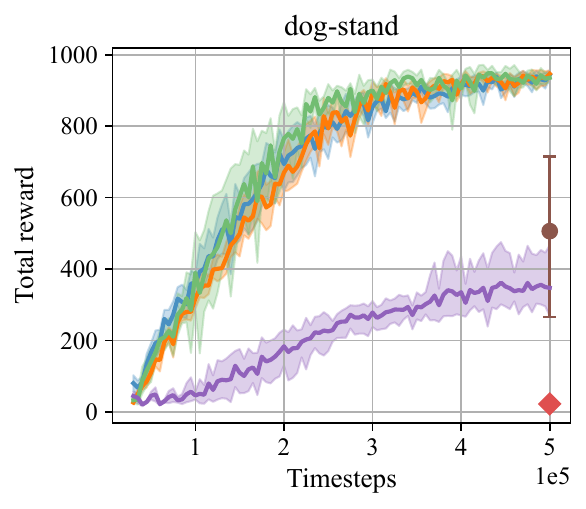}
    \end{minipage}
    \hfill
    \begin{minipage}[t]{0.245\textwidth}
        % \caption*{\quad \quad dog-trot}
        \centering
        \includegraphics[width=\linewidth]{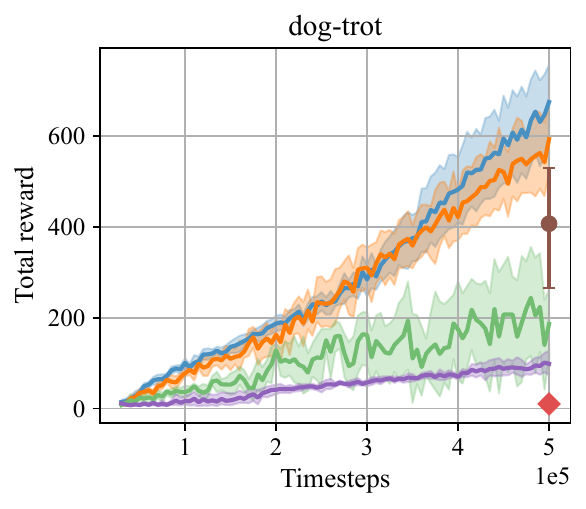}
    \end{minipage}
    \hfill
    \begin{minipage}[t]{0.245\textwidth}
        % \caption*{\quad \quad dog-walk}
        \centering
        \includegraphics[width=\linewidth]{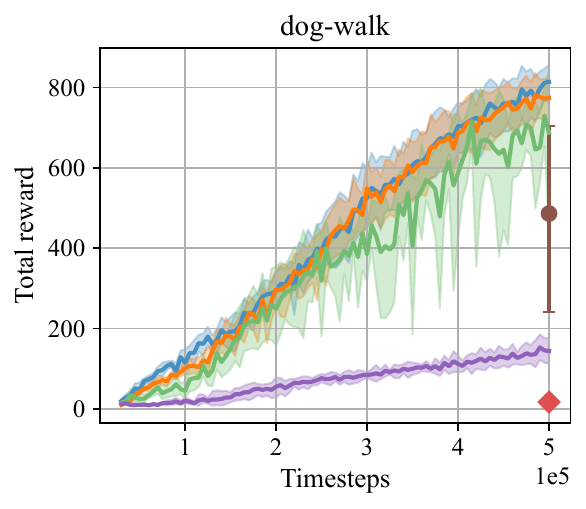}
    \end{minipage}
    \hfill
    \begin{minipage}[t]{0.245\textwidth}
        % \caption*{\quad \quad finger-spin}
        \centering
        \includegraphics[width=\linewidth]{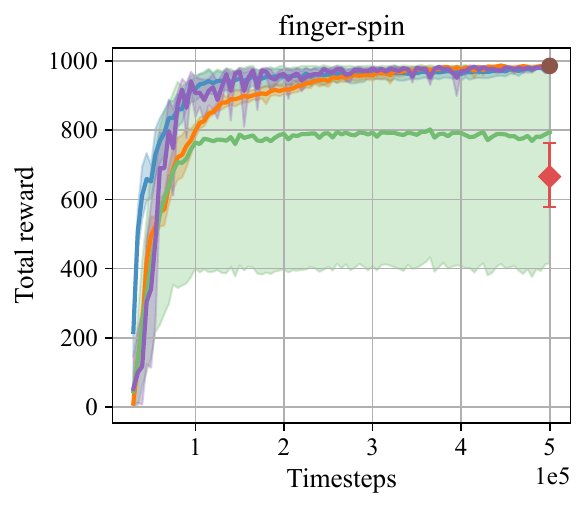}
    \end{minipage}

    \begin{minipage}[t]{0.245\textwidth}
        % \caption*{\quad \quad finger-turn\_easy}
        \centering
        \includegraphics[width=\linewidth]{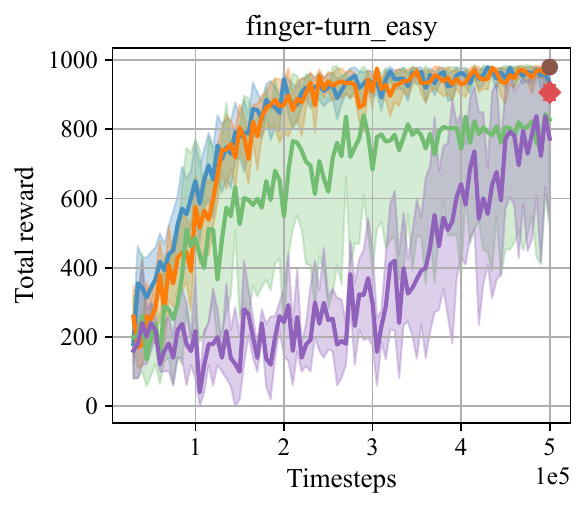}
    \end{minipage}
    \hfill
    \begin{minipage}[t]{0.245\textwidth}
        % \caption*{\quad \quad finger-turn\_hard}
        \centering
        \includegraphics[width=\linewidth]{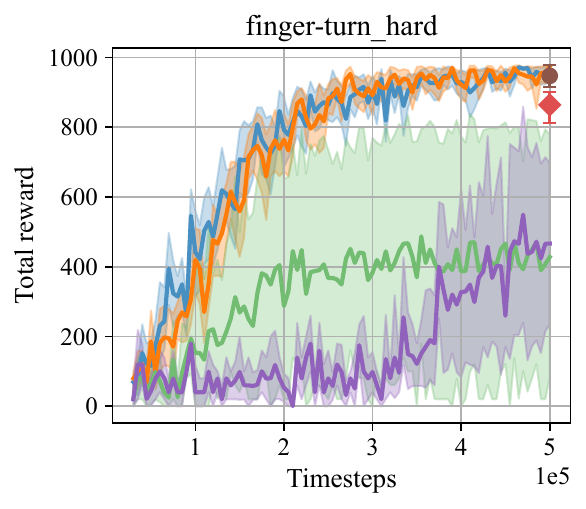}
    \end{minipage}
    \hfill
    \begin{minipage}[t]{0.245\textwidth}
        % \caption*{\quad \quad fish-swim}
        \centering
        \includegraphics[width=\linewidth]{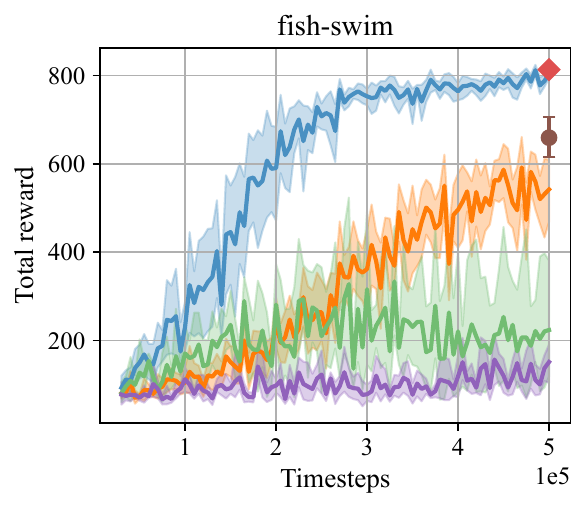}
    \end{minipage}
    \hfill
    \begin{minipage}[t]{0.245\textwidth}
        % \caption*{\quad \quad hopper-hop}
        \centering
        \includegraphics[width=\linewidth]{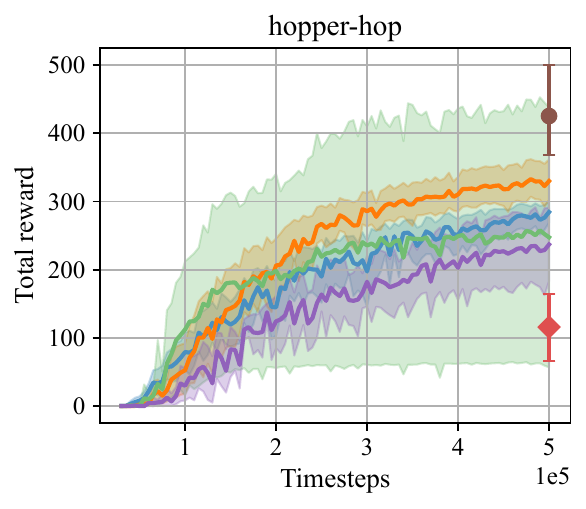}
    \end{minipage}

    \begin{minipage}[t]{0.245\textwidth}
        % \caption*{\quad \quad hopper-stand}
        \centering
        \includegraphics[width=\linewidth]{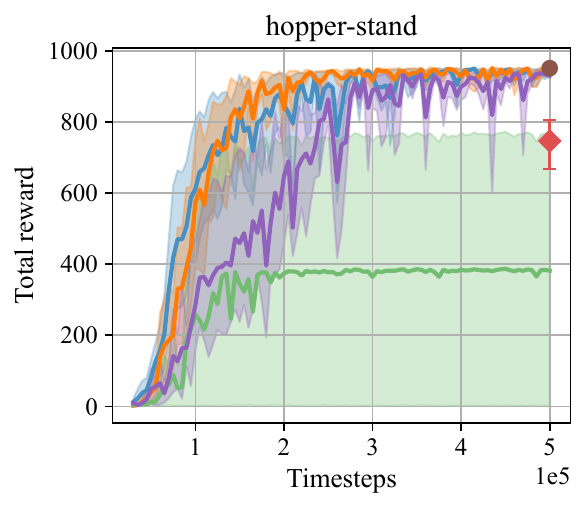}
    \end{minipage}
    \hfill
    \begin{minipage}[t]{0.245\textwidth}
        % \caption*{\quad \quad humanoid-run}
        \centering
        \includegraphics[width=\linewidth]{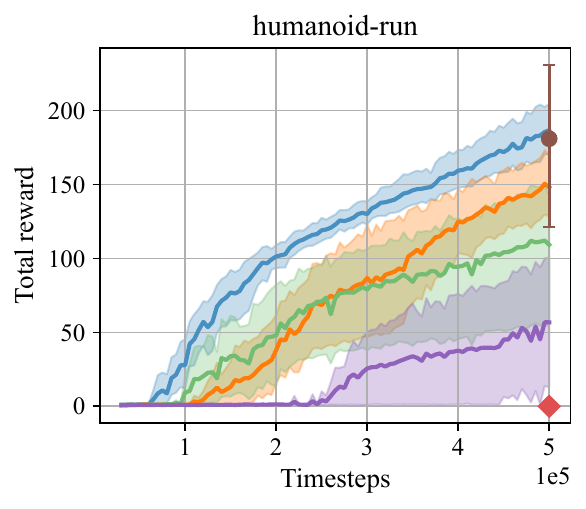}
    \end{minipage}
    \hfill
    \begin{minipage}[t]{0.245\textwidth}
        % \caption*{\quad \quad humanoid-stand}
        \centering
        \includegraphics[width=\linewidth]{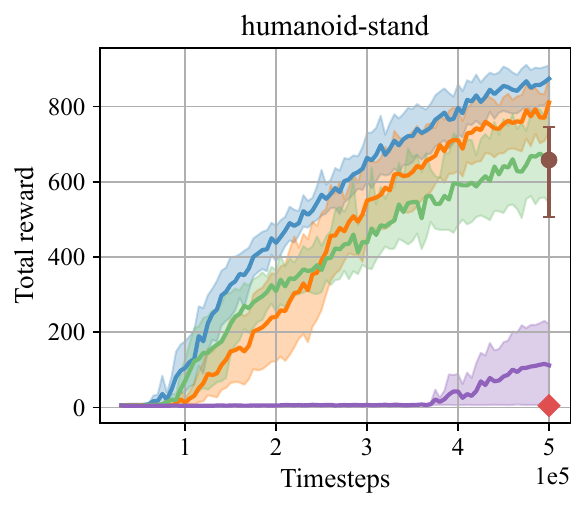}
    \end{minipage}
    \hfill
    \begin{minipage}[t]{0.245\textwidth}
        % \caption*{\quad \quad humanoid-walk}
        \centering
        \includegraphics[width=\linewidth]{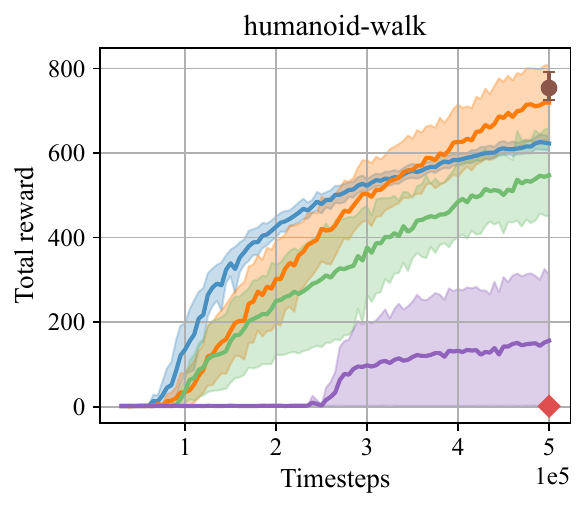}
    \end{minipage}

    \begin{minipage}[t]{0.245\textwidth}
        % \caption*{\quad \quad pendulum-swingup}
        \centering
        \includegraphics[width=\linewidth]{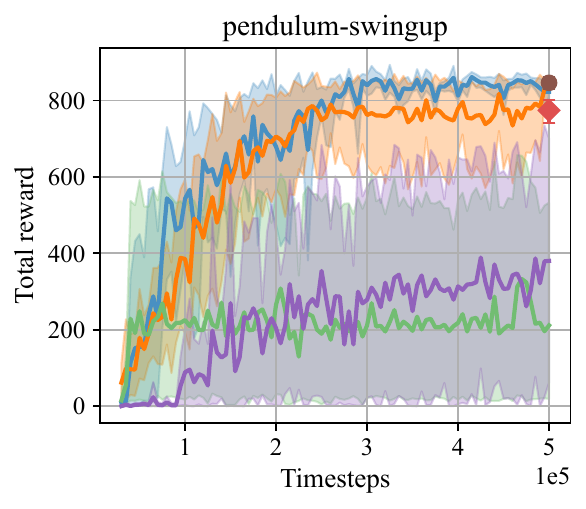}
    \end{minipage}
    \hfill
    \begin{minipage}[t]{0.245\textwidth}
        % \caption*{\quad \quad quadruped-run}
        \centering
        \includegraphics[width=\linewidth]{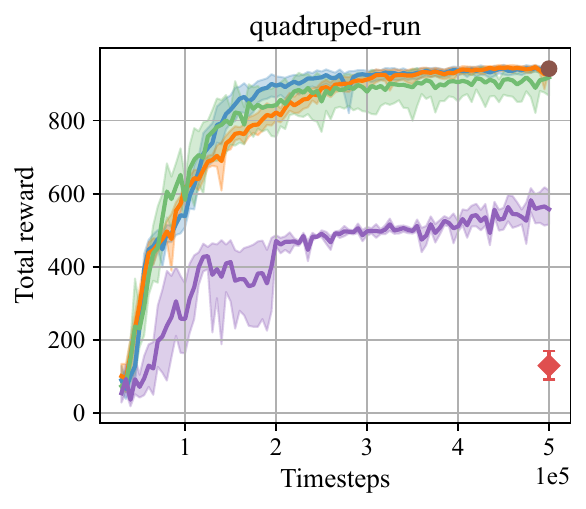}
    \end{minipage}
    \hfill
    \begin{minipage}[t]{0.245\textwidth}
        % \caption*{\quad \quad quadruped-walk}
        \centering
        \includegraphics[width=\linewidth]{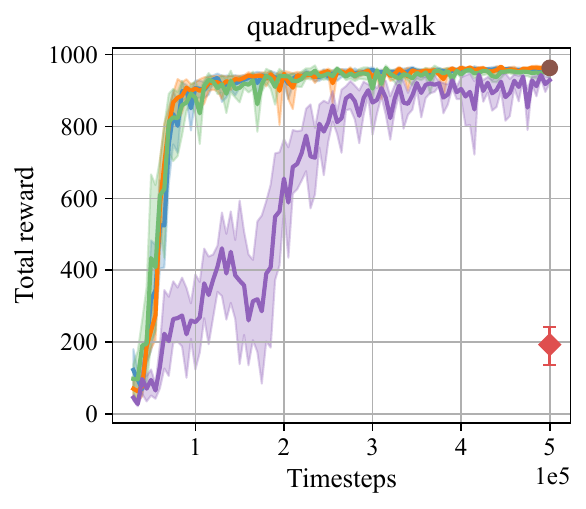}
    \end{minipage}
    \hfill
    \begin{minipage}[t]{0.245\textwidth}
        % \caption*{\quad \quad reacher-easy}
        \centering
        \includegraphics[width=\linewidth]{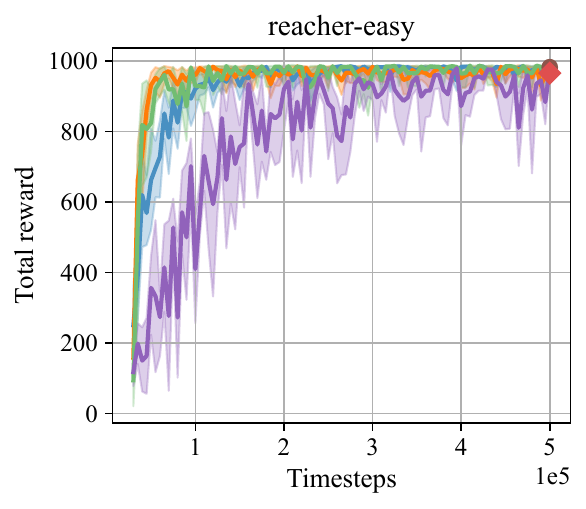}
    \end{minipage}

    \begin{minipage}[t]{0.245\textwidth}
        % \caption*{\quad \quad reacher-hard}
        \centering
        \includegraphics[width=\linewidth]{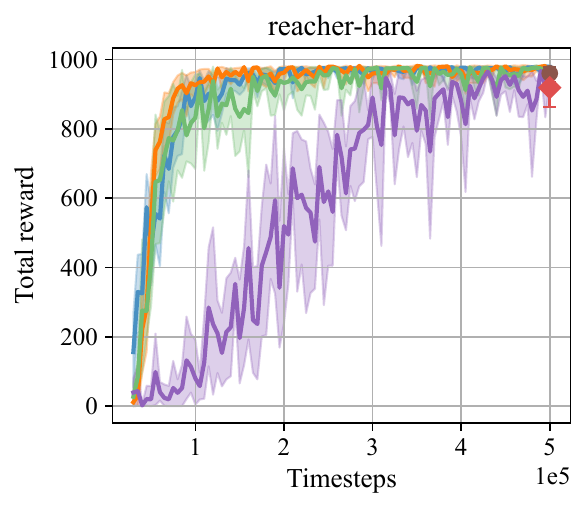}
    \end{minipage}
    \hfill
    \begin{minipage}[t]{0.245\textwidth}
        % \caption*{\quad \quad walker-run}
        \centering
        \includegraphics[width=\linewidth]{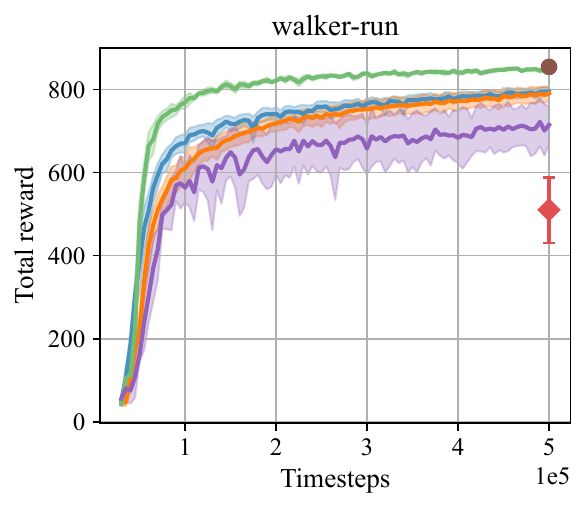}
    \end{minipage}
    \hfill
    \begin{minipage}[t]{0.245\textwidth}
        % \caption*{\quad \quad walker-stand}
        \centering
        \includegraphics[width=\linewidth]{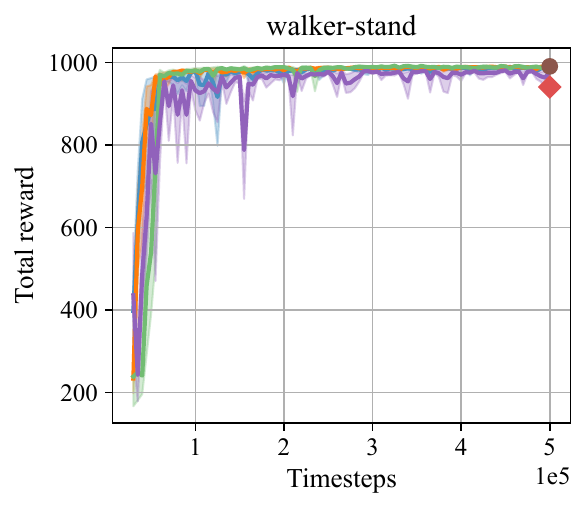}
    \end{minipage}
    \hfill
    \begin{minipage}[t]{0.245\textwidth}
       %  \caption*{\quad \quad walker-walk}
        \centering
        \includegraphics[width=\linewidth]{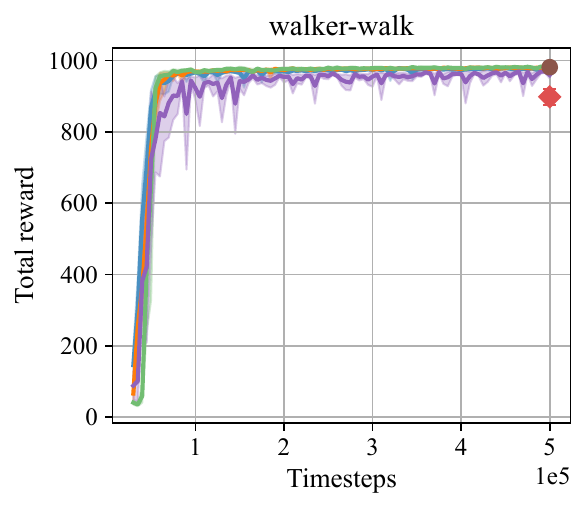}
    \end{minipage}

    \begin{minipage}[t]{0.8\textwidth}
        \centering
        % \vspace{1 em}
        \includegraphics[width=\linewidth]{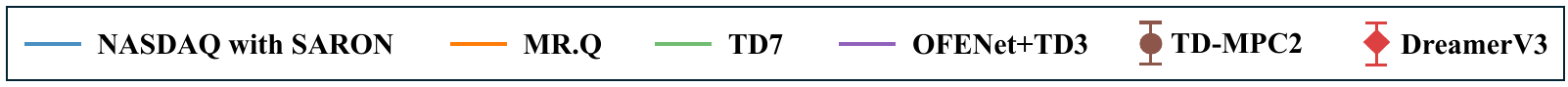}
    \end{minipage}
    
    \caption{
     Learning curves on \textbf{DMC (proprioceptive)}. Solid lines indicate average performance over 5 seeds, and shaded areas indicate the 95\% bootstrap confidence interval. Discrete points with 95\% bootstrap confidence interval denote the final results of TD-MPC2 and DreamerV3 reported in MR.Q.
    }
    \label{fig:main_result_dmc_p2}
\end{figure}

\begin{minipage}{\textwidth}
\subsection{DMC (visual)}
\label{app:full_dmc_v}
\begin{table}[H]
    \caption{
    Final performance on the \textbf{DMC (visual)} benchmark at 500k time steps (1M time steps in the original environment due to action repeat), averaged over 5 seeds. We use the results reported in MR.Q for DrQ-v2, TD-MPC2, and DreamerV3. The {\textcolor{gray}{[bracketed values]}} represent a 95\% bootstrap confidence interval. 
    }
    \centering
    \begin{adjustbox}{max width=\textwidth}
    \begin{tabular}{lcccccc}
        \toprule
        \textbf{Tasks} & \textbf{DrQ-v2} & \textbf{TD-MPC2} &
        \textbf{DreamerV3} & \textbf{MR.Q} & \textbf{NASDAQ} \\
        \midrule
        acrobot-swingup & 
        168 {\textcolor{gray}{[127, 219]}} & 
        197 {\textcolor{gray}{[179, 217]}} & 
        121 {\textcolor{gray}{[106, 145]}} & 
        315 {\textcolor{gray}{[268, 362]}} &
        264 {\textcolor{gray}{[171, 356]}} \\
        
        ball\_in\_cup-catch & 
        909 {\textcolor{gray}{[821, 973]}} & 
        932 {\textcolor{gray}{[899, 961]}} & 
        971 {\textcolor{gray}{[969, 973]}} & 
        975 {\textcolor{gray}{[971, 979]}} &
        975 {\textcolor{gray}{[973, 978]}} \\
        
        cartpole-balance & 
        993 {\textcolor{gray}{[990, 996]}} & 
        972 {\textcolor{gray}{[948, 991]}} & 
        998 {\textcolor{gray}{[997, 1000]}} & 
        998 {\textcolor{gray}{[997, 999]}} &
        997 {\textcolor{gray}{[996, 998]}} \\
        
        cartpole-balance\_sparse & 
        962 {\textcolor{gray}{[887, 1000]}} & 
        1000 {\textcolor{gray}{[1000, 1000]}} & 
        999 {\textcolor{gray}{[999, 1000]}} & 
        1000 {\textcolor{gray}{[1000, 1000]}} &
        1000 {\textcolor{gray}{[1000, 1000]}} \\
        
        cartpole-swingup & 
        864 {\textcolor{gray}{[854, 873]}} & 
        690 {\textcolor{gray}{[521, 813]}} & 
        725 {\textcolor{gray}{[603, 807]}} &  
        863 {\textcolor{gray}{[857, 870]}} & 
        873 {\textcolor{gray}{[867, 878]}} \\

        cartpole-swingup\_sparse & 
        774 {\textcolor{gray}{[741, 805]}} & 
        636 {\textcolor{gray}{[404, 804]}} & 
        547 {\textcolor{gray}{[351, 726]}} & 
        782 {\textcolor{gray}{[739, 818]}} & 
        835 {\textcolor{gray}{[827, 843]}} \\
        
        cheetah-run & 
        728 {\textcolor{gray}{[701, 753]}} & 
        431 {\textcolor{gray}{[267, 556]}} & 
        618 {\textcolor{gray}{[576, 661]}} & 
        759 {\textcolor{gray}{[752, 768]}} & 
        853 {\textcolor{gray}{[821, 885]}} \\
        
        dog-run & 
        10 {\textcolor{gray}{[9, 12]}} & 
        14 {\textcolor{gray}{[10, 18]}} & 
        9 {\textcolor{gray}{[6, 14]}} &   
        58 {\textcolor{gray}{[46, 74]}} & 
        63 {\textcolor{gray}{[48, 79]}} \\
        
        dog-stand & 
        43 {\textcolor{gray}{[37, 49]}} & 
        117 {\textcolor{gray}{[72, 148]}} & 
        61 {\textcolor{gray}{[30, 92]}} &  
        239 {\textcolor{gray}{[222, 258]}} & 
        244 {\textcolor{gray}{[216, 269]}}  \\
        
        dog-trot & 
        14 {\textcolor{gray}{[11, 18]}} & 
        20 {\textcolor{gray}{[14, 25]}} & 
        14 {\textcolor{gray}{[13, 16]}} &  
        65 {\textcolor{gray}{[57, 75]}} & 
        63 {\textcolor{gray}{[59, 68]}} \\

        dog-walk & 
        22 {\textcolor{gray}{[18, 29]}} & 
        22 {\textcolor{gray}{[17, 28]}} & 
        11 {\textcolor{gray}{[11, 12]}} &
        84 {\textcolor{gray}{[76, 95]}} & 
        83 {\textcolor{gray}{[75, 93]}} \\
        
        finger-spin & 
        860 {\textcolor{gray}{[787, 922]}} & 
        786 {\textcolor{gray}{[492, 984]}} & 
        656 {\textcolor{gray}{[544, 765]}} & 
        835 {\textcolor{gray}{[682, 982]}} & 
        986 {\textcolor{gray}{[985, 987]}} \\
        
        finger-turn\_easy & 
        503 {\textcolor{gray}{[399, 615]}} & 
        562 {\textcolor{gray}{[317, 779]}} & 
        491 {\textcolor{gray}{[447, 542]}} &  
        924 {\textcolor{gray}{[862, 974]}} & 
        915 {\textcolor{gray}{[873, 956]}} \\
        
        finger-turn\_hard & 
        223 {\textcolor{gray}{[121, 340]}} & 
        903 {\textcolor{gray}{[870, 940]}} & 
        494 {\textcolor{gray}{[401, 571]}} &  
        929 {\textcolor{gray}{[891, 967]}} & 
        914 {\textcolor{gray}{[871, 958]}}  \\
        
        fish-swim & 
        84 {\textcolor{gray}{[65, 107]}} & 
        43 {\textcolor{gray}{[21, 64]}} & 
        90 {\textcolor{gray}{[84, 96]}} &  
        70 {\textcolor{gray}{[63, 77]}} & 
        67 {\textcolor{gray}{[64, 70]}} \\

        hopper-hop & 
        224 {\textcolor{gray}{[170, 278]}} & 
        187 {\textcolor{gray}{[119, 238]}} & 
        205 {\textcolor{gray}{[125, 287]}} & 
        254 {\textcolor{gray}{[204, 299]}} & 
        283 {\textcolor{gray}{[256, 308]}} \\
        
        hopper-stand & 
        917 {\textcolor{gray}{[903, 931]}} & 
        582 {\textcolor{gray}{[321, 794]}} & 
        888 {\textcolor{gray}{[875, 900]}} & 
        925 {\textcolor{gray}{[919, 930]}} & 
        902 {\textcolor{gray}{[850, 934]}} \\
        
        humanoid-run & 
        1 {\textcolor{gray}{[1, 1]}} & 
        0 {\textcolor{gray}{[0, 1]}} & 
        1 {\textcolor{gray}{[1, 1]}} &  
        1 {\textcolor{gray}{[1, 2]}} & 
        1 {\textcolor{gray}{[1, 1]}} \\
        
        humanoid-stand & 
        6 {\textcolor{gray}{[7, 7]}} & 
        5 {\textcolor{gray}{[5, 7]}} & 
        5 {\textcolor{gray}{[5, 7]}} &  
        7 {\textcolor{gray}{[7, 8]}} & 
        6 {\textcolor{gray}{[5, 7]}}  \\
        
        humanoid-walk & 
        2 {\textcolor{gray}{[2, 2]}} & 
        1 {\textcolor{gray}{[1, 2]}} & 
        1 {\textcolor{gray}{[2, 2]}} &  
        3 {\textcolor{gray}{[2, 5]}} & 
        2 {\textcolor{gray}{[2, 3]}} \\

        pendulum-swingup & 
        838 {\textcolor{gray}{[813, 861]}} & 
        748 {\textcolor{gray}{[574, 850]}} & 
        761 {\textcolor{gray}{[709, 807]}} & 
        852 {\textcolor{gray}{[832, 874]}} &
        836 {\textcolor{gray}{[816, 858]}} \\
        
        quadruped-run & 
        459 {\textcolor{gray}{[412, 507]}} & 
        262 {\textcolor{gray}{[184, 330]}} & 
        328 {\textcolor{gray}{[255, 397]}} & 
        493 {\textcolor{gray}{[465, 517]}} & 
        471 {\textcolor{gray}{[458, 495]}} \\
        
        quadruped-walk & 
        750 {\textcolor{gray}{[699, 796]}} & 
        246 {\textcolor{gray}{[179, 310]}} & 
        316 {\textcolor{gray}{[260, 379]}} &  
        856 {\textcolor{gray}{[800, 912]}} & 
        809 {\textcolor{gray}{[748, 855]}} \\
        
        reacher-easy & 
        938 {\textcolor{gray}{[903, 973]}} & 
        956 {\textcolor{gray}{[932, 978]}} & 
        735 {\textcolor{gray}{[678, 796]}} &  
        977 {\textcolor{gray}{[974, 980]}} & 
        977 {\textcolor{gray}{[975, 979]}}  \\
        
        reacher-hard & 
        705 {\textcolor{gray}{[580, 831]}} & 
        911 {\textcolor{gray}{[867, 946]}} & 
        338 {\textcolor{gray}{[227, 461]}} &  
        951 {\textcolor{gray}{[913, 972]}} & 
        935 {\textcolor{gray}{[893, 977]}} \\

        walker-run & 
        546 {\textcolor{gray}{[475, 612]}} & 
        665 {\textcolor{gray}{[566, 719]}} & 
        669 {\textcolor{gray}{[615, 708]}} &  
        572 {\textcolor{gray}{[502, 638]}} & 
        676 {\textcolor{gray}{[651, 704]}} \\
        
        walker-stand & 
        980 {\textcolor{gray}{[977, 984]}} & 
        937 {\textcolor{gray}{[907, 962]}} & 
        969 {\textcolor{gray}{[966, 973]}} &  
        985 {\textcolor{gray}{[984, 986]}} & 
        987 {\textcolor{gray}{[985, 989]}}  \\
        
        walker-walk & 
        766 {\textcolor{gray}{[489, 957]}} & 
        958 {\textcolor{gray}{[952, 965]}} & 
        942 {\textcolor{gray}{[936, 949]}} &  
        967 {\textcolor{gray}{[962, 971]}} & 
        963 {\textcolor{gray}{[958, 968]}} \\
        \midrule
        \multicolumn{4}{l}{\textbf{Aggregate Results}} \\
        \midrule
        Mean & 510 & 492 & 463 & 598 & 606  \\
        Median & 626 & 572 & 493 & 809 & 836   \\
        IQM & 545 & 501 & 452 & 686 & 701 \\
        \bottomrule
    \end{tabular}
    \label{tab:main_result_dmc2}
    \end{adjustbox}
\end{table}
\end{minipage}

\begin{figure}[H]
    \centering
    \begin{minipage}[t]{0.245\textwidth}
        % \caption*{\quad \quad acrobot-swingup}
        \centering
        \includegraphics[width=\linewidth]{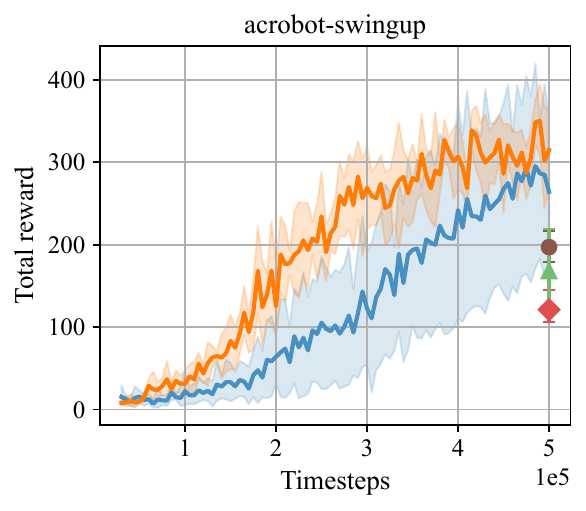}
    \end{minipage}
    \hfill
    \begin{minipage}[t]{0.245\textwidth}
        % % \caption*{\quad \quad ball\_in\_cup-catch}
        \centering
        \includegraphics[width=\linewidth]{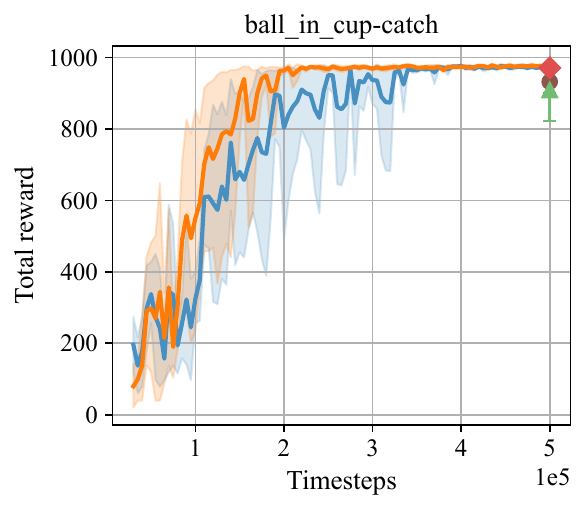}
    \end{minipage}
    \hfill
    \begin{minipage}[t]{0.245\textwidth}
        % \caption*{\quad \quad cartpole-balance}
        \centering
        \includegraphics[width=\linewidth]{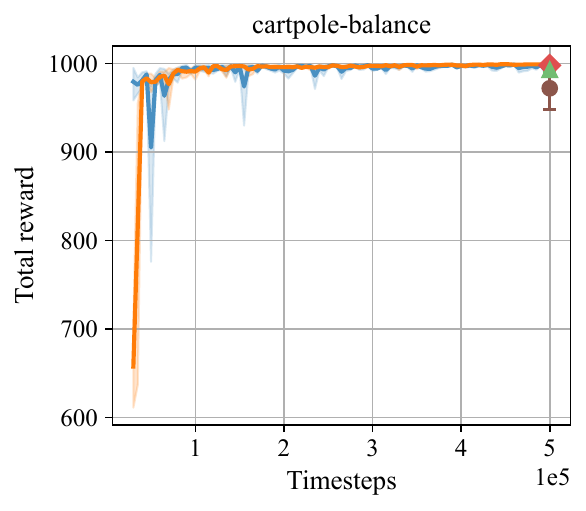}
    \end{minipage}
    \hfill
    \begin{minipage}[t]{0.245\textwidth}
        % \caption*{ \quad cartpole-balance\_sparse}
        \centering
        \includegraphics[width=\linewidth]{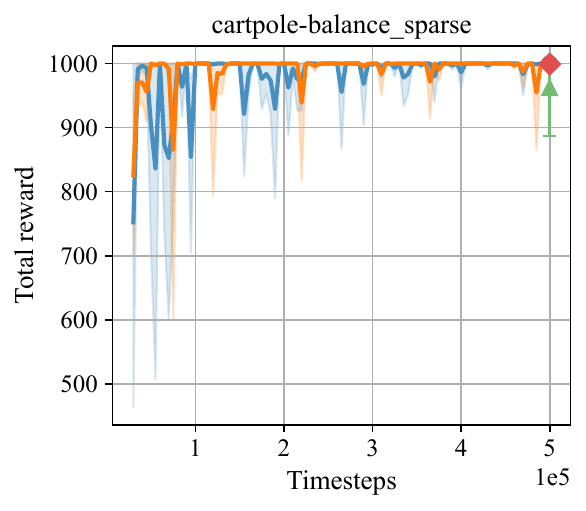}
    \end{minipage}

    \begin{minipage}[t]{0.245\textwidth}
        % \caption*{\quad \quad cartpole-swingup}
        \centering
        \includegraphics[width=\linewidth]{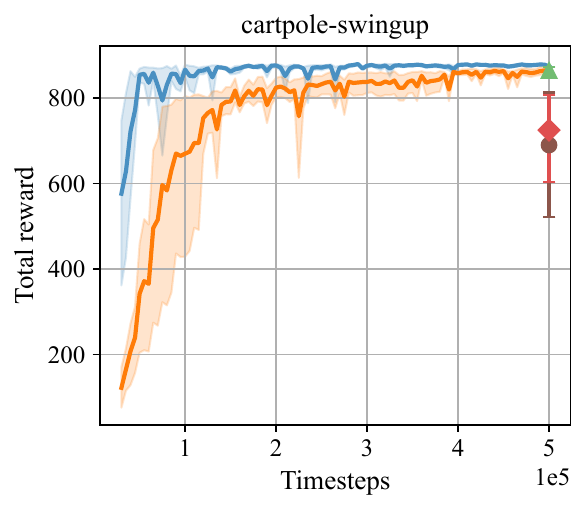}
    \end{minipage}
    \hfill
    \begin{minipage}[t]{0.245\textwidth}
        % \caption*{ \ \ cartpole-swingup\_sparse}
        \centering
        % \textbf{DMC (proprioceptive)}
        \includegraphics[width=\linewidth]{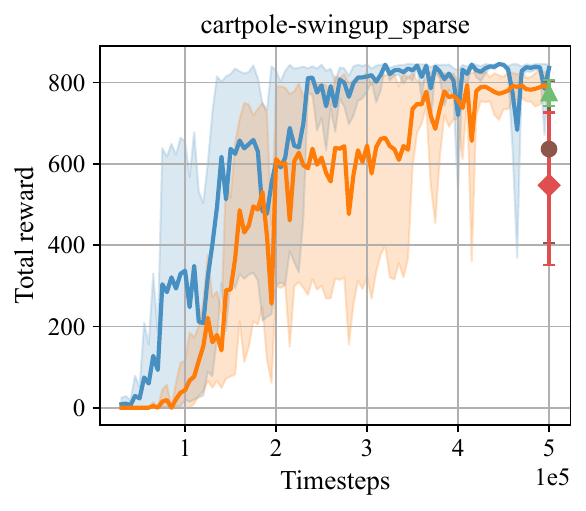}
    \end{minipage}
    \hfill
    \begin{minipage}[t]{0.245\textwidth}
        % \caption*{\quad \quad cheetah-run}
        \centering
        % \textbf{DMC (visual)}
        \includegraphics[width=\linewidth]{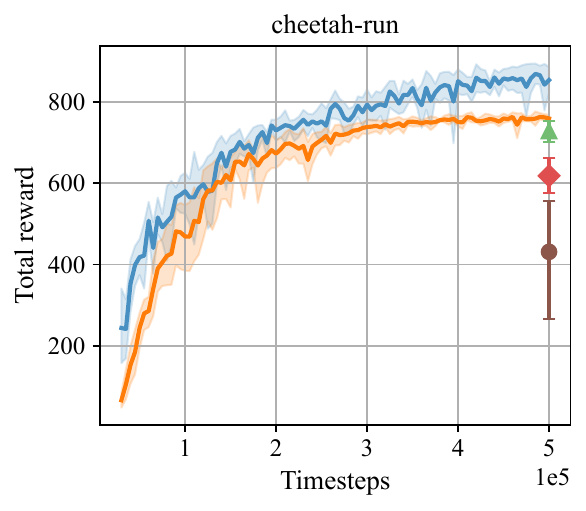}
    \end{minipage}
    \hfill
    \begin{minipage}[t]{0.245\textwidth}
        % \caption*{\quad \quad dog-run}
        \centering
        % \textbf{DMC (visual)}
        \includegraphics[width=\linewidth]{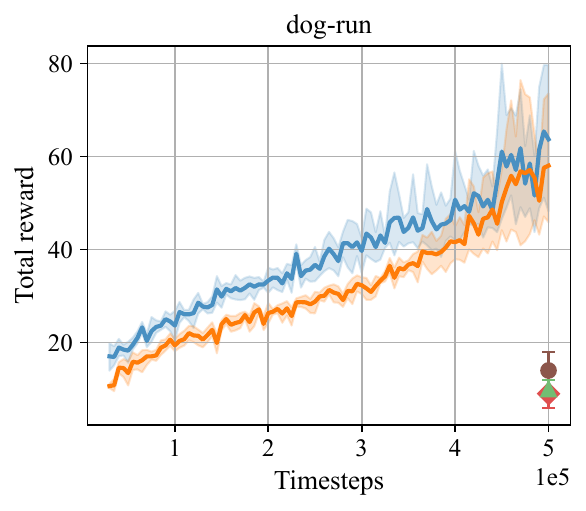}
    \end{minipage}
    \begin{minipage}[t]{0.245\textwidth}
        % \caption*{\quad \quad dog-stand}
        \centering
        \includegraphics[width=\linewidth]{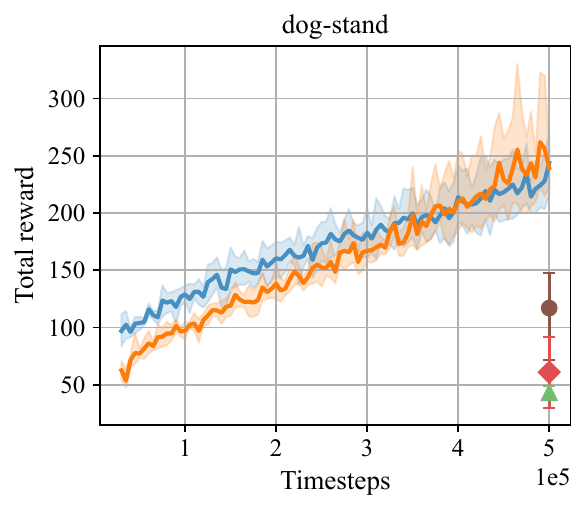}
    \end{minipage}
    \hfill
    \begin{minipage}[t]{0.245\textwidth}
        % \caption*{\quad \quad dog-trot}
        \centering
        \includegraphics[width=\linewidth]{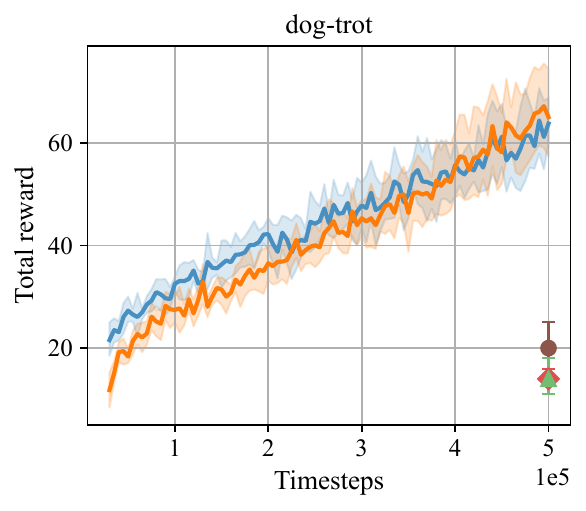}
    \end{minipage}
    \hfill
    \begin{minipage}[t]{0.245\textwidth}
        % \caption*{\quad \quad dog-walk}
        \centering
        \includegraphics[width=\linewidth]{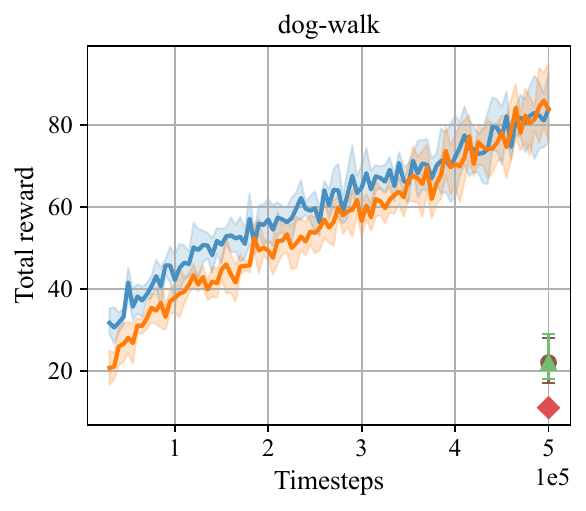}
    \end{minipage}
    \hfill
    \begin{minipage}[t]{0.245\textwidth}
        % \caption*{\quad \quad finger-spin}
        \centering
        \includegraphics[width=\linewidth]{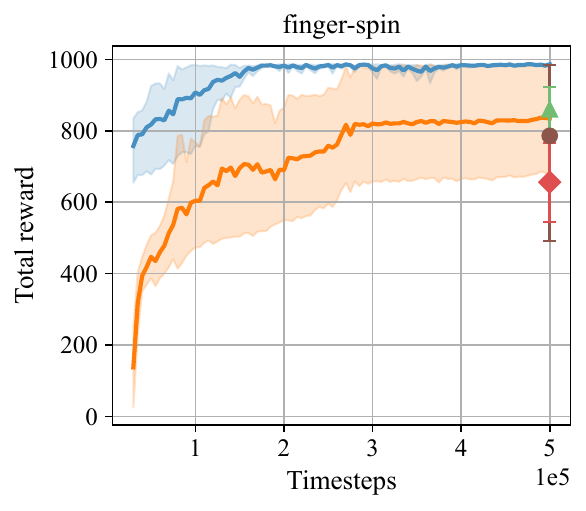}
    \end{minipage}

    \begin{minipage}[t]{0.245\textwidth}
        % \caption*{\quad \quad finger-turn\_easy}
        \centering
        \includegraphics[width=\linewidth]{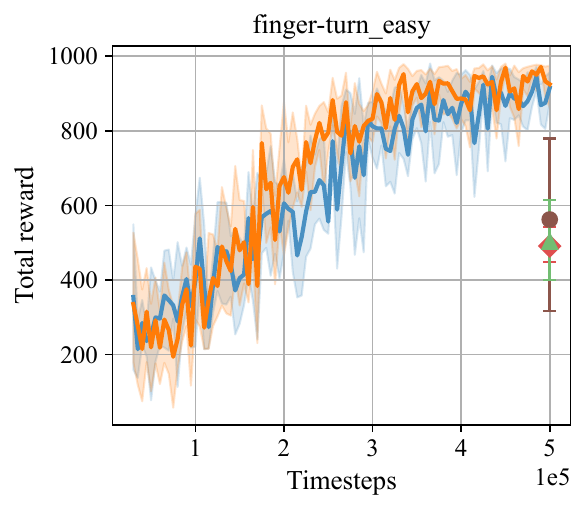}
    \end{minipage}
    \hfill
    \begin{minipage}[t]{0.245\textwidth}
        % \caption*{\quad \quad finger-turn\_hard}
        \centering
        \includegraphics[width=\linewidth]{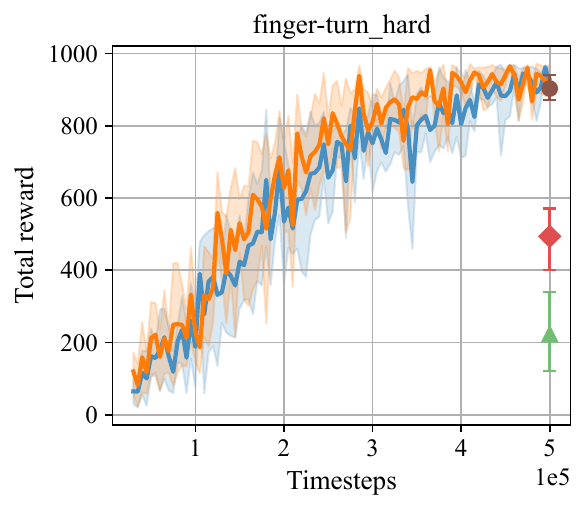}
    \end{minipage}
    \hfill
    \begin{minipage}[t]{0.245\textwidth}
        % \caption*{\quad \quad fish-swim}
        \centering
        \includegraphics[width=\linewidth]{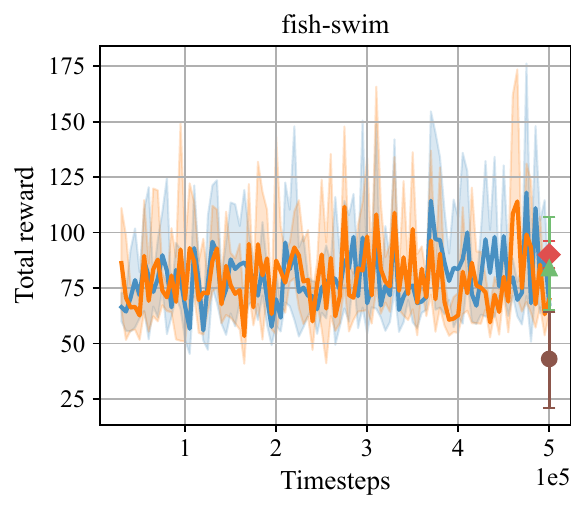}
    \end{minipage}
    \hfill
    \begin{minipage}[t]{0.245\textwidth}
        % \caption*{\quad \quad hopper-hop}
        \centering
        \includegraphics[width=\linewidth]{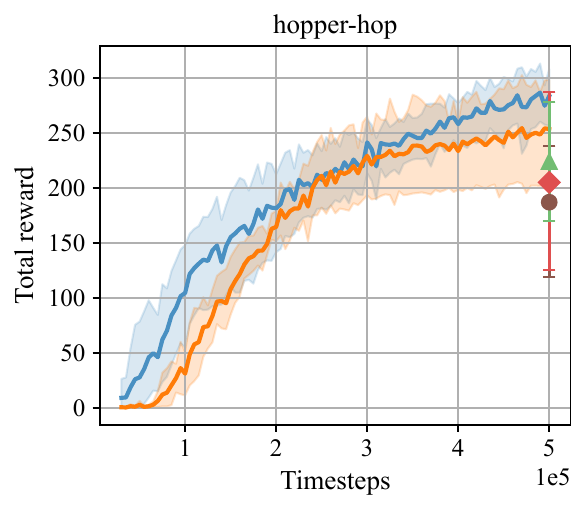}
    \end{minipage}

    \begin{minipage}[t]{0.245\textwidth}
        % \caption*{\quad \quad hopper-stand}
        \centering
        \includegraphics[width=\linewidth]{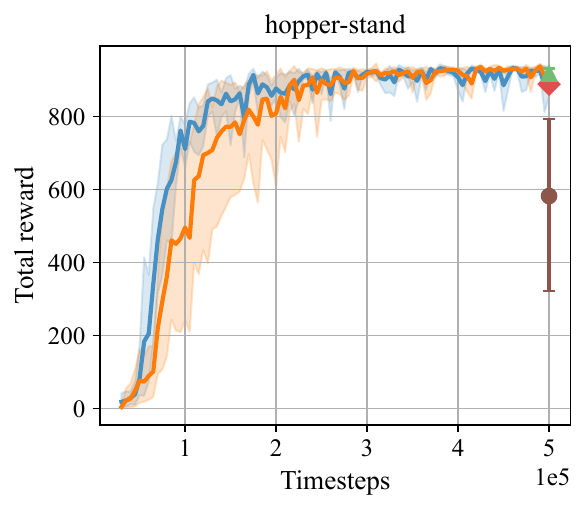}
    \end{minipage}
    \hfill
    \begin{minipage}[t]{0.245\textwidth}
        % \caption*{\quad \quad humanoid-run}
        \centering
        \includegraphics[width=\linewidth]{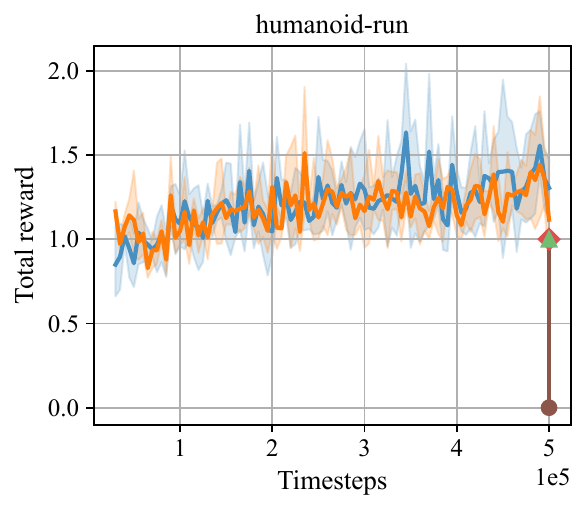}
    \end{minipage}
    \hfill
    \begin{minipage}[t]{0.245\textwidth}
        % \caption*{\quad \quad humanoid-stand}
        \centering
        \includegraphics[width=\linewidth]{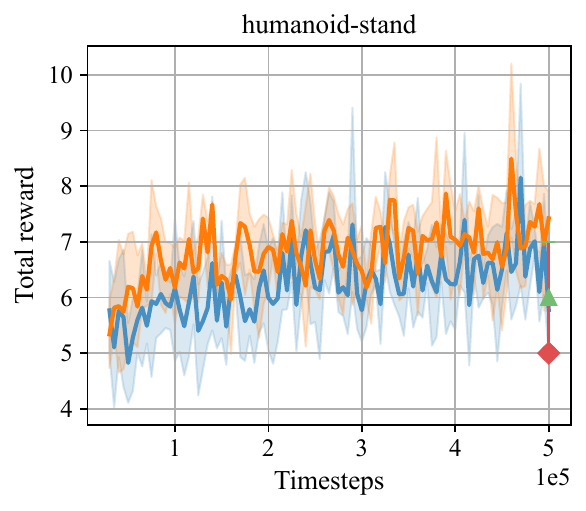}
    \end{minipage}
    \hfill
    \begin{minipage}[t]{0.245\textwidth}
        % \caption*{\quad \quad humanoid-walk}
        \centering
        \includegraphics[width=\linewidth]{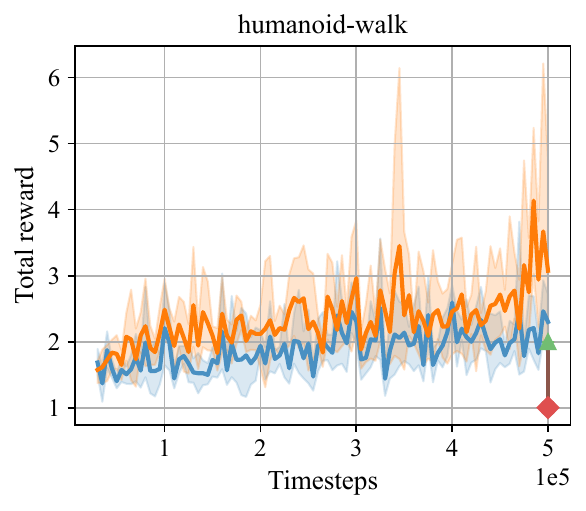}
    \end{minipage}

    \begin{minipage}[t]{0.245\textwidth}
        % \caption*{\quad \quad pendulum-swingup}
        \centering
        \includegraphics[width=\linewidth]{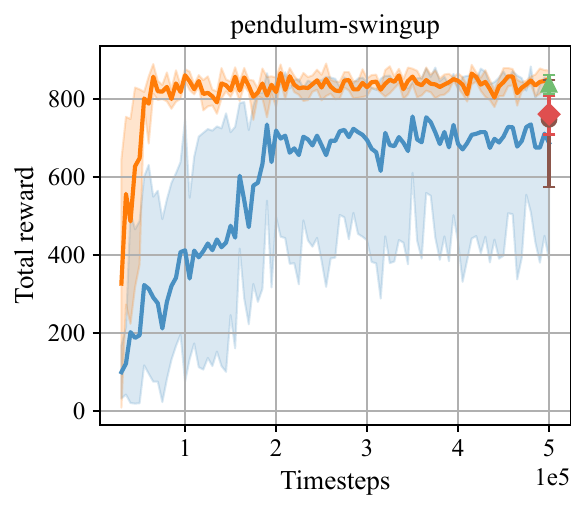}
    \end{minipage}
    \hfill
    \begin{minipage}[t]{0.245\textwidth}
        % \caption*{\quad \quad quadruped-run}
        \centering
        \includegraphics[width=\linewidth]{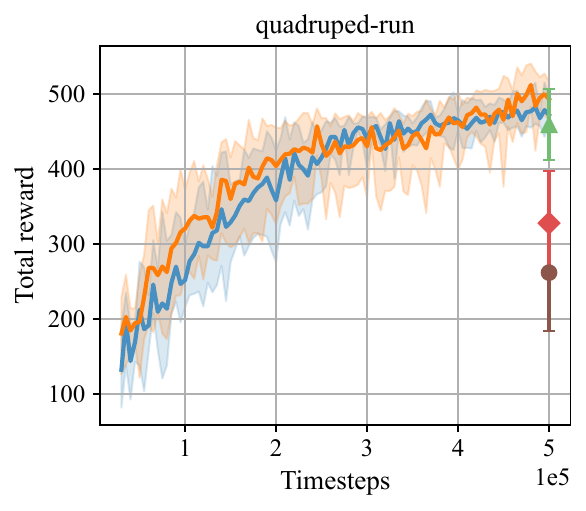}
    \end{minipage}
    \hfill
    \begin{minipage}[t]{0.245\textwidth}
        % \caption*{\quad \quad quadruped-walk}
        \centering
        \includegraphics[width=\linewidth]{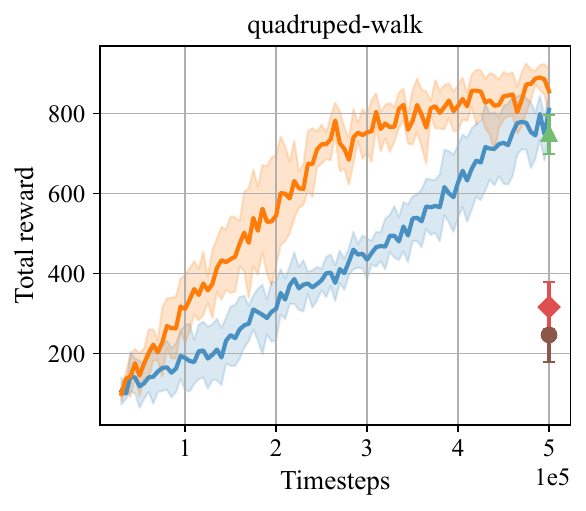}
    \end{minipage}
    \hfill
    \begin{minipage}[t]{0.245\textwidth}
        % \caption*{\quad \quad reacher-easy}
        \centering
        \includegraphics[width=\linewidth]{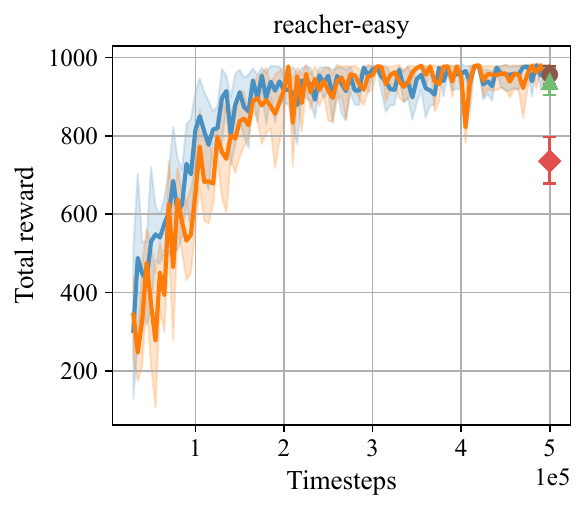}
    \end{minipage}

    \begin{minipage}[t]{0.245\textwidth}
        % \caption*{\quad \quad reacher-hard}
        \centering
        \includegraphics[width=\linewidth]{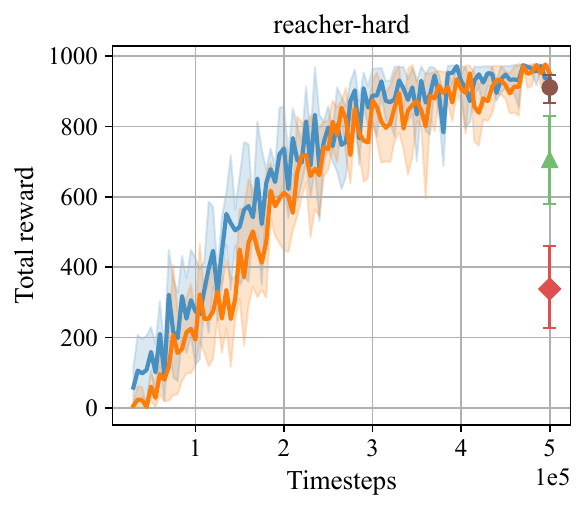}
    \end{minipage}
    \hfill
    \begin{minipage}[t]{0.245\textwidth}
        % \caption*{\quad \quad walker-run}
        \centering
        \includegraphics[width=\linewidth]{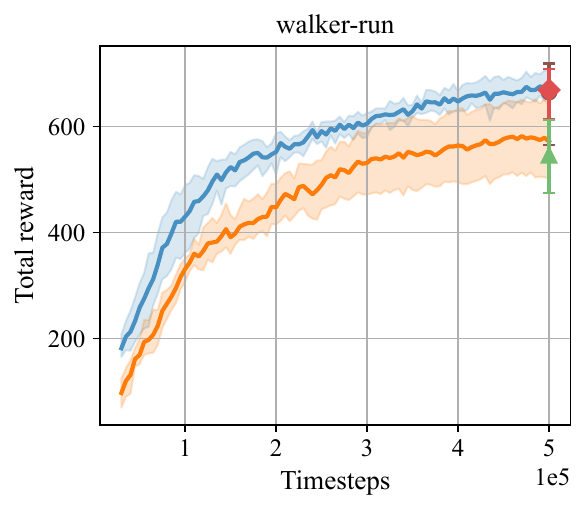}
    \end{minipage}
    \hfill
    \begin{minipage}[t]{0.245\textwidth}
        % \caption*{\quad \quad walker-stand}
        \centering
        \includegraphics[width=\linewidth]{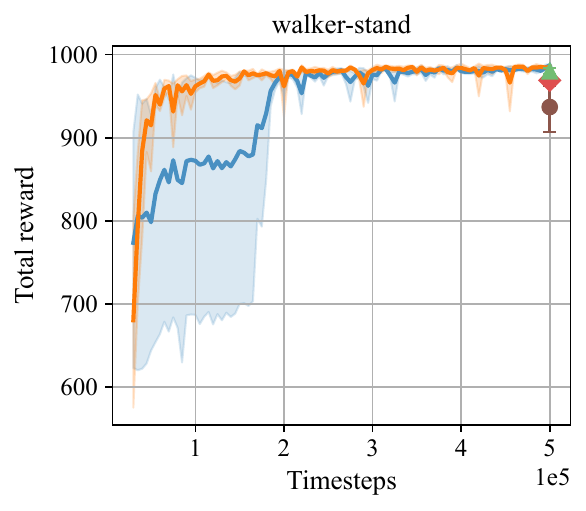}
    \end{minipage}
    \hfill
    \begin{minipage}[t]{0.245\textwidth}
        % \caption*{\quad \quad walker-walk}
        \centering
        \includegraphics[width=\linewidth]{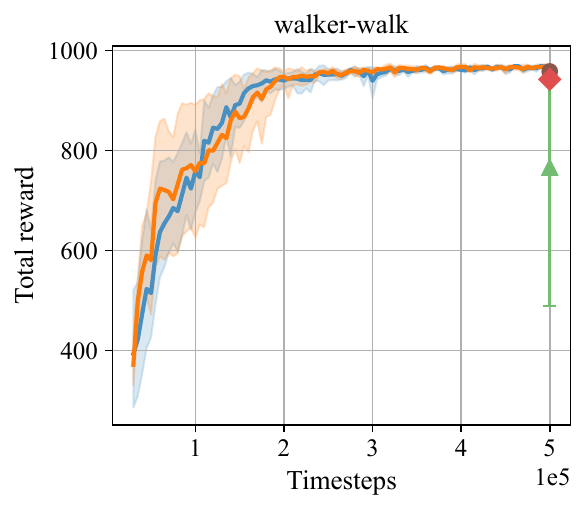}
    \end{minipage}

    \begin{minipage}[t]{0.62\textwidth}
        \centering
        % \vspace{1 em}
        \includegraphics[width=\linewidth]{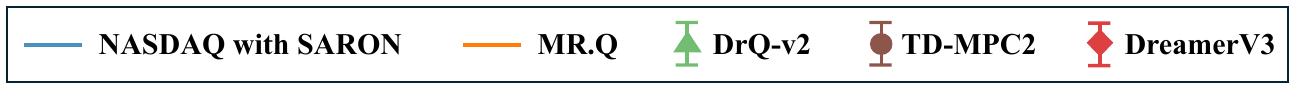}
    \end{minipage}
    \caption{
     Learning curves on \textbf{DMC (visual)}. Solid lines indicate average performance over 5 seeds, and shaded areas indicate the 95\% bootstrap confidence interval. Discrete points with 95\% bootstrap confidence interval denote the final results of DreamerV3 reported in MR.Q.
    }
    \label{fig:main_result_dmc_visual2}
\end{figure}

\begin{minipage}{\textwidth}
\subsection{Atari100k}
\label{app:full_atari}
\begin{table}[H]
    \caption{
    Final performance on the \textbf{Atari100k} benchmark at 100k time steps (400k time steps in the original environment due to an action repeat of 4), averaged over 5 seeds. The results of DreamerV3 are obtained from the original paper. The {\textcolor{gray}{[bracketed values]}} represent a 95\% bootstrap confidence interval. The aggregate mean, median, and interquartile mean (IQM) are computed over the human-normalized scores (see Appendix~\ref{app:benchmark}). 
    }
    \centering
    \begin{adjustbox}{max width=\textwidth}
    \begin{tabular}{lcccccc}
        \toprule
        \textbf{Tasks} & \textbf{Rainbow (enhanced)} & \textbf{SPR} &
         \textbf{MR.Q} & \textbf{NASDAQ} & \textbf{DreamerV3} \\
        \midrule
        Alien & 
        728 {\textcolor{gray}{[665, 791]}} & 
        923 {\textcolor{gray}{[734, 1102]}} & 
        923 {\textcolor{gray}{[815, 1030]}} &
        1183 {\textcolor{gray}{[951, 1480]}} &
        1118 \\
        
        Amidar & 
        223 {\textcolor{gray}{[203, 246]}} & 
        213 {\textcolor{gray}{[181, 255]}} & 
        179 {\textcolor{gray}{[149, 210]}} &
        158 {\textcolor{gray}{[113, 203]}} &
        97 \\
        
        Assault & 
        695 {\textcolor{gray}{[600, 807]}} & 
        671 {\textcolor{gray}{[617, 726]}} & 
        601 {\textcolor{gray}{[574, 634]}} &
        676 {\textcolor{gray}{[640, 715]}} &
        683 \\
        
        Asterix & 
        924 {\textcolor{gray}{[812, 1032]}} & 
        934 {\textcolor{gray}{[853, 1021]}} & 
        1110 {\textcolor{gray}{[1024, 1195]}} &
        1320 {\textcolor{gray}{[1112, 1514]}} &
        1062 \\
        
        BankHeist & 
        63 {\textcolor{gray}{[44, 82]}} & 
        180 {\textcolor{gray}{[34, 432]}} & 
        23 {\textcolor{gray}{[19, 27]}} & 
        35 {\textcolor{gray}{[22, 50]}} &
        398 \\

        BattleZone & 
        12422 {\textcolor{gray}{[10226, 14964]}} & 
        10634 {\textcolor{gray}{[7422, 14112]}} & 
        6350 {\textcolor{gray}{[4654, 8350]}} & 
        4880 {\textcolor{gray}{[2484, 9086]}} & 
        20300 \\
        
        Boxing & 
        28 {\textcolor{gray}{[20, 37]}} & 
        38 {\textcolor{gray}{[26, 53]}} & 
        73 {\textcolor{gray}{[68, 80]}} & 
        75 {\textcolor{gray}{[68, 80]}} &
        82 \\
        
        Breakout & 
        14 {\textcolor{gray}{[12, 16]}} & 
        13 {\textcolor{gray}{[11, 15]}} & 
        18 {\textcolor{gray}{[16, 21]}} & 
        17 {\textcolor{gray}{[14, 20]}} &
        10 \\
        
        ChopperCommand & 
        1820 {\textcolor{gray}{[1554, 2093]}} & 
        1425 {\textcolor{gray}{[1165, 1685]}} & 
        1453 {\textcolor{gray}{[1298, 1619]}} & 
        1781 {\textcolor{gray}{[1606, 1953]}} &
        2222 \\
        
        CrazyClimber & 
        17741 {\textcolor{gray}{[16024, 19592]}} & 
        22957 {\textcolor{gray}{[20368, 25563]}} & 
        56564 {\textcolor{gray}{[51652, 61476]}} & 
        65146 {\textcolor{gray}{[53208, 76493]}} &
        86225 \\

        DemonAttack & 
        1525 {\textcolor{gray}{[1365, 1677]}} & 
        1519 {\textcolor{gray}{[1305, 1732]}} & 
        804 {\textcolor{gray}{[687, 933]}} & 
        2013 {\textcolor{gray}{[1319, 2721]}} &
        577\\

        Freeway & 
        6 {\textcolor{gray}{[0, 17]}} & 
        21 {\textcolor{gray}{[10, 30]}} & 
        19 {\textcolor{gray}{[6, 31]}} & 
        13 {\textcolor{gray}{[0, 25]}} &
        0 \\
        
        Frostbite & 
        1949 {\textcolor{gray}{[794, 3105]}} & 
        2862 {\textcolor{gray}{[2511, 3162]}} & 
        2296 {\textcolor{gray}{[1267, 3074]}} & 
        1939 {\textcolor{gray}{[997, 2761]}} &
        3377 \\
        
        Gopher & 
        615 {\textcolor{gray}{[482, 748]}} & 
        500 {\textcolor{gray}{[373, 644]}} & 
        739 {\textcolor{gray}{[625, 853]}} & 
        806 {\textcolor{gray}{[676, 936]}} & 
        2160 \\
        
        Hero & 
        7227 {\textcolor{gray}{[6212, 8019]}} & 
        7924 {\textcolor{gray}{[7353, 8496]}} & 
        8197 {\textcolor{gray}{[7206, 9929]}} & 
        5804 {\textcolor{gray}{[4613, 6996]}} &
        13354 \\
        
        Jamesbond & 
        399 {\textcolor{gray}{[338, 472]}} & 
        399 {\textcolor{gray}{[344, 467]}} &
        391 {\textcolor{gray}{[340, 446]}} & 
        380 {\textcolor{gray}{[316, 427]}} &
        540 \\

        Kangaroo & 
        2262 {\textcolor{gray}{[977, 4025]}} & 
        3658 {\textcolor{gray}{[1719, 5598]}} & 
        2021 {\textcolor{gray}{[669, 4368]}} & 
        476 {\textcolor{gray}{[259, 648]}} &
        2643 \\
        
        Krull & 
        3862 {\textcolor{gray}{[3199, 4537]}} & 
        4149 {\textcolor{gray}{[3603, 4810]}} & 
        7720 {\textcolor{gray}{[7049, 8530]}} & 
        7515 {\textcolor{gray}{[7061, 7914]}} &
        8171 \\
        
        KungFuMaster & 
        13644 {\textcolor{gray}{[7467, 22636]}} & 
        14945 {\textcolor{gray}{[11669, 17586]}} & 
        17707 {\textcolor{gray}{[13608, 21533]}} & 
        15509 {\textcolor{gray}{[12738, 17771]}} &
        25900  \\
        
        MsPacman & 
        1464 {\textcolor{gray}{[1248, 1648]}} & 
        1584 {\textcolor{gray}{[1345, 1941]}} & 
        1761 {\textcolor{gray}{[1336, 2186]}} & 
        1363 {\textcolor{gray}{[1229, 1498]}} &
        1521  \\
        
        Pong & 
        -8 {\textcolor{gray}{[-14, -2]}} & 
        -5 {\textcolor{gray}{[-11, 3]}} & 
        13 {\textcolor{gray}{[7, 19]}} & 
        8 {\textcolor{gray}{[0, 17]}} &
        -4  \\

        PrivateEye & 
        80 {\textcolor{gray}{[40, 100]}} & 
        86 {\textcolor{gray}{[59, 100]}} & 
        3 {\textcolor{gray}{[0, 10]}} &
        77 {\textcolor{gray}{[37, 100]}} & 
        3238\\
        
        Qbert & 
        4015 {\textcolor{gray}{[3628, 4389]}} & 
        3259 {\textcolor{gray}{[2444, 4100]}} & 
        1907 {\textcolor{gray}{[928, 3420]}} & 
        839 {\textcolor{gray}{[588, 1192]}} &
        2921 \\
        
        RoadRunner & 
        10985 {\textcolor{gray}{[4326, 17396]}} & 
        11026 {\textcolor{gray}{[5063, 16555]}} & 
        12741 {\textcolor{gray}{[10030, 16573]}} & 
        13286 {\textcolor{gray}{[9238, 17276]}} & 
        19230 \\
        
        Seaquest & 
        560 {\textcolor{gray}{[418, 670]}} & 
        556 {\textcolor{gray}{[508, 605]}} & 
        576 {\textcolor{gray}{[531, 605]}} & 
        1022 {\textcolor{gray}{[834, 1219]}} &
        962  \\
        
        UpNDown & 
        6707 {\textcolor{gray}{[4963, 8268]}} & 
        8665 {\textcolor{gray}{[4665, 13810]}} & 
        5517 {\textcolor{gray}{[4213, 6821]}} & 
        11913 {\textcolor{gray}{[5180, 24252]}} &
        46910   \\

        \midrule
        \multicolumn{4}{l}{\textbf{Aggregate Results}} \\
        \midrule
        Mean & 0.54 & 0.65 & 0.91 & 0.93 & 1.25  \\
        Median & 0.29 & 0.42 & 0.40 & 0.34 & 0.49   \\
        IQM & 0.35 & 0.43 & 0.41 & 0.42 & 0.54 \\
        \bottomrule
    \end{tabular}
    \label{tab:main_result_atari}
    \end{adjustbox}
\end{table}
\end{minipage}

\begin{figure}[H]
    \centering
    \begin{minipage}[t]{0.245\textwidth}
        % \caption*{\quad \quad Alien}
        \centering
        \includegraphics[width=\linewidth]{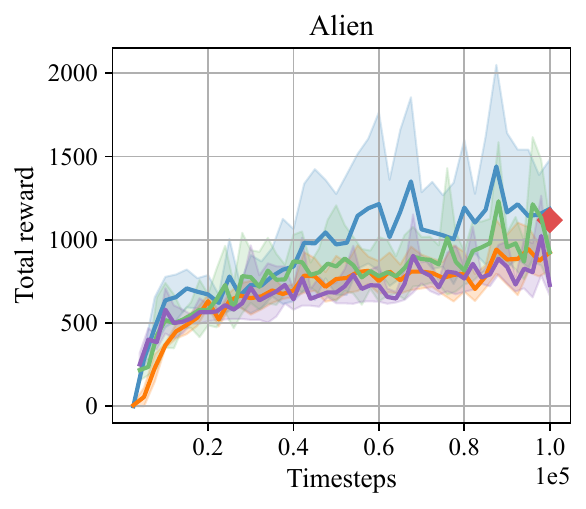}
    \end{minipage}
    \hfill
    \begin{minipage}[t]{0.245\textwidth}
        % \caption*{\quad \quad Amidar}
        \centering
        \includegraphics[width=\linewidth]{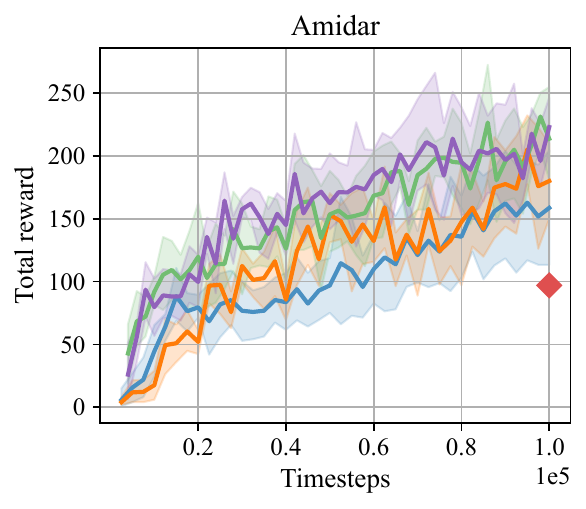}
    \end{minipage}
    \hfill
    \begin{minipage}[t]{0.245\textwidth}
        % \caption*{\quad \quad Assault}
        \centering
        \includegraphics[width=\linewidth]{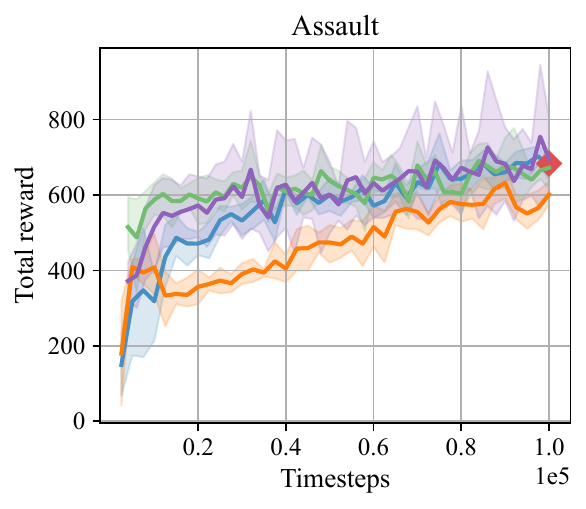}
    \end{minipage}
    \hfill
    \begin{minipage}[t]{0.245\textwidth}
        % \caption*{\quad \quad Asterix}
        \centering
        \includegraphics[width=\linewidth]{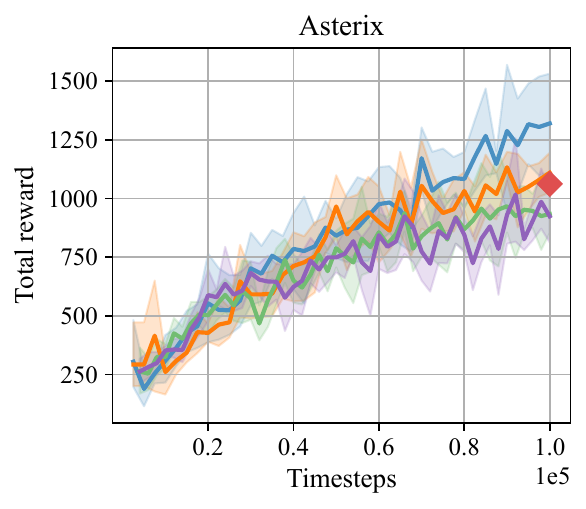}
    \end{minipage}

    \begin{minipage}[t]{0.245\textwidth}
        % \caption*{\quad \quad BankHeist}
        \centering
        \includegraphics[width=\linewidth]{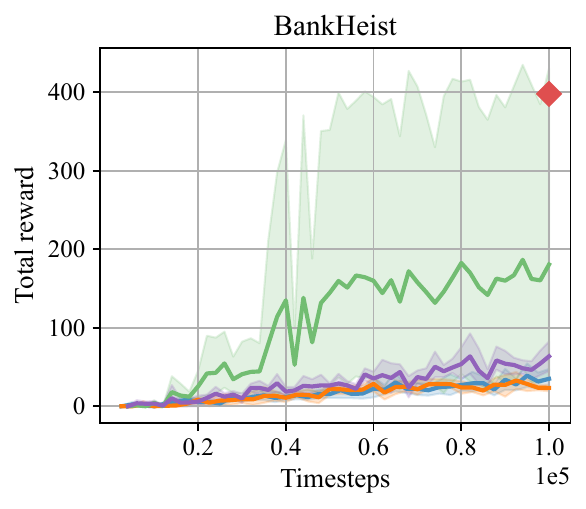}
    \end{minipage}
    \hfill
    \begin{minipage}[t]{0.245\textwidth}
        % \caption*{\quad \quad BattleZone}
        \centering
        \includegraphics[width=\linewidth]{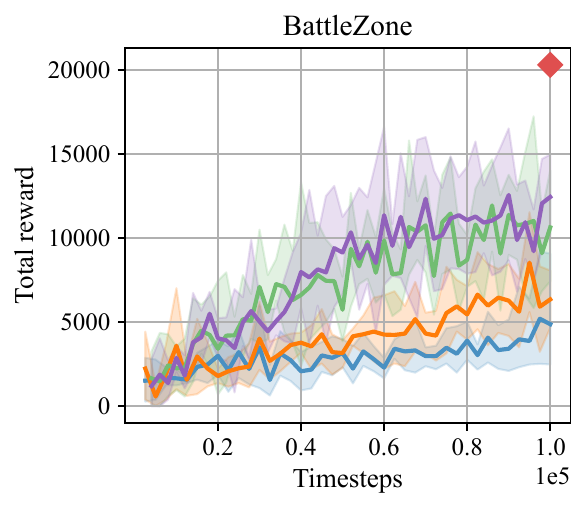}
    \end{minipage}
    \hfill
    \begin{minipage}[t]{0.245\textwidth}
        % \caption*{\quad \quad Boxing}
        \centering
        \includegraphics[width=\linewidth]{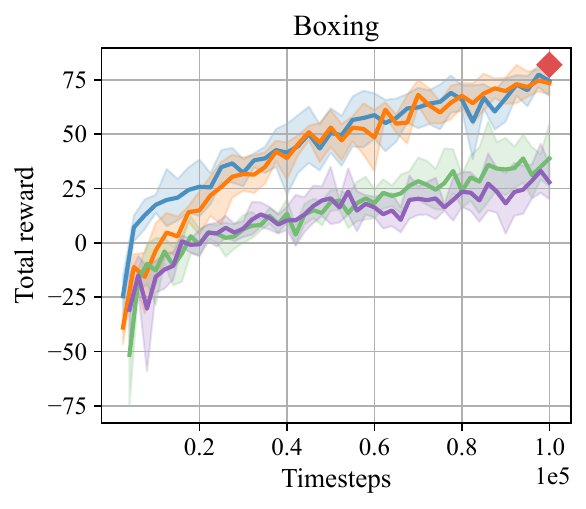}
    \end{minipage}
    \hfill
    \begin{minipage}[t]{0.245\textwidth}
        % \caption*{\quad \quad Breakout}
        \centering
        \includegraphics[width=\linewidth]{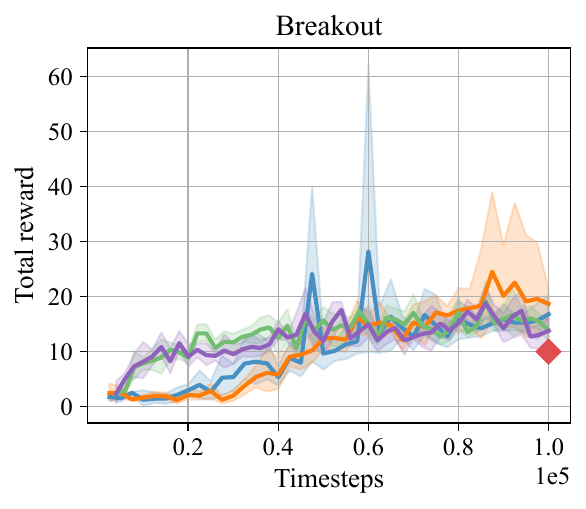}
    \end{minipage}

    \begin{minipage}[t]{0.245\textwidth}
        % \caption*{\quad \quad ChopperCommand}
        \centering
        \includegraphics[width=\linewidth]{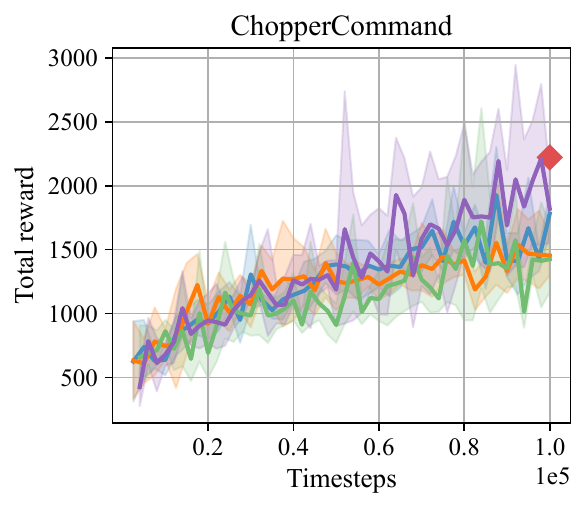}
    \end{minipage}
    \hfill
    \begin{minipage}[t]{0.245\textwidth}
        % \caption*{\quad \quad CrazyClimber}
        \centering
        \includegraphics[width=\linewidth]{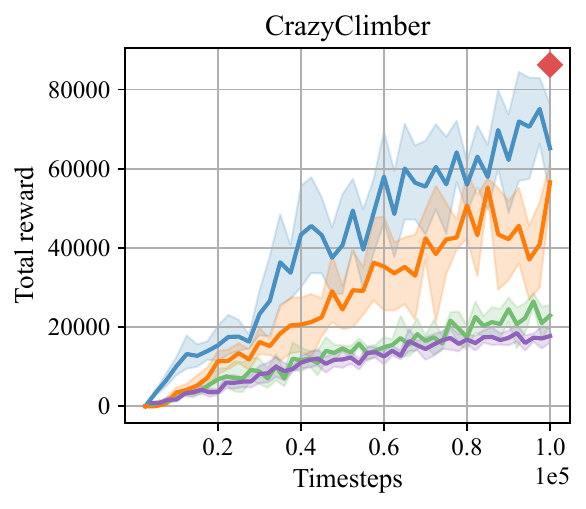}
    \end{minipage}
    \hfill
    \begin{minipage}[t]{0.245\textwidth}
        % \caption*{\quad \quad DemonAttack}
        \centering
        \includegraphics[width=\linewidth]{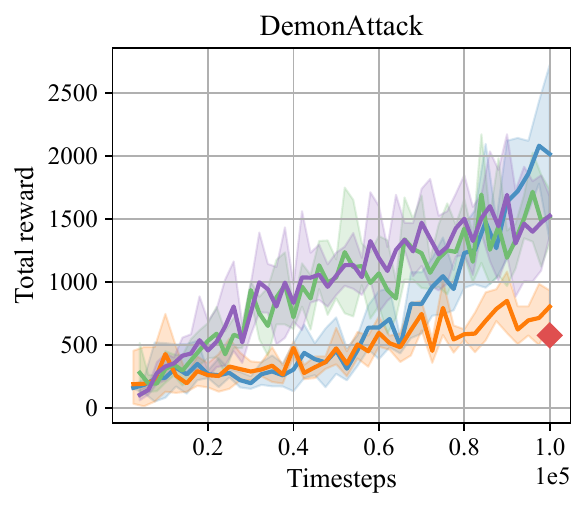}
    \end{minipage}
    \hfill
    \begin{minipage}[t]{0.245\textwidth}
        % \caption*{\quad \quad Freeway}
        \centering
        \includegraphics[width=\linewidth]{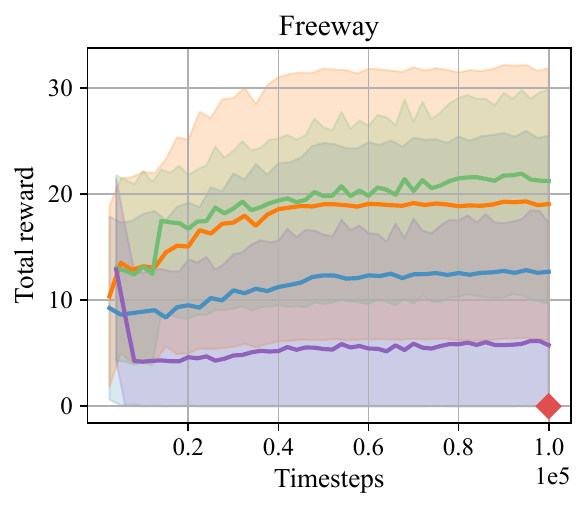}
    \end{minipage}
    \begin{minipage}[t]{0.245\textwidth}
        % \caption*{\quad \quad Frostbite}
        \centering
        \includegraphics[width=\linewidth]{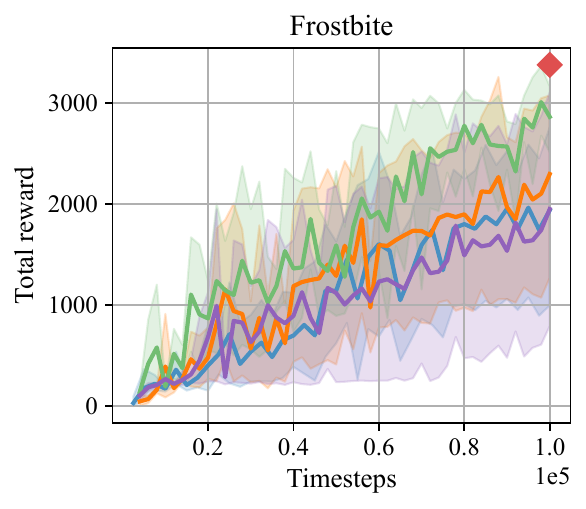}
    \end{minipage}
    \hfill
    \begin{minipage}[t]{0.245\textwidth}
        % \caption*{\quad \quad Gopher}
        \centering
        \includegraphics[width=\linewidth]{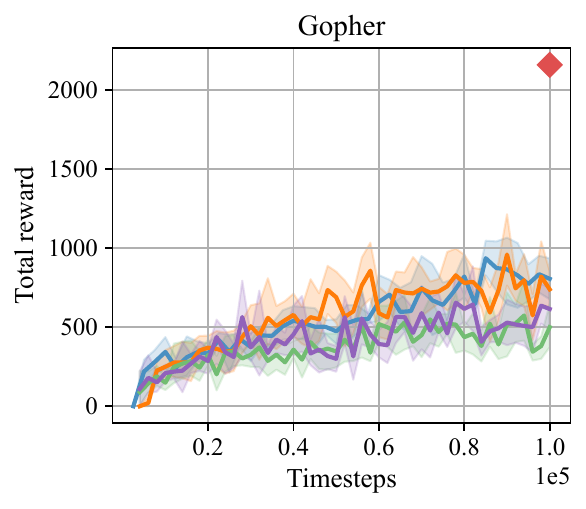}
    \end{minipage}
    \hfill
    \begin{minipage}[t]{0.245\textwidth}
        % \caption*{\quad \quad Hero}
        \centering
        \includegraphics[width=\linewidth]{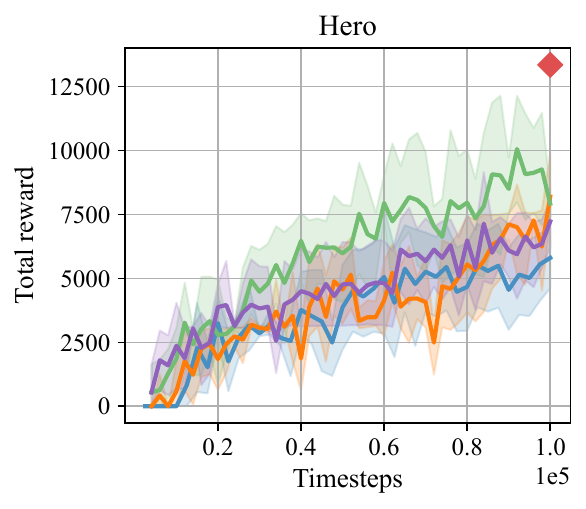}
    \end{minipage}
    \hfill
    \begin{minipage}[t]{0.245\textwidth}
        % \caption*{\quad \quad Jamesbond}
        \centering
        \includegraphics[width=\linewidth]{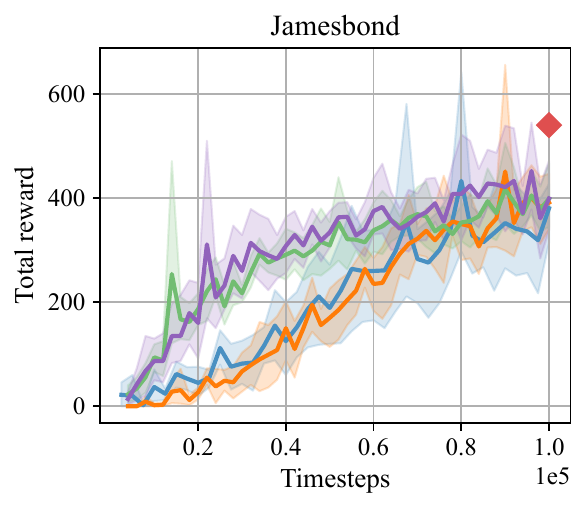}
    \end{minipage}

    \begin{minipage}[t]{0.245\textwidth}
        % \caption*{\quad \quad Kangaroo}
        \centering
        \includegraphics[width=\linewidth]{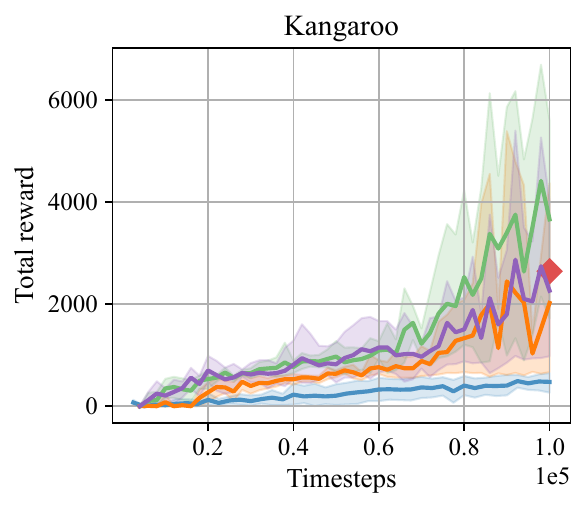}
    \end{minipage}
    \hfill
    \begin{minipage}[t]{0.245\textwidth}
        % \caption*{\quad \quad Krull}
        \centering
        \includegraphics[width=\linewidth]{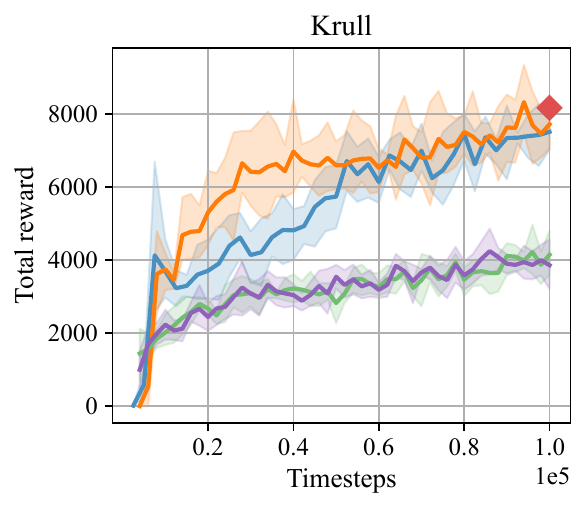}
    \end{minipage}
    \hfill
    \begin{minipage}[t]{0.245\textwidth}
        % \caption*{\quad \quad KungFuMaster}
        \centering
        \includegraphics[width=\linewidth]{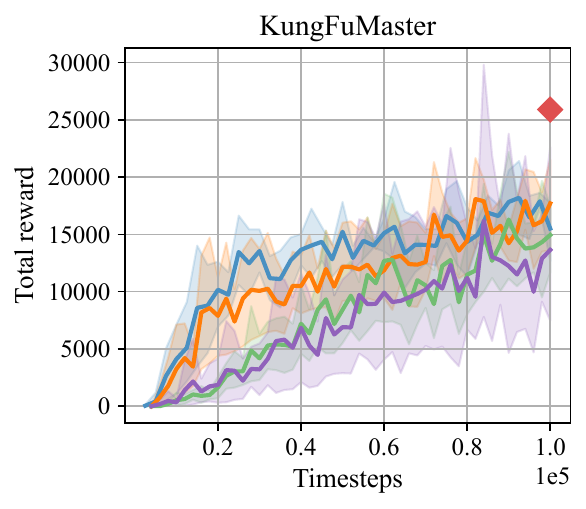}
    \end{minipage}
    \hfill
    \begin{minipage}[t]{0.245\textwidth}
        % \caption*{\quad \quad MsPacman}
        \centering
        \includegraphics[width=\linewidth]{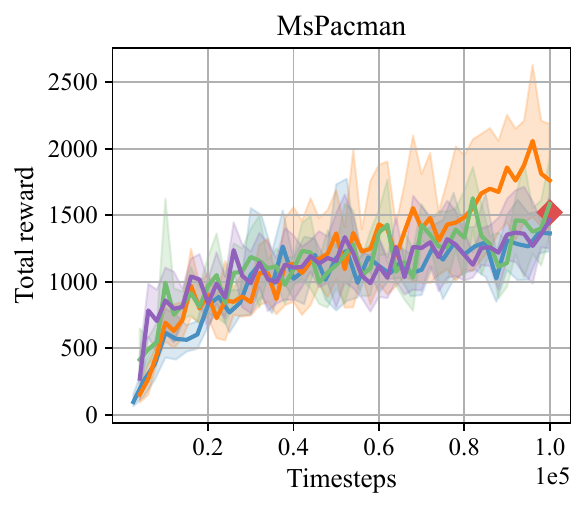}
    \end{minipage}

    \begin{minipage}[t]{0.245\textwidth}
        % \caption*{\quad \quad Pong}
        \centering
        \includegraphics[width=\linewidth]{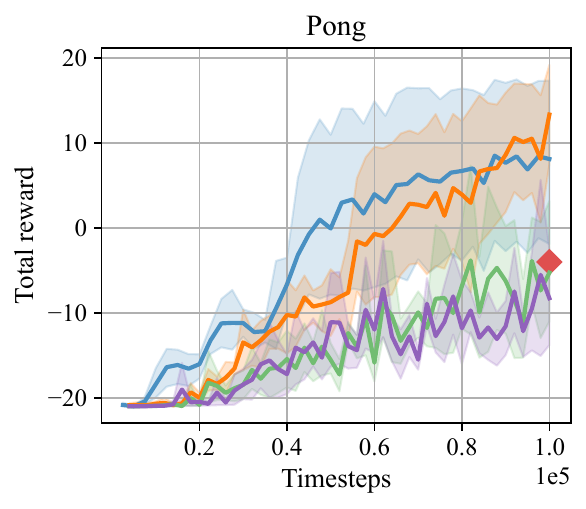}
    \end{minipage}
    \hfill
    \begin{minipage}[t]{0.245\textwidth}
        % \caption*{\quad \quad PrivateEye}
        \centering
        \includegraphics[width=\linewidth]{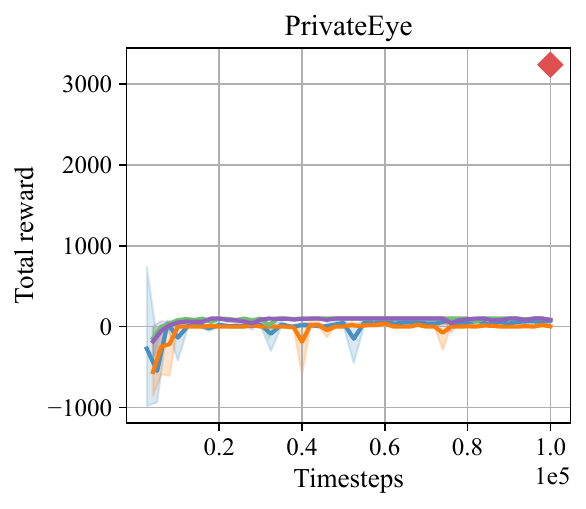}
    \end{minipage}
    \hfill
    \begin{minipage}[t]{0.245\textwidth}
        % \caption*{\quad \quad Qbert}
        \centering
        \includegraphics[width=\linewidth]{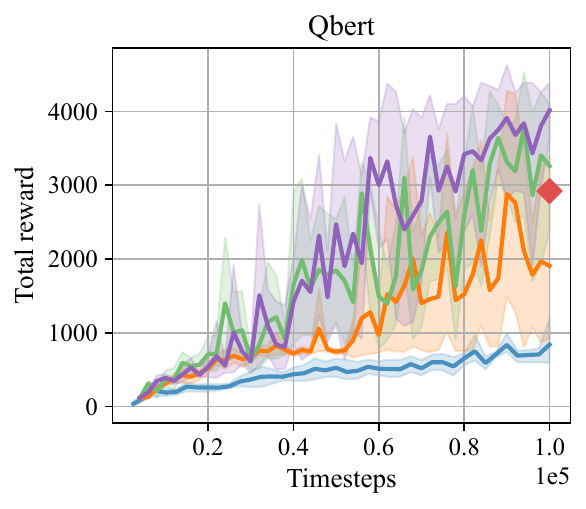}
    \end{minipage}
    \hfill
    \begin{minipage}[t]{0.245\textwidth}
        % \caption*{\quad \quad RoadRunner}
        \centering
        \includegraphics[width=\linewidth]{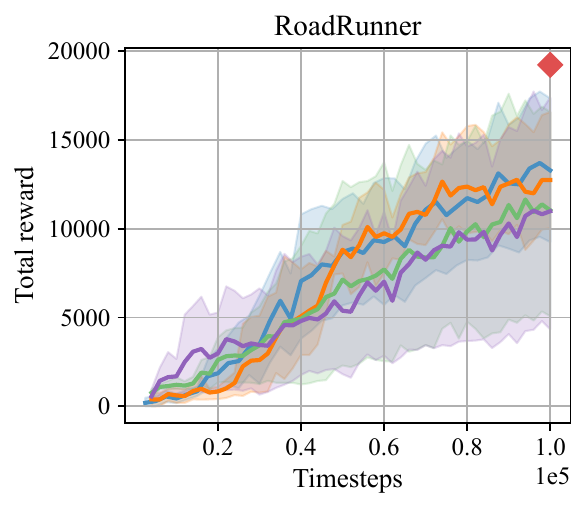}
    \end{minipage}
    \begin{minipage}{\textwidth}
        \begin{minipage}[t]{0.245\textwidth}
            % \caption*{\quad \quad Seaquest}
            \centering
            \includegraphics[width=\linewidth]{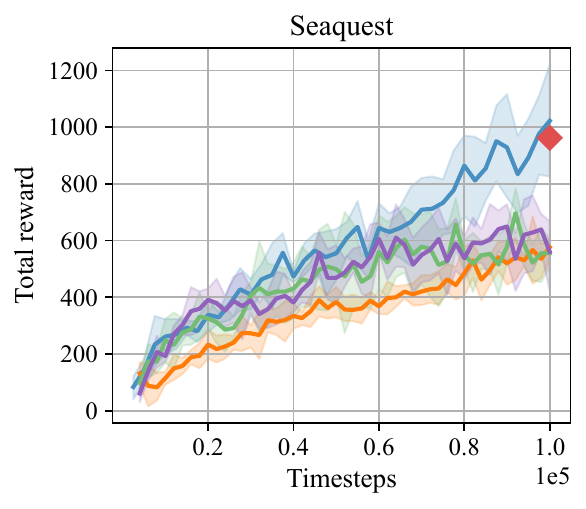}
        \end{minipage}
        % \hfill
        \begin{minipage}[t]{0.245\textwidth}
           % \caption*{\quad \quad UpNDown}
            \centering
            \includegraphics[width=\linewidth]{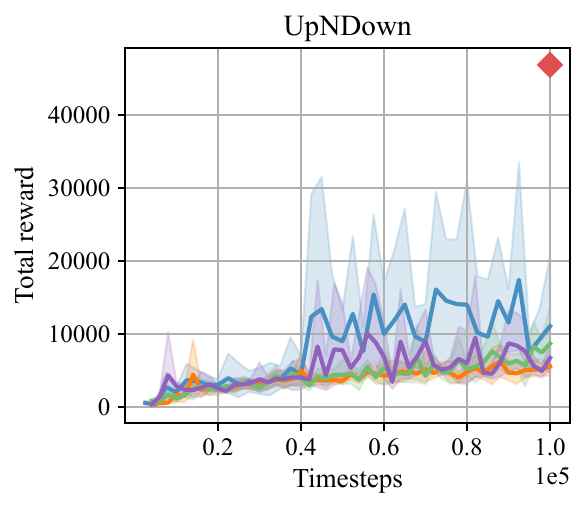}
        \end{minipage}
     \end{minipage}
 
    \begin{minipage}[t]{0.75\textwidth}
        \centering
        % \vspace{1 em}
        \includegraphics[width=\linewidth]{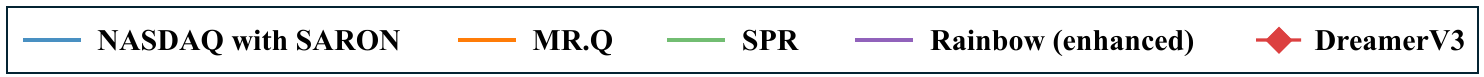}
    \end{minipage}
    \caption{
     Learning curves on \textbf{Atari100k}. Solid lines indicate average performance over 5 seeds, and shaded areas capture the 95\% bootstrap confidence interval. Discrete points denote the final results reported in DreamerV3.
    }
    \label{fig:main_result_atari2}
\end{figure}

\begin{minipage}{\textwidth}
\subsection{Ablation results of observation normalization}
\label{app:ab_on}
\begin{table}[H]
    \caption{
    Final average results of ablation variants on \textbf{Gym} over 5 seeds. The {\textcolor{gray}{[bracketed values]}} represent a 95\% bootstrap confidence interval. For comparison, we also include the results of the original OFENet+TD3 and NASDAQ with SARON (complete) without reporting confidence intervals.
    }
    \centering

    \begin{adjustbox}{max width=\textwidth}
    \begin{tabular}{l cc ccccc}
        \toprule
        \multirow{2}{*}{\textbf{Tasks}}  & \multicolumn{2}{c}{\textbf{OFENet+TD3}} & \multicolumn{5}{c}{\textbf{NASDAQ}} \\
        
        \cmidrule(lr){2-3} \cmidrule(lr){4-8}
        & Original & with SARON & w/o ON, w/o Aux & with simple ON, w/o Aux & with SARON, w/o Aux  & with simple ON & with SARON \\
        % \midrule
        % \multicolumn{8}{l}{\textbf{Gym}} \\
        \midrule

        Ant-5
        & 8156 & 8125 {\textcolor{gray}{[6912, 9048]}} & 6716 {\textcolor{gray}{[6263, 7078]}} & 
        7248 {\textcolor{gray}{[6972, 7478]}} & 6745 {\textcolor{gray}{[5824, 7453]}} & 7939 {\textcolor{gray}{[7842, 8039]}} & 7871 \\
        Humanoid-5
        & 6063 & 8470 {\textcolor{gray}{[8052, 8835]}} & 4977 {\textcolor{gray}{[4811, 5317]}} & 
        9658 {\textcolor{gray}{[9486, 9821]}} & 9258 {\textcolor{gray}{[8852, 9664]}} & 10270 {\textcolor{gray}{[10084, 10475]}} & 9938 \\        
        HalfCheetah-v5
        & 13548 & 16998 {\textcolor{gray}{[16458, 17350]}} & 16233 {\textcolor{gray}{[16015, 16525]}} & 
        16374 {\textcolor{gray}{[16062, 16658]}} & 16597 {\textcolor{gray}{[16335, 16859]}} & 16626 {\textcolor{gray}{[16138, 17112]}} & 16742 \\   
        Hopper-v5
        & 2853 & 1942 {\textcolor{gray}{[1764, 2193]}} & 1959 {\textcolor{gray}{[1608, 2433]}} & 
        1697 {\textcolor{gray}{[1449, 1855]}} & 1700 {\textcolor{gray}{[1582, 1860]}} & 1932 {\textcolor{gray}{[1659, 2255]}} & 2856 \\   
        Walker2d-v5
        & 6009 & 6207 {\textcolor{gray}{[5765, 6618]}} & 5460 {\textcolor{gray}{[5279, 5719]}} & 
        5515 {\textcolor{gray}{[5097, 5855]}} & 4653 {\textcolor{gray}{[3318, 5746]}} & 6020 {\textcolor{gray}{[5479, 6422]}} & 6077 \\  
        
        \midrule
        \multicolumn{4}{l}{\textbf{Aggregate Results}} \\
        \midrule
        Mean
        & 1.48 & 1.64 & 1.29 & 1.63 & 1.55 & 1.75 & 1.79 \\ 
        Median
        & 1.24 & 1.29 & 1.20 & 1.21 & 1.22 & 1.25 & 1.26 \\ 
        IQM
        & 1.44 & 1.54 & 1.35 & 1.40 & 1.30 & 1.50 & 1.50 \\ 
        \bottomrule
    \end{tabular}
    \end{adjustbox}
    \label{tab:full_ab1}
\end{table}
\begin{table}[H]
    \caption{
    Final average results of ablated variants on \textbf{DMC (proprioceptive)} over 5 seeds. The {\textcolor{gray}{[bracketed values]}} represent a 95\% bootstrap confidence interval. For comparison, we also include the results of the original OFENet+TD3 and NASDAQ with SARON (complete) without confidence intervals.
    }
    \centering

    \begin{adjustbox}{max width=\textwidth}
    \begin{tabular}{l cc ccccc}
        \toprule
        \multirow{2}{*}{\textbf{Tasks}}  & \multicolumn{2}{c}{\textbf{OFENet+TD3}} & \multicolumn{5}{c}{\textbf{NASDAQ}} \\
        
        \cmidrule(lr){2-3} \cmidrule(lr){4-8}
        & Original & with SARON & w/o ON, w/o Aux & with simple ON, w/o Aux & with SARON, w/o Aux  & with simple ON & with SARON \\

        % \midrule
        % \multicolumn{8}{l}{\textbf{DMC (proprioceptive)}} \\
        \midrule
        acrobot-swingup & 
        40 & 30 {\textcolor{gray}{[11, 49]}} & 234 {\textcolor{gray}{[201, 267]}} & 
        160 {\textcolor{gray}{[106, 215]}} & 155 {\textcolor{gray}{[65, 245]}} & 180 {\textcolor{gray}{[68, 291]}} & 242 \\
        
        ball\_in\_cup-catch & 
        981 & 980 {\textcolor{gray}{[977, 983]}} & 984 {\textcolor{gray}{[980, 988]}} & 
        978 {\textcolor{gray}{[975, 983]}} & 982 {\textcolor{gray}{[981, 984]}} & 981 {\textcolor{gray}{[978, 982]}} & 981 \\
        
        cartpole-balance & 
        985 & 939 {\textcolor{gray}{[859, 997]}} & 997 {\textcolor{gray}{[995, 999]}} & 
        998 {\textcolor{gray}{[997, 999]}} & 997 {\textcolor{gray}{[997, 998]}} & 998 {\textcolor{gray}{[995, 999]}} & 996 \\
        
        cartpole-balance\_sparse & 
        678 & 997 {\textcolor{gray}{[992, 1000]}} & 1000 {\textcolor{gray}{[1000, 1000]}} & 
        1000 {\textcolor{gray}{[1000, 1000]}} & 1000 {\textcolor{gray}{[1000, 1000]}} & 1000 {\textcolor{gray}{[1000, 1000]}} & 1000 \\
        
        cartpole-swingup & 
        867 & 872 {\textcolor{gray}{[868, 876]}} & 880 {\textcolor{gray}{[879, 881]}} & 
        881 {\textcolor{gray}{[880, 882]}} & 882 {\textcolor{gray}{[881, 883]}} & 880 {\textcolor{gray}{[878, 882]}} & 881 \\

        cartpole-swingup\_sparse & 
        168 & 325 {\textcolor{gray}{[0, 656]}} & 169 {\textcolor{gray}{[0, 508]}} & 
        169 {\textcolor{gray}{[0, 508]}} & 322 {\textcolor{gray}{[0, 644]}} & 174 {\textcolor{gray}{[0, 500]}} & 827 \\
        
        cheetah-run & 
        908 & 911 {\textcolor{gray}{[907, 915]}} & 890 {\textcolor{gray}{[845, 914]}} & 
        917 {\textcolor{gray}{[914, 921]}} & 918 {\textcolor{gray}{[915, 921]}} & 920 {\textcolor{gray}{[919, 921]}} & 920 \\
        
        dog-run & 
        66 & 7 {\textcolor{gray}{[5, 8]}} & 56 {\textcolor{gray}{[47, 65]}} & 
        404 {\textcolor{gray}{[388, 431]}} & 392 {\textcolor{gray}{[337, 448]}} & 313 {\textcolor{gray}{[260, 374]}} & 296 \\
        
        dog-stand & 
        347 & 28 {\textcolor{gray}{[18, 37]}} & 246 {\textcolor{gray}{[199, 305]}} & 
        920 {\textcolor{gray}{[895, 940]}} & 905 {\textcolor{gray}{[859, 953]}} & 953 {\textcolor{gray}{[930, 972]}} & 944 \\
        
        dog-trot & 
        98 & 8 {\textcolor{gray}{[6, 9]}} & 75 {\textcolor{gray}{[66, 90]}} & 
        672 {\textcolor{gray}{[526, 817]}} & 580 {\textcolor{gray}{[503, 685]}} & 619 {\textcolor{gray}{[576, 685]}} & 645 \\

        dog-walk & 
        144 & 12 {\textcolor{gray}{[10, 15]}} & 92 {\textcolor{gray}{[50, 127]}} & 
        776 {\textcolor{gray}{[691, 838]}} & 792 {\textcolor{gray}{[710, 851]}} & 618 {\textcolor{gray}{[317, 812]}} & 807 \\
        
        finger-spin & 
        977 & 981 {\textcolor{gray}{[978, 984]}} & 952 {\textcolor{gray}{[934, 970]}} & 
        969 {\textcolor{gray}{[950, 982]}} & 957 {\textcolor{gray}{[929, 979]}} & 983 {\textcolor{gray}{[979, 986]}} & 979 \\
        
        finger-turn\_easy & 
        771 & 979 {\textcolor{gray}{[972, 985]}} & 461 {\textcolor{gray}{[177, 745]}} & 
        813 {\textcolor{gray}{[624, 963]}} & 830 {\textcolor{gray}{[745, 917]}} & 957 {\textcolor{gray}{[917, 984]}} & 933 \\
        
        finger-turn\_hard & 
        466 & 944 {\textcolor{gray}{[905, 970]}} & 402 {\textcolor{gray}{[209, 594]}} & 
        644 {\textcolor{gray}{[411, 878]}} & 781 {\textcolor{gray}{[619, 894]}} & 887 {\textcolor{gray}{[823, 951]}} & 959 \\
        
        fish-swim & 
        151 & 176 {\textcolor{gray}{[112, 244]}} & 762 {\textcolor{gray}{[746, 778]}} & 
        799 {\textcolor{gray}{[780, 818]}} & 787 {\textcolor{gray}{[772, 802]}} & 792 {\textcolor{gray}{[763, 818]}} & 803 \\

        hopper-hop & 
        236 & 244 {\textcolor{gray}{[167, 361]}} & 177 {\textcolor{gray}{[96, 259]}} & 
        202 {\textcolor{gray}{[157, 242]}} & 159 {\textcolor{gray}{[115, 202]}} & 256 {\textcolor{gray}{[221, 291]}} & 284 \\
        
        hopper-stand & 
        930 & 706 {\textcolor{gray}{[506, 879]}} & 610 {\textcolor{gray}{[355, 865]}} & 
        663 {\textcolor{gray}{[361, 945]}} & 847 {\textcolor{gray}{[714, 952]}} & 926 {\textcolor{gray}{[892, 946]}} & 949 \\
        
        humanoid-run & 
        57 & 152 {\textcolor{gray}{[137, 166]}} & 25 {\textcolor{gray}{[1, 71]}} & 
        146 {\textcolor{gray}{[137, 155]}} & 135 {\textcolor{gray}{[125, 146]}} & 164 {\textcolor{gray}{[154, 181]}} & 186 \\
        
        humanoid-stand & 
        112 & 505 {\textcolor{gray}{[397, 663]}} & 171 {\textcolor{gray}{[7, 499]}} & 
        538 {\textcolor{gray}{[327, 721]}} & 712 {\textcolor{gray}{[583, 827]}} & 892 {\textcolor{gray}{[871, 907]}} & 860 \\
        
        humanoid-walk & 
        156 & 310 {\textcolor{gray}{[112, 508]}} & 84 {\textcolor{gray}{[2, 247]}} & 
        519 {\textcolor{gray}{[495, 553]}} & 534 {\textcolor{gray}{[505, 566]}} & 677 {\textcolor{gray}{[603, 789]}} & 622 \\

        pendulum-swingup & 
        380 & 227 {\textcolor{gray}{[2, 488]}} & 841 {\textcolor{gray}{[826, 856]}} & 
        205 {\textcolor{gray}{[20, 516]}} & 557 {\textcolor{gray}{[243, 871]}} & 870 {\textcolor{gray}{[856, 884]}} & 822 \\
        
        quadruped-run & 
        852 & 582 {\textcolor{gray}{[547, 617]}} & 865 {\textcolor{gray}{[834, 898]}} & 
        910 {\textcolor{gray}{[886, 935]}} & 940 {\textcolor{gray}{[922, 951]}} & 943 {\textcolor{gray}{[936, 948]}} & 935 \\
        
        quadruped-walk & 
        929 & 948 {\textcolor{gray}{[933, 961]}} & 951 {\textcolor{gray}{[938, 964]}} & 
        960 {\textcolor{gray}{[954, 965]}} & 945 {\textcolor{gray}{[933, 955]}} & 958 {\textcolor{gray}{[952, 963]}} & 960 \\
        
        reacher-easy & 
        977 & 900 {\textcolor{gray}{[777, 981]}} & 986 {\textcolor{gray}{[985, 988]}} & 
        965 {\textcolor{gray}{[922, 988]}} & 986 {\textcolor{gray}{[985, 987]}} & 985 {\textcolor{gray}{[983, 986]}} & 983 \\
        
        reacher-hard & 
        956 & 963 {\textcolor{gray}{[950, 976]}} & 958 {\textcolor{gray}{[917, 981]}} & 
        982 {\textcolor{gray}{[979, 985]}} & 981 {\textcolor{gray}{[979, 983]}} & 979 {\textcolor{gray}{[977, 980]}} & 978 \\

        walker-run & 
        714 & 770 {\textcolor{gray}{[751, 788]}} & 762 {\textcolor{gray}{[726, 786]}} & 
        802 {\textcolor{gray}{[790, 812]}} & 798 {\textcolor{gray}{[788, 808]}} & 770 {\textcolor{gray}{[715, 809]}} & 794 \\
        
        walker-stand & 
        979 & 985 {\textcolor{gray}{[982, 989]}} & 987 {\textcolor{gray}{[986, 989]}} & 
        987 {\textcolor{gray}{[984, 990]}} & 986 {\textcolor{gray}{[978, 993]}} & 988 {\textcolor{gray}{[985, 990]}} & 987 \\
        
        walker-walk & 
        959 & 963 {\textcolor{gray}{[957, 970]}} & 976 {\textcolor{gray}{[976, 977]}} & 
        973 {\textcolor{gray}{[970, 976]}} & 975 {\textcolor{gray}{[972, 979]}} & 977 {\textcolor{gray}{[974, 982]}} & 976 \\
        
        \midrule
        \multicolumn{4}{l}{\textbf{Aggregate Results}} \\
        \midrule
        Mean
        & 567 & 587 & 593 & 713 & 744 & 773 & 805 \\ 
        Median
        & 696 & 738 & 762 & 808 & 839 & 906 & 927 \\ 
        IQM
        & 600 & 656 & 645 & 804 & 834 & 886 & 898 \\ 
        \bottomrule
    \end{tabular}
    \end{adjustbox}
    \label{tab:full_ab2}
\end{table}
\end{minipage}

\begin{minipage}{\textwidth}
\subsection{Ablation results of auxiliary tasks}
\label{app:full_ab}
\begin{table}[H]
    \caption{
    Complete ablation results of auxiliary tasks on \textbf{Gym}. The {\textcolor{gray}{[bracketed values]}} represent a 95\% bootstrap confidence interval. The aggregate mean, median, and interquartile mean (IQM) are computed over the Deep-TD3-normalized scores (see Appendix~\ref{app:benchmark}).
    }
    \centering

    \begin{adjustbox}{max width=\textwidth}
    \begin{tabular}{l ccc}
        \toprule
        \multirow{2}{*}{\textbf{Tasks}}  & \multicolumn{3}{c}{\textbf{NASDAQ with SARON}} \\
        
        \cmidrule(lr){2-4} &
         $(\lambda_\text{Rec} =0, \lambda_\text{n-step} =0)$ & 
         $(\lambda_\text{Rec} >0, \lambda_\text{n-step} =0)$ &
         $(\lambda_\text{Rec} >0, \lambda_\text{n-step} >0)$ \\
        % \midrule
        % \multicolumn{3}{l}{\textbf{Gym}} \\
        \midrule
        Ant-5
        & 6745 {\textcolor{gray}{[5824, 7453]}} & 
        7871 {\textcolor{gray}{[7664, 8103]}} & 
        7813 {\textcolor{gray}{[7618, 8011]}} \\
        Humanoid-5
        & 9258 {\textcolor{gray}{[8852, 9664]}} & 
        9938 {\textcolor{gray}{[9577, 10251]}} & 
        10392 {\textcolor{gray}{[10280, 10509]}} \\
        HalfCheetah-5
        & 16597 {\textcolor{gray}{[16335, 16859]}} & 
        16742 {\textcolor{gray}{[16476, 17031]}} & 
        16444 {\textcolor{gray}{[16090, 16749]}} \\
        Hopper-5
        & 1700 {\textcolor{gray}{[1582, 1860]}} & 
        2856 {\textcolor{gray}{[2340, 3372]}}  & 
        2412 {\textcolor{gray}{[1883, 3022]}} \\
        Waler2d-5
        & 4653 {\textcolor{gray}{[3318, 5746]}} & 
        6077 {\textcolor{gray}{[5661, 6375]}} & 
        5853 {\textcolor{gray}{[5400, 6191]}} \\
        
        \midrule
        \multicolumn{3}{l}{\textbf{Aggregate Results}} \\
        \midrule
        Mean
        & 1.55 & 1.79 & 1.78  \\ 
        Median
        & 1.22 & 1.26 & 1.21  \\ 
        IQM
        & 1.30 & 1.50 & 1.47 \\ 
        \bottomrule
    \end{tabular}
    \end{adjustbox}
    \label{tab:full_ab4}
\end{table}
\end{minipage}
\begin{table}[H]
    \caption{
    Complete ablation results of auxiliary tasks on \textbf{DMC (proprioceptive)}. The {\textcolor{gray}{[bracketed values]}} represent a 95\% bootstrap confidence interval.
    }
    \centering

    \begin{adjustbox}{max width=\textwidth}
    \begin{tabular}{l ccc}
        \toprule
        \multirow{2}{*}{\textbf{Tasks}}  & \multicolumn{3}{c}{\textbf{NASDAQ with SARON}} \\
        
        \cmidrule(lr){2-4} &
         $(\lambda_\text{Rec} =0, \lambda_\text{n-step} =0)$ & 
         $(\lambda_\text{Rec} >0, \lambda_\text{n-step} =0)$ &
         $(\lambda_\text{Rec} >0, \lambda_\text{n-step} >0)$ \\
        % \midrule
        % \multicolumn{3}{l}{\textbf{DMC (proprioceptive)}} \\
        \midrule
        acrobot-swingup
        & 155 {\textcolor{gray}{[65, 245]}} & 
        242 {\textcolor{gray}{[200, 279]}} & 
        259 {\textcolor{gray}{[221, 297]}} \\
        ball\_in\_cup-catch
        & 982 {\textcolor{gray}{[981, 984]}} & 
        981 {\textcolor{gray}{[978, 983]}} & 
        979 {\textcolor{gray}{[976, 984]}} \\
        cartpole-balance
        & 997 {\textcolor{gray}{[997, 998]}} & 
        996 {\textcolor{gray}{[994, 998]}} & 
        998 {\textcolor{gray}{[996, 999]}} \\
        cartpole-balance\_sparse
        & 1000 {\textcolor{gray}{[1000, 1000]}} & 
        1000 {\textcolor{gray}{[1000, 1000]}}  & 
        1000 {\textcolor{gray}{[1000, 1000]}} \\
        cartpole-swingup
        & 882 {\textcolor{gray}{[881, 883]}} & 
        881 {\textcolor{gray}{[880, 882]}} & 
        878 {\textcolor{gray}{[870, 882]}} \\
        cartpole-swingup\_sparse  
        & 322 {\textcolor{gray}{[0, 644]}} & 
        827 {\textcolor{gray}{[814, 837]}} & 
        324 {\textcolor{gray}{[0, 655]}} \\
        cheetah-run  
        & 918 {\textcolor{gray}{[915, 921]}} & 
        920 {\textcolor{gray}{[918, 921]}} & 
        918 {\textcolor{gray}{[917, 920]}} \\
        dog-run  
        & 392 {\textcolor{gray}{[337, 448]}} & 
        296 {\textcolor{gray}{[272, 315]}} & 
        294 {\textcolor{gray}{[279, 311]}} \\
        dog-stand  
        & 905 {\textcolor{gray}{[857, 953]}} & 
        944 {\textcolor{gray}{[933, 955]}} & 
        926 {\textcolor{gray}{[891, 954]}} \\
        dog-trot  
        & 580 {\textcolor{gray}{[503, 685]}} & 
        645 {\textcolor{gray}{[552, 738]}} & 
        672 {\textcolor{gray}{[594, 755]}} \\
        dog-walk 
        & 792 {\textcolor{gray}{[710, 851]}} & 
        807 {\textcolor{gray}{[761, 856]}} & 
        808 {\textcolor{gray}{[754, 864]}} \\
        finger-spin  
        & 957 {\textcolor{gray}{[929, 979]}} & 
        979 {\textcolor{gray}{[974, 982]}} & 
        982 {\textcolor{gray}{[980, 985]}} \\
        finger-turn\_easy  
        & 830 {\textcolor{gray}{[745, 917]}} & 
        933 {\textcolor{gray}{[890, 971]}} & 
        975 {\textcolor{gray}{[968, 982]}} \\
        finger-turn\_hard  
        & 781 {\textcolor{gray}{[619, 894]}} & 
        959 {\textcolor{gray}{[936, 973]}} & 
        973 {\textcolor{gray}{[969, 977]}} \\
        fish-swim 
        & 787 {\textcolor{gray}{[772, 802]}} & 
        803 {\textcolor{gray}{[793, 813]}} & 
        792 {\textcolor{gray}{[768, 816]}} \\
        hopper-hop 
        & 159 {\textcolor{gray}{[115, 202]}} & 
        284 {\textcolor{gray}{[268, 300]}} & 
        176 {\textcolor{gray}{[109, 232]}} \\
        hopper-stand  
        & 847 {\textcolor{gray}{[714, 952]}} & 
        949 {\textcolor{gray}{[944, 953]}} & 
        814 {\textcolor{gray}{[553, 946]}} \\
        humanoid-run  
        & 135 {\textcolor{gray}{[125, 146]}} & 
        186 {\textcolor{gray}{[170, 204]}} & 
        159 {\textcolor{gray}{[153, 165]}} \\
        humanoid-stand  
        & 712 {\textcolor{gray}{[583, 827]}} & 
        860 {\textcolor{gray}{[797, 909]}} & 
        809 {\textcolor{gray}{[683, 921]}} \\
        humanoid-walk  
        & 534 {\textcolor{gray}{[505, 566]}} & 
        622 {\textcolor{gray}{[606, 639]}} & 
        651 {\textcolor{gray}{[586, 761]}} \\
        pendulum-swingup  
        & 557 {\textcolor{gray}{[243, 871]}} & 
        822 {\textcolor{gray}{[808, 838]}} & 
        348 {\textcolor{gray}{[20, 675]}} \\
        quadruped-run  
        & 940 {\textcolor{gray}{[922, 951]}} & 
        935 {\textcolor{gray}{[909, 942]}} & 
        943 {\textcolor{gray}{[938, 948]}} \\
        quadruped-walk  
        & 945 {\textcolor{gray}{[933, 955]}} & 
        960 {\textcolor{gray}{[954, 966]}} & 
        968 {\textcolor{gray}{[958, 976]}} \\
        reacher-easy  
        & 986 {\textcolor{gray}{[985, 987]}} & 
        983 {\textcolor{gray}{[982, 985]}} & 
        984 {\textcolor{gray}{[981, 987]}} \\
        reacher-hard  
        & 981 {\textcolor{gray}{[979, 983]}} & 
        978 {\textcolor{gray}{[974, 983]}} & 
        959 {\textcolor{gray}{[914, 983]}} \\
        walker-run  
        & 798 {\textcolor{gray}{[788, 808]}} & 
        794 {\textcolor{gray}{[786, 803]}} & 
        795 {\textcolor{gray}{[782, 807]}} \\
        walker-stand  
        & 986 {\textcolor{gray}{[978, 993]}} & 
        987 {\textcolor{gray}{[985, 989]}} & 
        988 {\textcolor{gray}{[986, 991]}} \\
        walker-walk  
        & 975 {\textcolor{gray}{[972, 979]}} & 
        976 {\textcolor{gray}{[975, 977]}} & 
        975 {\textcolor{gray}{[970, 980]}} \\
        
        \midrule
        \multicolumn{3}{l}{\textbf{Aggregate Results}} \\
        \midrule
        Mean
        & 744 & 805 & 762  \\ 
        Median
        & 839 & 927 & 898  \\ 
        IQM
        & 834 & 898 & 874 \\ 
        \bottomrule
    \end{tabular}
    \end{adjustbox}
    \label{tab:full_ab3}
\end{table}

\begin{table}[H]
    \caption{
    Complete ablation results of auxiliary tasks on \textbf{DMC (visual)}. The {\textcolor{gray}{[bracketed values]}} represent a 95\% bootstrap confidence interval.
    }
    \centering

    \begin{adjustbox}{max width=\textwidth}
    \begin{tabular}{l ccc}
        \toprule
        \multirow{2}{*}{\textbf{Tasks}}  & \multicolumn{3}{c}{\textbf{NASDAQ}} \\
        
        \cmidrule(lr){2-4} &
         $(\lambda_\text{Rec} =0, \lambda_\text{n-step} =0)$ & 
         $(\lambda_\text{Rec} >0, \lambda_\text{n-step} =0)$ &
         $(\lambda_\text{Rec} >0, \lambda_\text{n-step} >0)$ \\
        % \midrule
        % \multicolumn{3}{l}{\textbf{DMC (visual)}} \\
        \midrule
        acrobot-swingup
        & 312 {\textcolor{gray}{[227, 377]}} & 
        331 {\textcolor{gray}{[250, 412]}} & 
        264 {\textcolor{gray}{[171, 356]}} \\
        ball\_in\_cup-catch
        & 700 {\textcolor{gray}{[351, 973]}} & 
        971 {\textcolor{gray}{[965, 977]}} & 
        975 {\textcolor{gray}{[973, 978]}} \\
        cartpole-balance
        & 997 {\textcolor{gray}{[995, 999]}} & 
        996 {\textcolor{gray}{[991, 999]}} & 
        997 {\textcolor{gray}{[996, 998]}} \\
        cartpole-balance\_sparse
        & 998 {\textcolor{gray}{[993, 1000]}} & 
        997 {\textcolor{gray}{[990, 1000]}}  & 
        1000 {\textcolor{gray}{[1000, 1000]}} \\
        cartpole-swingup
        & 873 {\textcolor{gray}{[864, 880]}} & 
        878 {\textcolor{gray}{[875, 880]}} & 
        873 {\textcolor{gray}{[867, 878]}} \\
        cartpole-swingup\_sparse  
        & 0 {\textcolor{gray}{[0, 0]}} & 
        841 {\textcolor{gray}{[837, 845]}} & 
        835 {\textcolor{gray}{[827, 843]}} \\
        cheetah-run  
        & 705 {\textcolor{gray}{[671, 738]}} & 
        812 {\textcolor{gray}{[766, 858]}} & 
        853 {\textcolor{gray}{[821, 885]}} \\
        dog-run  
        & 25 {\textcolor{gray}{[23, 26]}} & 
        54 {\textcolor{gray}{[49, 61]}} & 
        63 {\textcolor{gray}{[48, 79]}} \\
        dog-stand  
        & 133 {\textcolor{gray}{[127, 142]}} & 
        195 {\textcolor{gray}{[177, 213]}} & 
        244 {\textcolor{gray}{[216, 269]}} \\
        dog-trot  
        & 35 {\textcolor{gray}{[32, 38]}} & 
        72 {\textcolor{gray}{[65, 79]}} & 
        63 {\textcolor{gray}{[59, 68]}} \\
        dog-walk 
        & 48 {\textcolor{gray}{[45, 51]}} & 
        88 {\textcolor{gray}{[77, 97]}} & 
        83 {\textcolor{gray}{[75, 93]}} \\
        finger-spin  
        & 522 {\textcolor{gray}{[131, 913]}} & 
        983 {\textcolor{gray}{[981, 985]}} & 
        986 {\textcolor{gray}{[985, 987]}} \\
        finger-turn\_easy  
        & 640 {\textcolor{gray}{[523, 930]}} & 
        902 {\textcolor{gray}{[817, 968]}} & 
        915 {\textcolor{gray}{[873, 956]}} \\
        finger-turn\_hard  
        & 330 {\textcolor{gray}{[193, 467]}} & 
        881 {\textcolor{gray}{[802, 950]}} & 
        914 {\textcolor{gray}{[871, 958]}} \\
        fish-swim 
        & 75 {\textcolor{gray}{[63, 91]}} & 
        84 {\textcolor{gray}{[61, 111]}} & 
        67 {\textcolor{gray}{[64, 70]}} \\
        hopper-hop 
        & 62 {\textcolor{gray}{[0, 186]}} & 
        276 {\textcolor{gray}{[258, 298]}} & 
        283 {\textcolor{gray}{[256, 308]}} \\
        hopper-stand  
        & 144 {\textcolor{gray}{[5, 413]}} & 
        923 {\textcolor{gray}{[915, 931]}} & 
        902 {\textcolor{gray}{[850, 934]}} \\
        humanoid-run  
        & 1 {\textcolor{gray}{[1, 1]}} & 
        1 {\textcolor{gray}{[1, 1]}} &  
        1 {\textcolor{gray}{[1, 1]}} \\
        humanoid-stand  
        & 6 {\textcolor{gray}{[6, 7]}} & 
        8 {\textcolor{gray}{[7, 9]}} & 
        6 {\textcolor{gray}{[5, 7]}} \\
        humanoid-walk  
        & 2 {\textcolor{gray}{[2, 2]}} & 
        3 {\textcolor{gray}{[3, 4]}} & 
        2 {\textcolor{gray}{[2, 3]}} \\
        pendulum-swingup  
        & 581 {\textcolor{gray}{[283, 850]}} & 
        592 {\textcolor{gray}{[349, 807]}} & 
        836 {\textcolor{gray}{[816, 858]}} \\
        quadruped-run  
        & 413 {\textcolor{gray}{[394, 431]}} & 
        463 {\textcolor{gray}{[459, 467]}} &
        471 {\textcolor{gray}{[458, 495]}} \\
        quadruped-walk  
        & 662 {\textcolor{gray}{[557, 729]}} & 
        772 {\textcolor{gray}{[706, 839]}} & 
        809 {\textcolor{gray}{[748, 855]}} \\
        reacher-easy  
        & 974 {\textcolor{gray}{[971, 976]}} & 
        938 {\textcolor{gray}{[867, 976]}} & 
        977 {\textcolor{gray}{[975, 979]}} \\
        reacher-hard  
        & 178 {\textcolor{gray}{[68, 290]}} & 
        931 {\textcolor{gray}{[852, 973]}} & 
        935 {\textcolor{gray}{[893, 977]}} \\
        walker-run  
        & 501 {\textcolor{gray}{[464, 528]}} & 
        554 {\textcolor{gray}{[423, 653]}} & 
        676 {\textcolor{gray}{[651, 704]}} \\
        walker-stand  
        & 973 {\textcolor{gray}{[969, 977]}} & 
        986 {\textcolor{gray}{[983, 989]}} & 
        987 {\textcolor{gray}{[985, 989]}} \\
        walker-walk  
        & 912 {\textcolor{gray}{[836, 955]}} & 
        964 {\textcolor{gray}{[962, 966]}} & 
        963 {\textcolor{gray}{[958, 968]}} \\
        
        \midrule
        \multicolumn{3}{l}{\textbf{Aggregate Results}} \\
        \midrule
        Mean
        & 422 & 589 & 606  \\ 
        Median
        & 372 & 792 & 836  \\ 
        IQM
        & 375 & 668 & 701 \\ 
        \bottomrule
    \end{tabular}
    \end{adjustbox}
    \label{tab:full_ab5}
\end{table}

\begin{table}[H]
    \caption{
    Complete ablation results of auxiliary tasks on \textbf{Atari100k}. The {\textcolor{gray}{[bracketed values]}} represent a 95\% bootstrap confidence interval. The aggregate mean, median, and interquartile mean (IQM) are computed over the human-normalized score (see Appendix~\ref{app:benchmark}).
    }
    \centering

    \begin{adjustbox}{max width=\textwidth}
    \begin{tabular}{l ccc}
        \toprule
        \multirow{2}{*}{\textbf{Tasks}}  & \multicolumn{3}{c}{\textbf{NASDAQ}} \\
        
        \cmidrule(lr){2-4} &
         $(\lambda_\text{Rec} =0, \lambda_\text{n-step} =0)$ & 
         $(\lambda_\text{Rec} >0, \lambda_\text{n-step} =0)$ &
         $(\lambda_\text{Rec} >0, \lambda_\text{n-step} >0)$ \\
        % \midrule
        % \multicolumn{3}{l}{\textbf{Atari100k}} \\
        \midrule

        Alien & 
        900 {\textcolor{gray}{[824, 970]}} & 
        915 {\textcolor{gray}{[775, 1032]}} & 
        1183 {\textcolor{gray}{[951, 1480]}} \\
        
        Amidar & 
        149 {\textcolor{gray}{[135, 161]}} & 
        154 {\textcolor{gray}{[137, 175]}} & 
        158 {\textcolor{gray}{[113, 203]}} \\
        
        Assault & 
        431 {\textcolor{gray}{[298, 565]}} & 
        637 {\textcolor{gray}{[604, 672]}} & 
        676 {\textcolor{gray}{[640, 715]}} \\
        
        Asterix & 
        899 {\textcolor{gray}{[737, 1060]}} & 
        1235 {\textcolor{gray}{[1095, 1375]}} & 
        1320 {\textcolor{gray}{[1112, 1514]}} \\
        
        BankHeist & 
        20 {\textcolor{gray}{[10, 28]}} & 
        30 {\textcolor{gray}{[21, 42]}} & 
        35 {\textcolor{gray}{[22, 50]}} \\

        BattleZone & 
        5544 {\textcolor{gray}{[2180, 12128]}} & 
        3058 {\textcolor{gray}{[2670, 3540]}} & 
        4880 {\textcolor{gray}{[2484, 9086]}} \\
        
        Boxing & 
        24 {\textcolor{gray}{[21, 30]}} & 
        65 {\textcolor{gray}{[60, 70]}} & 
        75 {\textcolor{gray}{[68, 80]}} \\
        
        Breakout & 
        1 {\textcolor{gray}{[1, 1]}} & 
        18 {\textcolor{gray}{[15, 22]}} & 
        17 {\textcolor{gray}{[14, 20]}} \\
        
        ChopperCommand & 
        1679 {\textcolor{gray}{[1404, 1926]}} & 
        1682 {\textcolor{gray}{[1340, 1964]}} & 
        1781 {\textcolor{gray}{[1606, 1953]}} \\
        
        CrazyClimber & 
        23282 {\textcolor{gray}{[21142, 25339]}} & 
        57314 {\textcolor{gray}{[35783, 73445]}} & 
        65146 {\textcolor{gray}{[53208, 76493]}} \\

        DemonAttack & 
        160 {\textcolor{gray}{[153, 165]}} & 
        1821 {\textcolor{gray}{[1580, 2039]}} & 
        2013 {\textcolor{gray}{[1319, 2721]}} \\

        Freeway & 
        25 {\textcolor{gray}{[12, 31]}} & 
        0 {\textcolor{gray}{[0, 0]}} & 
        13 {\textcolor{gray}{[0, 25]}} \\
        
        Frostbite & 
        1612 {\textcolor{gray}{[1478, 1766]}} & 
        2211 {\textcolor{gray}{[1241, 2856]}} & 
        1939 {\textcolor{gray}{[997, 2761]}} \\
        
        Gopher & 
        406 {\textcolor{gray}{[339, 498]}} & 
        673 {\textcolor{gray}{[592, 763]}} & 
        806 {\textcolor{gray}{[676, 936]}} \\
        
        Hero & 
        3570 {\textcolor{gray}{[2733, 5117]}} & 
        3897 {\textcolor{gray}{[2790, 6852]}} & 
        5804 {\textcolor{gray}{[4613, 6996]}} \\
        
        Jamesbond & 
        117 {\textcolor{gray}{[31, 289]}} & 
        349 {\textcolor{gray}{[299, 400]}} & 
        380 {\textcolor{gray}{[316, 427]}} \\

        Kangaroo & 
        1978 {\textcolor{gray}{[147, 5211]}} & 
        94 {\textcolor{gray}{[7, 222]}} & 
        476 {\textcolor{gray}{[259, 648]}} \\
        
        Krull & 
        3111 {\textcolor{gray}{[2898, 3288]}} & 
        6596 {\textcolor{gray}{[6093, 7121]}} & 
        7515 {\textcolor{gray}{[7061, 7914]}} \\
        
        KungFuMaster & 
        8569 {\textcolor{gray}{[6936, 10431]}} & 
        11372 {\textcolor{gray}{[9778, 12967]}} & 
        15509 {\textcolor{gray}{[12738, 17771]}} \\
        
        MsPacman & 
        1435 {\textcolor{gray}{[1130, 1776]}} & 
        1166 {\textcolor{gray}{[999, 1324]}} & 
        1363 {\textcolor{gray}{[1229, 1498]}} \\
        
        Pong & 
        -2 {\textcolor{gray}{[-9, 7]}} & 
        10 {\textcolor{gray}{[0, 18]}} & 
        8 {\textcolor{gray}{[0, 17]}} \\

        PrivateEye & 
        40 {\textcolor{gray}{[0, 80]}} &
        20 {\textcolor{gray}{[0, 60]}} & 
        77 {\textcolor{gray}{[37, 100]}} \\
        
        Qbert & 
        2074 {\textcolor{gray}{[1165, 2986]}} & 
        697 {\textcolor{gray}{[637, 750]}} & 
        839 {\textcolor{gray}{[588, 1192]}} \\
        
        RoadRunner & 
        8662 {\textcolor{gray}{[1869, 15455]}} & 
        11611 {\textcolor{gray}{[9485, 15026]}} & 
        13286 {\textcolor{gray}{[9238, 17276]}} \\
        
        Seaquest & 
        442 {\textcolor{gray}{[366, 527]}} & 
        798 {\textcolor{gray}{[699, 900]}} & 
        1022 {\textcolor{gray}{[834, 1219]}} \\
        
        UpNDown & 
        4580 {\textcolor{gray}{[4224, 4961]}} & 
        28976 {\textcolor{gray}{[18459, 38070]}} & 
        11913 {\textcolor{gray}{[5180, 24252]}} \\
        
        \midrule
        \multicolumn{3}{l}{\textbf{Aggregate Results}} \\
        \midrule
        Mean
        & 0.38 & 0.86 & 0.93  \\ 
        Median
        & 0.16 & 0.16 & 0.34  \\ 
        IQM
        & 0.23 & 0.36 & 0.42 \\ 
        \bottomrule
    \end{tabular}
    \end{adjustbox}
    \label{tab:full_ab_atari}
\end{table}

\begin{minipage}{\textwidth}
% section4.1的结果
\subsection{Results on the diagnostic set of tasks}
\label{app:add_result}
\hfill
\begin{figure}[H]
    \centering
    % % ===== 第一组标题 =====
    % \textbf{Auxiliary loss across observation dimensions}

    \begin{minipage}[t]{0.32\textwidth}
        % \caption*{\quad \quad HalfCheetah-v5}
        \centering
        \includegraphics[width=\linewidth]{images/prelab2/Humanoid-v5-aux_loss.pdf}
    \end{minipage}
    \hfill
    \begin{minipage}[t]{0.32\textwidth}
        % \caption*{\quad \quad Hopper-v5}
        \centering
        \includegraphics[width=\linewidth]{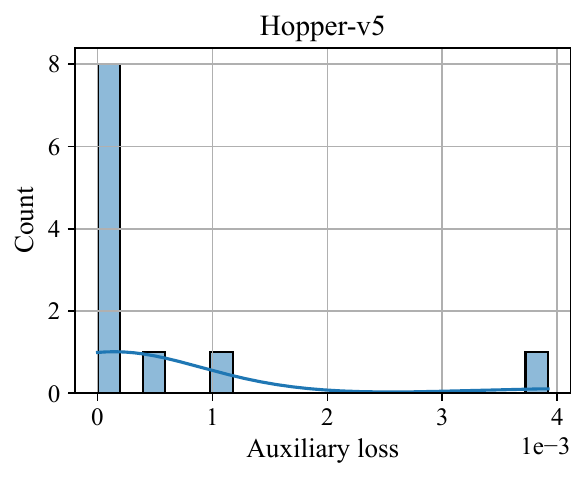}
    \end{minipage}
    \hfill
    \begin{minipage}[t]{0.32\textwidth}
        % \caption*{\quad \quad Walker2d-v5}
        \centering
        \includegraphics[width=\linewidth]{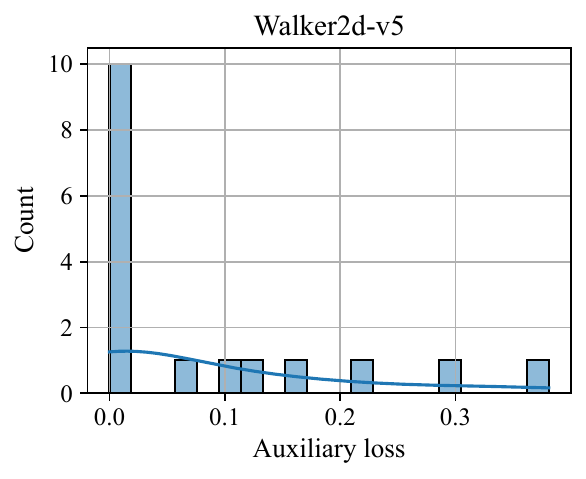}
    \end{minipage}

    % % ===== 第二组标题 =====
    % \textbf{Standard deviation across observation dimensions}

    \begin{minipage}[t]{0.32\textwidth}
        % \caption*{\quad \quad HalfCheetah-v5}
        \centering
        \includegraphics[width=\linewidth]{images/prelab2/Humanoid-v5-std.pdf}
    \end{minipage}
    \hfill
    \begin{minipage}[t]{0.32\textwidth}
        % \caption*{\quad \quad Hopper-v5}
        \centering
        \includegraphics[width=\linewidth]{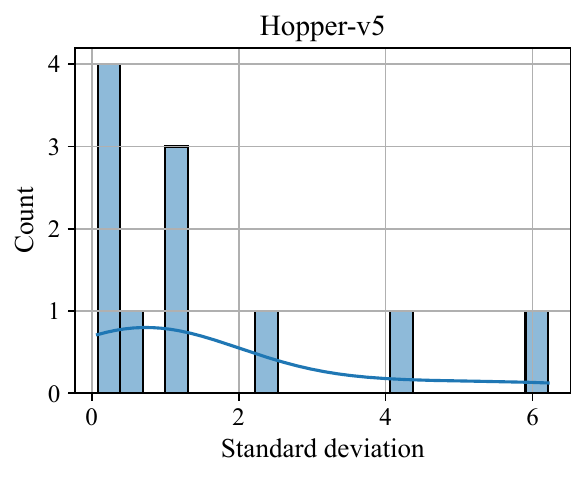}
    \end{minipage}
    \hfill
    \begin{minipage}[t]{0.32\textwidth}
        % \caption*{\quad \quad Walker2d-v5}
        \centering
        \includegraphics[width=\linewidth]{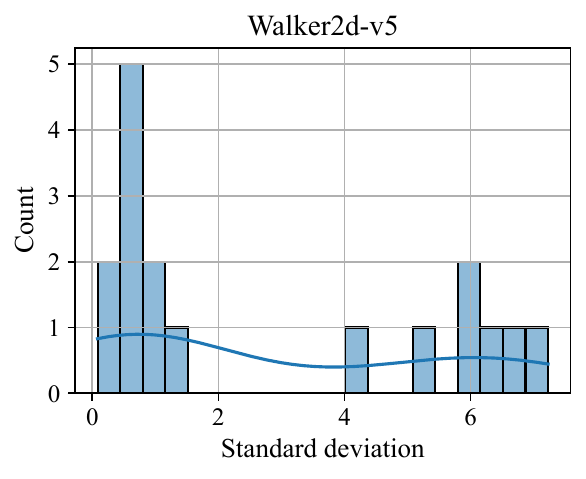}
    \end{minipage}

    \caption{Histograms and kernel density-estimated PDFs of per-dimension statistics, obtained from a single run of \textbf{OFENet+TD3} on three tasks. 
    \textit{Top}: distribution of auxiliary losses for each observation dimension, where the value for each dimension is computed by averaging the recorded auxiliary losses over the final 20k time steps.
    \textit{Bottom}: distribution of standard deviation over dimensions, computed from the final 20k samples.}
    \label{tab:prelab_extra}
\end{figure}
\hfill
\end{minipage}

\begin{figure}[H]
    \centering
    % ===== 第一组标题 =====
    \begin{minipage}[t]{0.32\textwidth}
        % \caption*{\quad Ant-v5 \\ \quad  ($r=0.42, \rho=0.81$)}
        % \vspace{0.7 em}
        \centering
        \includegraphics[width=\linewidth]{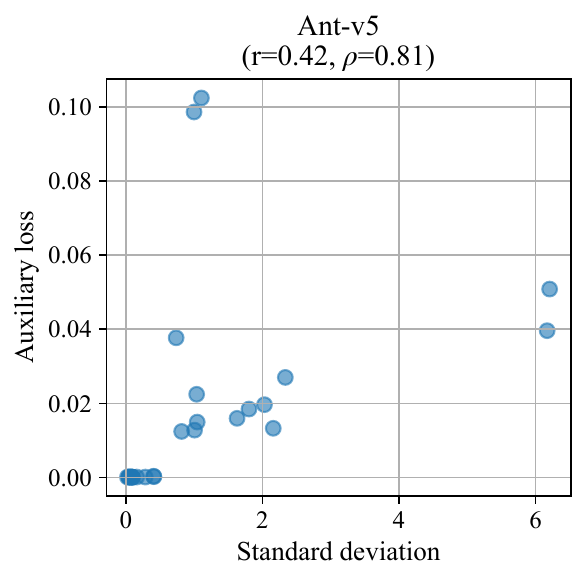}
    \end{minipage}
    \hfill
    \begin{minipage}[t]{0.32\textwidth}
        % \caption*{ Humanoid-v5 \\  ($r=0.94, \rho=0.38$)}
        \centering
        \includegraphics[width=\linewidth]{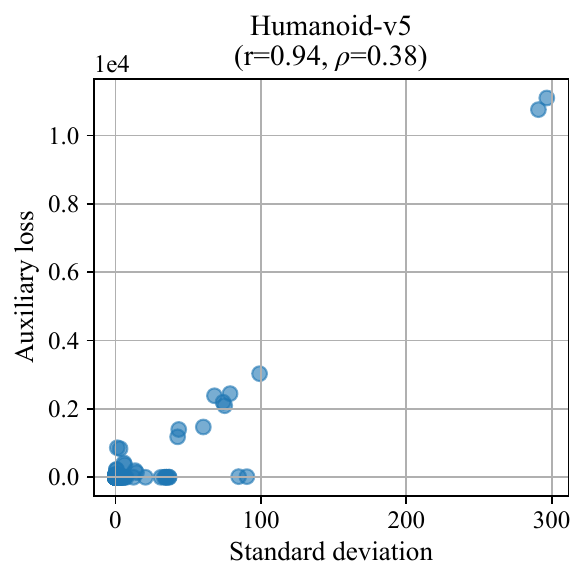}
    \end{minipage}
    \hfill
    \begin{minipage}[t]{0.32\textwidth}
        % \caption*{\quad dog-run \\ \quad ($r=0.62, \rho=0.64$)}
        \centering
        \includegraphics[width=\linewidth]{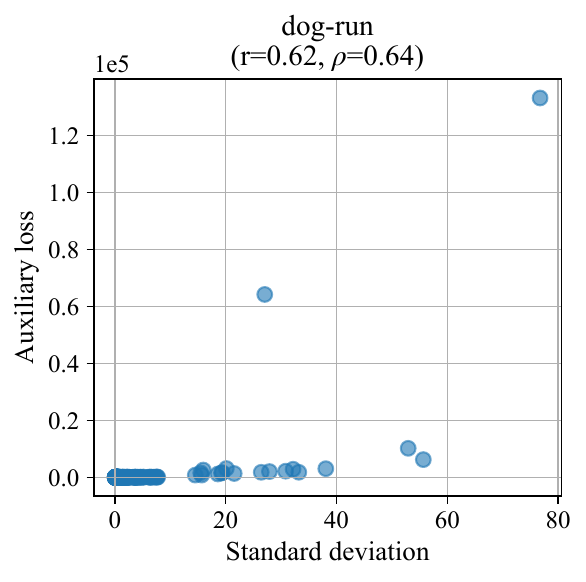}
    \end{minipage}
    % \vspace{1 em}
    \begin{minipage}[t]{0.32\textwidth}

        \centering
        % \caption*{ HalfCheetah-v5 \\ ($r=0.96, \rho=0.92$)}
        \includegraphics[width=\linewidth]{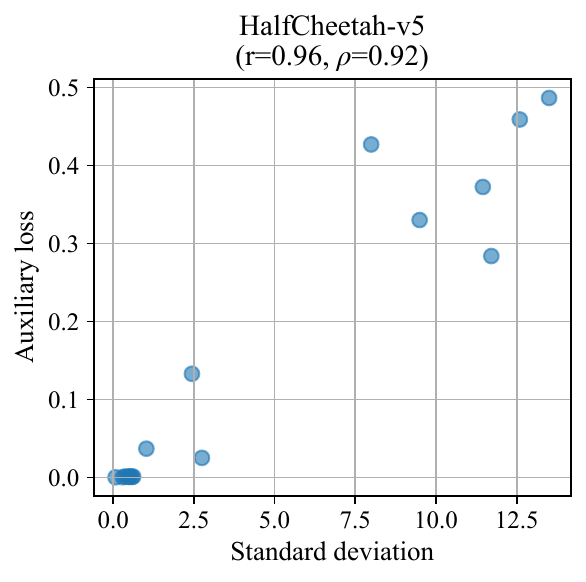}
    \end{minipage}
    \hfill
    \begin{minipage}[t]{0.32\textwidth}
        % \caption*{ Hopper-v5 \\  ($r=0.91, \rho=0.94$)}
        \centering
        \includegraphics[width=\linewidth]{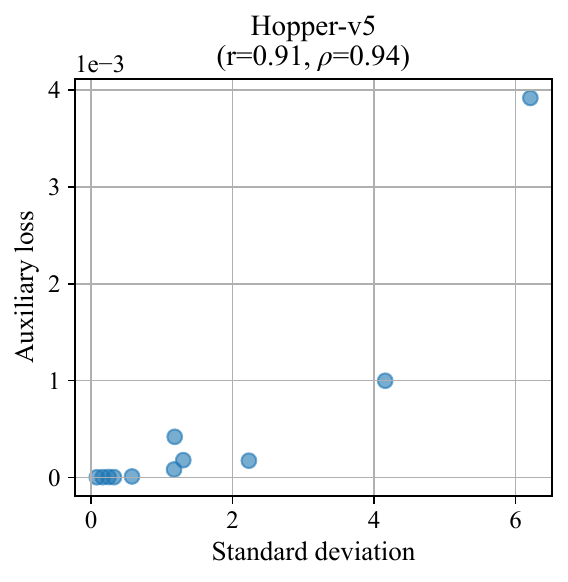}
    \end{minipage}
    \hfill
    \begin{minipage}[t]{0.32\textwidth}
        % \caption*{Walker2d-v5 \\ ($r=0.85, \rho=0.94$)}
        \centering
        % \vspace{0.7 em}
        \includegraphics[width=\linewidth]{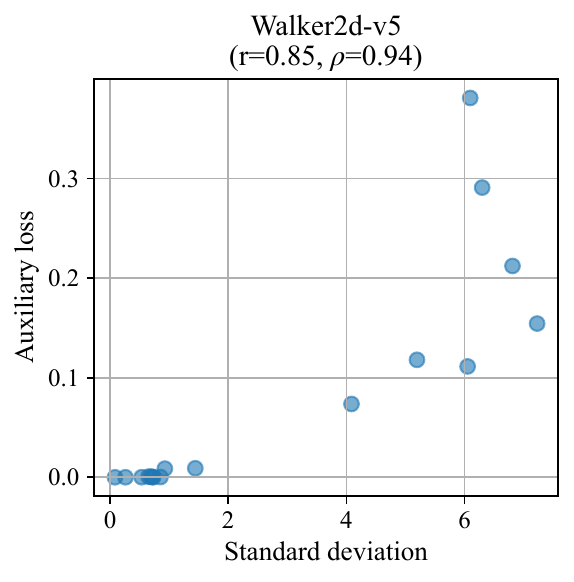}
    \end{minipage}

    \caption{Scatter plots of auxiliary loss versus standard deviation for each observation dimension on the diagnostic set of tasks. Data is obtained from a single run of \textbf{OFENet+TD3}. Each point corresponds to one observation dimension. Subplot titles report the Spearman rank ($\rho$) and Pearson correlation coefficient ($r$), quantifying the monotonic and linear relationships, respectively.}
    \label{tab:prelab_scatter}
\end{figure}

\begin{figure}[H]
    \centering
    % ===== 第一组标题 =====
    \begin{minipage}[t]{0.32\textwidth}
        \centering
        \includegraphics[width=\linewidth]{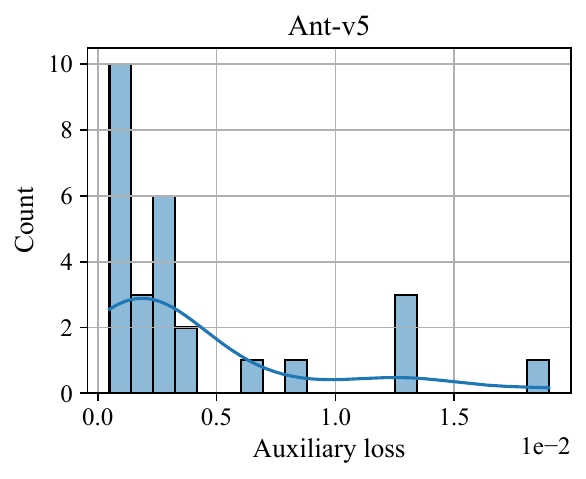}
    \end{minipage}
    \hfill
    \begin{minipage}[t]{0.32\textwidth}
        \centering
        \includegraphics[width=\linewidth]{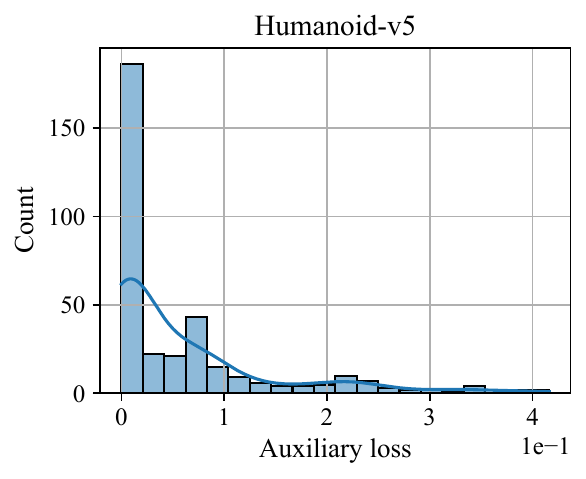}
    \end{minipage}
    \hfill
    \begin{minipage}[t]{0.32\textwidth}
        \centering
        \includegraphics[width=\linewidth]{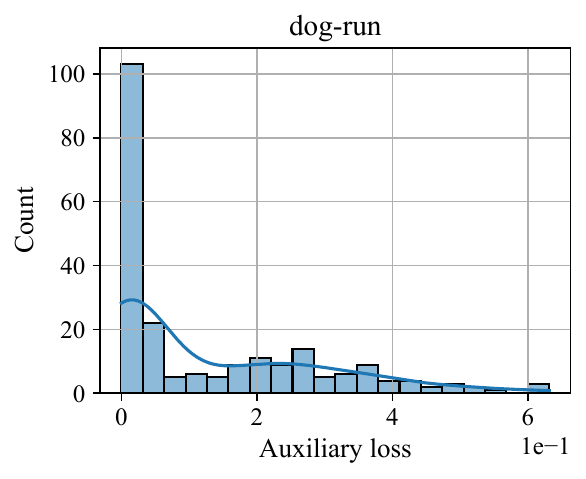}
    \end{minipage}
    % \vspace{1 em}
    \begin{minipage}[t]{0.32\textwidth}
        \centering
        \includegraphics[width=\linewidth]{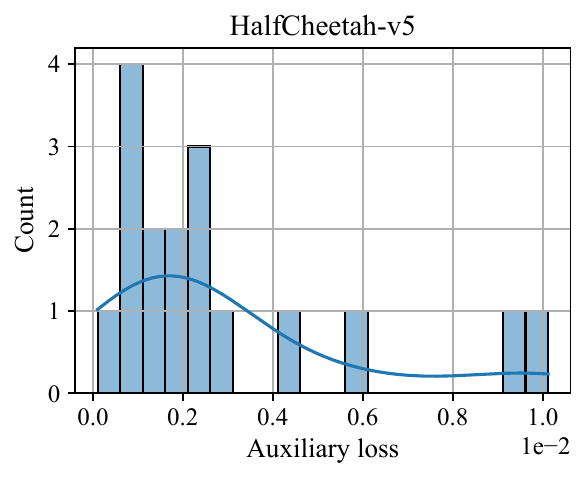}
    \end{minipage}
    \hfill
    \begin{minipage}[t]{0.32\textwidth}
        \centering
        \includegraphics[width=\linewidth]{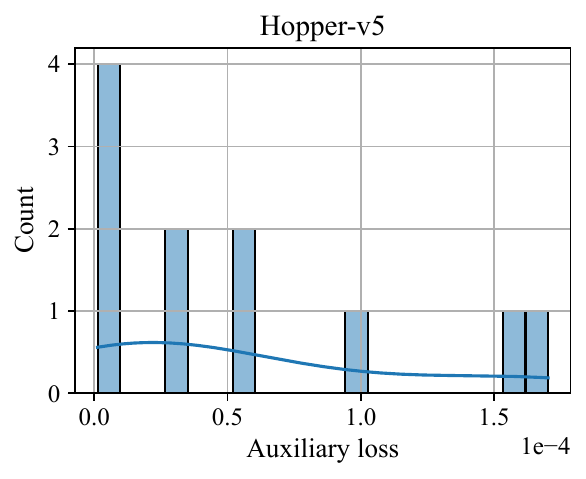}
    \end{minipage}
    \hfill
    \begin{minipage}[t]{0.32\textwidth}
        \centering
        \includegraphics[width=\linewidth]{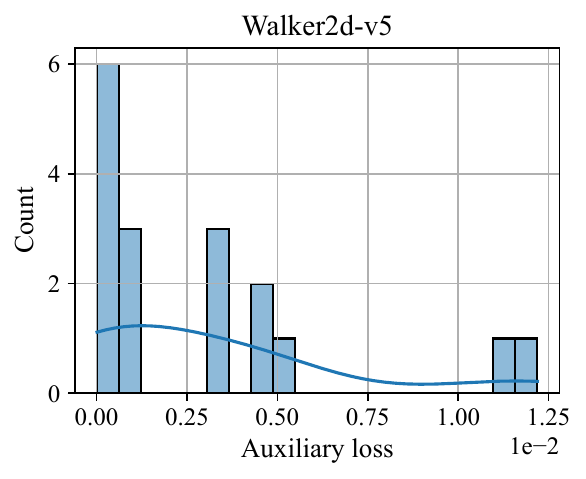}
    \end{minipage}

    \caption{Histograms and kernel density estimated PDFs of per-dimension auxiliary loss across the diagnostic set of tasks. Results are obtained from a single run of \textbf{OFENet+TD3 with SARON}. The value for each dimension is computed by averaging the recorded auxiliary losses over the final 20k time steps.}
    \label{tab:after_obs_norm}
\end{figure}

\section{Broader impacts}
\label{app:impacts}
Our proposed methods, SARON and NASDAQ, enable observation-predictive RL to match or exceed state-of-the-art model-based and self-predictive RL in performance while requiring significantly less training wall-time. This could help reduce energy consumption and carbon emissions when training agents. In addition, our experimental design promotes fairer comparisons among methods that share underlying principles (e.g., fine-grained alignment of design choices and parameter control described in Section~\ref{sec:exp}). We hope this inspires the community to adopt more rigorous and fair experimental practices.

As with any work that improves efficiency, our methods lower the barrier to using RL, which may lead to an increased risk of misuse. We acknowledge that improvements in efficiency should be accompanied by advances in social oversight and governance mechanisms.

% \subsection{Ablation Study}
% \subsection   {Sensitivity Analysis}

% % ====== Appendix C ======
% \section{Proofs}
% \label{app:proof}

% \subsection{Proof of Theorem 1}

%%%%%%%%%%%%%%%%%%%%%%%%%%%%%%%%%%%%%%%%%%%%%%%%%%%%%%%%%%%%

% \newpage
% \input{checklist.tex}

%%%%%%%%%%%%%%%%%%%%%%%%%%%%%%%%%%%%%%%%%%%%%%%%%%%%%%%%%%%%

\end{document}